\pdfoutput=1

\documentclass[twoside]{article}

\usepackage[dvipsnames]{xcolor}
\usepackage[accepted]{aistats2021}

\usepackage[activate={true},final,tracking=true,kerning=true,spacing=true,factor=1100,stretch=10,shrink=20]{microtype}
\usepackage{graphicx}
\usepackage{subcaption}
\usepackage{booktabs} %
\usepackage{multirow}
\usepackage{multicol}
\usepackage[most]{tcolorbox}
\usepackage[T1]{fontenc}

\usepackage{standalone}
\usepackage{tikz}
\usepackage{pgfplots}
\usepackage{pgf}
\usepackage{amsmath}
\usetikzlibrary{calc}
\usetikzlibrary{positioning}
\usetikzlibrary{angles,quotes}
\usetikzlibrary{backgrounds}
\usetikzlibrary{fit}
\usetikzlibrary{arrows}
\usetikzlibrary{arrows.meta}
\usetikzlibrary{shapes.symbols}
\usetikzlibrary{shadings}
\usetikzlibrary{shapes}
\usetikzlibrary{fadings}
\usetikzlibrary{bayesnet}
\usetikzlibrary{matrix}
\usetikzlibrary{plotmarks}
\usetikzlibrary{intersections}
\usetikzlibrary{pgfplots.fillbetween}
\pgfplotsset{compat=1.14}
\usepackage{siunitx}

\usepackage{colortbl}
\definecolor{mediumgray}{gray}{0.7}
\definecolor{lightgray}{gray}{0.85}
\definecolor{lightlightgray}{gray}{0.9}
\definecolor{C1}{HTML}{1F77B4}
\definecolor{C2}{HTML}{FF7F0E}
\definecolor{C3}{HTML}{2CA02C}
\definecolor{C4}{HTML}{D62728}
\definecolor{C5}{HTML}{9467BD}
\colorlet{C1light}{C1!70!white}
\colorlet{C2light}{C2!70!white}
\colorlet{C3light}{C3!70!white}
\colorlet{C4light}{C4!70!white}
\colorlet{C5light}{C5!70!white}
\colorlet{C1lighter}{C1!50!white}
\colorlet{C2lighter}{C2!50!white}
\colorlet{C3lighter}{C3!50!white}
\colorlet{C4lighter}{C4!50!white}
\colorlet{C5lighter}{C5!50!white}
\colorlet{C1vlight}{C1!20!white}
\colorlet{C2vlight}{C2!20!white}
\colorlet{C3vlight}{C3!20!white}
\colorlet{C4vlight}{C4!20!white}
\colorlet{C5vlight}{C5!20!white}
\colorlet{linkcolor}{violet}
\colorlet{citecolor}{RedOrange}  %
\colorlet{urlcolor}{Aquamarine}

\usepackage{amsmath}
\usepackage{amsfonts}       %
\usepackage{mathtools}
\usepackage{xspace}
\usepackage{bold-extra}
\usepackage{setspace}

\usepackage{paralist}
\usepackage{amsthm}
\usepackage{adjustbox}

\usepackage{placeins}

\usepackage[
    natbib,
    backend=biber,
    style=authoryear-comp,
    backref=false,
    maxbibnames=99,
    giveninits=false,
    maxcitenames=2,
    doi=false,
    isbn=false,
    url=false,
    uniquename=false,
    sortcites=false,
    dashed=false,
]{biblatex}
\AtEveryBibitem{%
	\clearlist{address}
	\clearfield{pages}
	\clearfield{volume}
	\clearfield{date}
	\clearfield{eprint}
	\clearfield{isbn}
	\clearfield{issn}
	\clearlist{location}
	\clearfield{month}
	\clearfield{series}

	\ifentrytype{book}{}{%
		\clearlist{publisher}
		\clearname{editor}
	}
}
\usepackage{caption}
\DeclareCaptionFont{textsc}{\bfseries\selectfont}
\usepackage{hyperref}
    \newcommand\myshade{85}
    \hypersetup{
      linkcolor  = linkcolor!\myshade!black,
      citecolor  = citecolor!\myshade!black,
      urlcolor   = urlcolor!\myshade!black,
      colorlinks = true,
    }

\usepackage[capitalize]{cleveref}
    \Crefname{table}{Tab.}{Tabs.}
    \Crefname{appendix}{App.}{Apps.}
    \Crefname{section}{Sec.}{Secs.}

\makeatletter
\newcommand*\iftodonotes{\if@todonotes@disabled\expandafter\@secondoftwo\else\expandafter\@firstoftwo\fi}  %
\makeatother

\definecolor{dandelion}{HTML}{FFD464}

\usepackage{tcolorbox}
\usepackage{contour}
\usepackage{array}
    \newcolumntype{C}{>{$}c<{$}}
    \newcolumntype{L}{>{$}l<{$}}
    \newcolumntype{R}{>{$}r<{$}}

\makeatletter
\renewcommand{\paragraph}{%
  \@startsection{paragraph}{4}%
  {\z@}{.25ex \@plus 0.5ex \@minus .2ex}{-1em}%
  {\normalfont\normalsize\bfseries}%
}
\makeatother

\newcounter{remark}%
\newcommand{\remark}[1]{\refstepcounter{remark} \textbf{Remark~\theremark.}\,\, \emph{#1}}

\newcommand{\transpose}{^\mathrm{\textsf{\tiny T}}}

\newcount\colveccount
\newcommand*\colvec[1]{
	\global\colveccount#1
	\begin{bmatrix}
		\colvecnext
	}
	\def\colvecnext#1{
		#1
		\global\advance\colveccount-1
		\ifnum\colveccount>0
		\\
		\expandafter\colvecnext
		\else
	\end{bmatrix}
	\fi
}

\newcommand{\kron}{\otimes}

\newcommand{\inv}{^{-1}}

\renewcommand{\i}{\mathrm{i}}
\newcommand{\e}{\mathrm{e}}

\newcommand{\Reals}{\mathbb{R}}

\makeatletter
\providecommand*{\calcd}%
   {\@ifnextchar^{\DIfF}{\DIfF^{}}}
\def\DIfF^#1{%
   \mathop{\mathrm{\mathstrut d}}%
      \nolimits^{#1}\gobblespace
}
\def\gobblespace{%
   \futurelet\diffarg\opspace}
\def\opspace{%
   \let\DiffSpace\!%
   \ifx\diffarg(%
      \let\DiffSpace\relax
   \else
      \ifx\diffarg\[%
         \let\DiffSpace\relax
      \else
         \ifx\diffarg\{%
            \let\DiffSpace\relax
         \fi\fi\fi\DiffSpace}
\makeatother

\renewcommand{\d}[1]{\ensuremath{\calcd #1}}

\newcommand\B[1]{{\mathbf{#1}}}
\newcommand\C[1]{{\mathcal{#1}}}

\renewcommand{\vec}{\pmb}

\newcommand{\vb}{\ensuremath{\mathbf{b}}}
\newcommand{\vf}{\ensuremath{\mathbf{f}}}
\newcommand{\vg}{\ensuremath{\mathbf{g}}}
\newcommand{\vk}{\ensuremath{\mathbf{k}}}

\newcommand{\vm}{\ensuremath{\mathbf{m}}}
\newcommand{\vr}{\ensuremath{\mathbf{r}}}

\newcommand{\vp}{\ensuremath{\mathbf{p}}}
\newcommand{\vw}{\ensuremath{\mathbf{w}}}
\newcommand{\vx}{\ensuremath{\mathbf{x}}}
\newcommand{\vxp}{\ensuremath{\mathbf{x}^\prime}}
\newcommand{\vxstar}{\ensuremath{\vx^\ast}}
\newcommand{\vy}{\ensuremath{\mathbf{y}}}

\newcommand{\vS}{\ensuremath{\mathbf{S}}}
\newcommand{\vW}{\ensuremath{\mathbf{W}}}
\newcommand{\vQ}{\ensuremath{\mathbf{Q}}}

\newcommand{\vD}{\ensuremath{\mathbf{D}}}
\newcommand{\vM}{\ensuremath{\mathbf{M}}}

\newcommand{\vtheta}{\boldsymbol{\theta}\xspace}

\newcommand{\vmu}{\boldsymbol{\mu}}

\newcommand{\vsigma}{\boldsymbol{\sigma}}
\newcommand{\vSigma}{\boldsymbol{\Sigma}}
\newcommand{\vLambda}{\boldsymbol{\Lambda}}
\newcommand{\vL}{\boldsymbol{L}}

\newcommand{\vI}{\boldsymbol{I}}
\newcommand{\vK}{\boldsymbol{K}}

\DeclareMathOperator{\Exp}{\mathbb{E}}

\DeclarePairedDelimiterX{\infdivx}[2]{(}{)}{%
	#1\;\delimsize\|\;#2%
}

\DeclareMathOperator{\Cov}{Cov}
\renewcommand{\Exp}{\mathbb{E}}

\newcommand{\given}{\ensuremath{\vert}}

\newcommand{\expect}[1]{\Exp\left[#1\right]}

\newcommand{\X}{\mathbb{X}}
\newcommand{\N}{\mathcal{N}}

\newcommand{\g}{\,|\,}

\newcommand{\T}{^{\intercal}}

\newcommand{\q}{\quad}

\renewcommand{\t}{\theta}

\renewcommand{\vec}{\boldsymbol}

\renewcommand{\O}{\mathcal{O}}

\newcommand{\f}{\vec{f}}
\newcommand{\y}{\vec{y}}
\newcommand{\x}{\vec{x}}

\newcommand{\wrt}{w.r.t\xspace}

\usepackage{nicefrac}

\newcommand{\D}{\mathcal{D}}
\newcommand{\R}{\mathcal{R}}
\newcommand{\reals}{\mathbb{R}}
\newcommand{\order}[1]{\ensuremath{\mathcal{O}\left(#1\right)}}

\newcommand{\bnn}{\textsc{bnn}\xspace}

\newcommand{\ggn}{\textsc{ggn}\xspace}

\newcommand{\gp}{\textsc{gp}\xspace}
\newcommand{\glm}{\textsc{glm}\xspace}

\newcommand{\dnntogp}{\textsc{dnn2gp}\xspace}
\newcommand{\GGN}{\textsc{ggn}\xspace}
\newcommand{\kfac}{\textsc{kfac}\xspace}
\newcommand{\dig}{\textsc{diag}\xspace}

\newcommand{\flin}{\ensuremath{\vf_\text{lin}^{\vtheta^\ast}}}

\newcommand{\likelihood}{\ensuremath{p(\D \given \vtheta)}\xspace}
\newcommand{\prior}{\ensuremath{p(\vtheta)}\xspace}
\newcommand{\posterior}{\ensuremath{p(\vtheta\given\D)}\xspace}
\newcommand{\approxposterior}{\ensuremath{q(\vtheta)}\xspace}
\newcommand{\onelikelihood}{\ensuremath{p(\vy\given\vf(\vx, \vtheta))}\xspace}
\newcommand{\onelikelihoodshort}{\ensuremath{p(\vy\given\vf)}\xspace}

\newcommand{\jac}{\ensuremath{\mathcal{J}_{\vtheta}(\vx)}\xspace}
\newcommand{\jacast}{\ensuremath{\mathcal{J}_{\vtheta^\ast}(\vx)}\xspace}
\newcommand{\jacastshort}{\ensuremath{\mathcal{J}_{\vtheta^\ast}}\xspace}
\newcommand{\jacshort}{\ensuremath{\mathcal{J}}\xspace}
\newcommand{\hessshort}{\ensuremath{\mathcal{H}}\xspace}
\newcommand{\hess}{\ensuremath{\mathcal{H}_{\vtheta}(\vx)}\xspace}
\newcommand{\network}{\ensuremath{\vf(\vx, \vtheta)}\xspace}
\newcommand{\networkast}{\ensuremath{\vf(\vx, \vtheta^\ast)}\xspace}
\newcommand{\networklin}{\ensuremath{\vf_\text{lin}^{\vtheta^\ast}\!(\vx, \vtheta)}\xspace}

\newcommand{\networklinn}{\ensuremath{\vf_\text{lin}^{\vtheta^\ast}\!(\vx_n, \vtheta)}\xspace}

\newcommand{\thetamap}{\ensuremath{\vtheta_\text{MAP}}\xspace}
\newcommand{\thetastar}{\ensuremath{\vtheta^\ast}}

\newcommand{\wmap}{\ensuremath{\vw_\text{MAP}}\xspace}
\newcommand{\bmap}{\ensuremath{\vb_\text{MAP}}\xspace}

\newcommand{\pms}[1]{\ensuremath{{\scriptstyle\pm #1}}}

\newcommand*\circled[2]{\tikz[baseline=(char.base)]{\node[shape=circle,draw, #1] (char) {#2};}}

\begin{document}

\addtocontents{toc}{\protect\setcounter{tocdepth}{-1}}

\newcommand{\ourtitle}{Improving predictions of Bayesian neural nets via local linearization}
\twocolumn[
\aistatstitle{\ourtitle}

\aistatsauthor{ Alexander Immer${}^\ast$ \And Maciej Korzepa \And  Matthias Bauer${}^\ast$}

\aistatsaddress{ Department of Computer Science \\ ETH Zurich, Switzerland \\\vspace{-0.8em} \\ Max Planck ETH Center \\ for Learning Systems \And  Technical University of Denmark\\ Copenhagen, Denmark \And DeepMind \\ London, UK }
]

\newtheorem{theorem}{Theorem}

\begin{abstract}
The generalized Gauss-Newton (\ggn) approximation is often used to make practical Bayesian deep learning approaches scalable by replacing a second order derivative with a product of first order derivatives. In this paper we argue that the \ggn approximation should be understood as a local linearization of the underlying Bayesian neural network (\bnn), which turns the \bnn into a generalized linear model (\glm). Because we use this linearized model for posterior inference, we should also predict using this modified model instead of the original one. We refer to this modified predictive as ``\glm predictive'' and show that it effectively resolves common underfitting problems of the Laplace approximation. It extends previous results in this vein to general likelihoods and has an equivalent Gaussian process formulation, which enables alternative inference schemes for {\bnn}s in function space. We demonstrate the effectiveness of our approach on several standard classification datasets and on out-of-distribution detection.
We provide an implementation at \url{https://github.com/AlexImmer/BNN-predictions}.
\vspace{-0.5em}
\end{abstract}

\section{Introduction}

Inference in Bayesian neural networks ({\bnn}s) %
usually requires posterior approximations due to intractable integrals and high computational cost.
Given such an approximate posterior of the parameters, we can make predictions at new locations by combining the posterior with the original Bayesian neural network likelihood.

\begin{figure}[tb]
    \centering
    \tcbox[top=0pt,left=0pt,right=0pt, bottom=0pt, colback=white, colframe=C1!75!black]{  \begin{tikzpicture}%
\node[align=left] (bnn) {$p_\text{\bnn}(\vy\given\vx, \mathcal{D})=$ \\ $\,\,\int\! q(\vtheta) p(\vy | \vf(\vx, \vtheta))\calcd{\vtheta}$};
  \node[above=0 of bnn] (bnn_title) {\bnn predictive (\cref{eq:bnn_predictive})};
\node[right=1. of bnn, align=left] (bglm) {$p_\text{\glm}(\vy\given\vx,\mathcal{D})=$ \\ $\,\,\int\! q(\vtheta) p(\vy | \textcolor{C1}{\flin}\!(\vx, \vtheta))\calcd{\vtheta}$};
  \node[above=0 of bglm] (bglm_title) {\glm predictive (\cref{eq:glm_predictive})};
  \draw[->, >={stealth'},  line width=0.7pt,shorten <=0.75pt,shorten >=0.75pt, C1] (bnn) -- (bglm) node[pos=0.5, above] (ggn) {\small \GGN};
  \node[below =0.6  of ggn, C1] (flin) {\footnotesize $\networklin = \networkast + \left.\nabla_{\vtheta}\vf(\vx, \vtheta)\right\vert_{\vtheta=\thetastar}\transpose(\vtheta - \vtheta^\ast)$};
  \draw(flin) edge[out=65,in=270,->, >={stealth'}, shorten <=-1pt, C1] (ggn);
  
  \end{tikzpicture}}
    \vskip-0.25em
    \caption{The generalized Gauss Newton approximation (\GGN) turns a Bayesian neural network (\bnn) into a generalized linear model (\glm) with same likelihood distribution, but network function $\vf(\vx, \vtheta)$ linearized around $\vtheta^\ast$. %
    When using \ggn, we should also use the \glm in the predictive. %
    \approxposterior is an approximate posterior and \thetastar{} is found by MAP estimation, \protect\cref{eq:map_objective}.}
    \label{fig:bnn_glm_predictive}
    \vspace{-0.7em}
\end{figure}

One common posterior approximation is the Laplace approximation \citep{mackay1992bayesian}, which has recently seen a revival for modern neural networks \citep{khan2019approximate, ritter2018scalable}. It approximates the posterior by a Gaussian around its maximum and has become computationally feasible through further approximations, most of which build on the \emph{generalized Gauss-Newton approximation} (\ggn; \citet{martens2015optimizing}). The \ggn replaces an expensive second order derivative by a product of first order derivatives, and is often jointly applied with approximate inference in {\bnn}s using the Laplace approximation \citep{ritter2018scalable,foresee1997gauss,foong2019between} or variational approximations \citep{khan2018fast,zhang2018noisy}.

Recently, \citet{foong2019between} showed empirically that predictions using a ``linearized Laplace'' predictive distribution in this setting %
can match or outperform other approximate inference approaches, such as mean field variational inference (MFVI) in the \emph{original} \bnn model \citep{blundell2015weight} and provide better ``in-between'' uncertainties for regression. %
Here we explain that their approach relies on an implicit change in probabilistic model due to the \ggn approximation.

More specifically, we argue that the \ggn approximation should be considered separately from approximate posterior inference:
(1) the \ggn approximation locally linearizes the underlying probabilistic model in its parameters %
and gives rise to a generalized linear model (\glm); (2) approximate inference %
such as through the Laplace approximation
enables posterior inference in this linearized \glm.
Because we have done inference in a \emph{modified} probabilistic model (the \glm), we should also predict with this modified model. We call the resulting predictive that uses locally linearized neural network features the ``\glm predictive'' in contrast to the normally used ``\bnn predictive'' that uses the original \bnn features in the likelihood, see \cref{fig:bnn_glm_predictive}.

Our approach generalizes previous results by \citet{khan2019approximate} and \citet{foong2019between} to non-Gaussian likelihoods. It explains why the \glm predictive works well compared to the \bnn predictive, which can show underfitting for Laplace posteriors \citep{lawrence2001phd}, especially when combined with the \ggn approximation \citep{ritter2018scalable}. %
We demonstrate that our proposed \glm predictive resolves these underfitting problems and consistently outperforms the \bnn predictive by a wide margin on several datasets; it is on par or better than the neural network MAP or MFVI.
Further, the \glm in weight space can be viewed as an equivalent Gaussian process (\gp) in function space, which enables complementary inference approximations. Finally, we show that the proposed \glm predictive can be successfully used for out-of-distribution detection.

\section{Background}
\label{sec:background}
In this paper we consider supervised learning tasks with inputs $\vx_n \in \reals^D$ and outputs $\vy_n \in \reals^C$ (e.g. regression) or $\vy_n \in \{0, 1\}^C$ (e.g. classification), $\D=\{(\vx_n, \vy_n)\}_{n=1}^N$. %
We introduce features $\vf(\vx, \vtheta)$ with parameters $\vtheta\in \reals^P$ and use a likelihood function \likelihood to map them to the outputs $\vy$ using an inverse link function $\vg\inv$, $\expect{\vy} = \vg\inv(\vf(\vx, \vtheta))$, such as the sigmoid or softmax:
\begin{align}
    \likelihood = \textstyle\prod_{n=1}^N p(\vy_n \given \vf(\vx_n, \vtheta)),\label{eq:likelihood}
\end{align}
In Bayesian deep learning (BDL) we impose a prior $\prior$ on the likelihood parameters and aim to compute their posterior given the data, $\posterior$; a typical choice is to assume a Gaussian prior $\prior=\mathcal{N}\left(\vm_0, \vS_0\right)$. Given a parameter posterior $\posterior$, we make probabilistic predictions for new inputs $\vx^{\ast}$ using the posterior predictive
\begin{align}
    p(\vy^\ast\given\vx^\ast, \D) = \Exp_{p(\vtheta\given\D)}[p\left(\vy^\ast\given\vf(\vx^\ast, \vtheta)\right)].
\end{align}
Exact posterior inference %
requires computation of a high-dimensional integral, the \emph{model evidence} or \emph{marginal likelihood} $p(\vy\given\vx) = \int \likelihood \prior \calcd{\vtheta}$, and is often infeasible. We therefore have to resort to approximate posterior inference techniques, such as mean field variational inference or the Laplace approximation, that approximate  $\approxposterior \approx \posterior$.

\paragraph{Mean-field VI.} Popular in recent years, \emph{mean-field variational inference} (MFVI) approximates the posterior $\posterior$ by a factorized variational distribution $\approxposterior$ optimized using an evidence lower bound (ELBO) to the marginal likelihood \citep{blundell2015weight}.

\paragraph{MAP.} Many practical approaches compute the \emph{maximum a posteriori} (MAP) solution $\thetamap=\arg\max_{\vtheta}\ell(\vtheta, \D)$ and return a point estimate $q(\vtheta) = \delta(\vtheta-\thetamap)$ or a distribution $\approxposterior$ around \thetamap; here $\ell(\vtheta, \D)$ denotes the log joint distribution
\begin{equation}
    \ell(\vtheta, \D) = \textstyle\sum_{n=1}^N \log p(\vy_n \given \vf(\vx_n,\vtheta)) + \log p(\vtheta), \label{eq:map_objective}
\end{equation}

\paragraph{Laplace.} The \emph{Laplace approximation} \citep{mackay1992bayesian} approximates the posterior by a Gaussian $q(\vtheta) = \mathcal{N}(\thetamap, \Sigma)$ around the mode $\thetamap$ with covariance $\Sigma$ given by the Hessian of the posterior
\begin{align}
    \Sigma = -\left[\left.\nabla^2_{\vtheta\vtheta} \log p(\vtheta|\mathcal{D})\right\vert_{\vtheta=\thetamap}\right]^{-1}.
\end{align}
To compute $\Sigma$, we need to compute the Hessian of \cref{eq:map_objective}; the prior terms are usually trivial, such that we focus on the log likelihood. We express the involved Jacobian and Hessian of the log likelihood \emph{per data point} through the \emph{Jacobian} $\jacshort \in \reals^{C \times P}$ and \emph{Hessian} $\hessshort \in \reals^{C \times P \times P}$ of the feature extractor $\vf(\vx, \vtheta)$, $\left[\jac\right]_{ci} = \frac{\partial f_c(\vx, \vtheta)}{\partial \theta_i}$ and $\left[\hess\right]_{cij} = \frac{\partial^2 f_c(\vx, \vtheta)}{\partial \theta_i \partial \theta_j}$, respectively:
\begin{align}
    \nabla_{\vtheta} \log\onelikelihood  & = \jac\transpose \vr(\vy; \vf) \label{eq:likelihood_jacobian}\\
    \begin{split}\nabla^2_{\vtheta\vtheta}  \log\onelikelihood  & =  \hess\transpose\vr(\vy; \vf) \\
    & \qquad - \jac \transpose \vLambda(\vy; \vf)\jac. \end{split} \label{eq:likelihood_hessian}
\end{align}
We can interpret $\vr(\vy; \vf) = \nabla_{\vf}\log\onelikelihoodshort$ as a residual and $\vLambda(\vy; \vf) = - \nabla^2_{\vf\vf}\log\onelikelihoodshort$ as per-input noise.
\begin{figure*}[tb]
    \centering
    \renewcommand{\edge}[3][]{ %
	\foreach \x in {#2} { %
		\foreach \y in {#3} { %
			\path (\x) edge [->, >={stealth'},  line width=0.7pt,
			#1] (\y) ;%
		} ;
	} ;
}

\tikzstyle{every node}=[font=\normalsize]
  \begin{tikzpicture}
  \node[align=center] (bnn) {$p_\text{\bnn}(\D, \vtheta)\propto$ \\ $p(\vtheta)\prod_i p(\vy_n | \vf(\vx_n, \vtheta))$};
  \node[above=0 of bnn] (bnn_title) {\large \bnn\vphantom{y}};
  \node[right=2 of bnn, align=center] (bglm) {$p_\text{\glm}(\D, \vtheta)\propto$ \\ $p(\vtheta)\prod_n p(\vy_n | \flin(\vx_n, \vtheta))$};
  \node[above=0 of bglm] (bglm_title) {Bayesian \glm};
  \node[right=2.75 of bglm, align=center] (ggp) {$p_{\gp}(\D, \vf)\propto$\\$p_\text{GP}(\vf) \prod_n p(\vy_n | \vf_n)$};
  \node at (bglm_title-|ggp) {\large \gp};
  \node[below right =0.3 and 0.5 of bglm.south west, C4, anchor=north east] (flin) {\footnotesize $\networklin = \networkast + \jacast(\vtheta - \vtheta^\ast)$};
  \draw(flin) edge[out=65,in=270,->, >={stealth'}, shorten <=-1pt, C4] ($(bnn)!0.5!(bglm)$);
  \draw[->, >={stealth'},  line width=1.25pt,shorten <=0.75pt,shorten >=0.75pt, C4] (bnn) -- (bglm) node[pos=0.5, above, text width=2cm, align=center] (ggn) {\GGN} node[pos=0.5, below] {\contour{white}{\footnotesize(\cref{sec:ggn_bnn_glm})}};
  
  \draw[<->,>={stealth'}, dashed, line width=1.25pt, shorten <=0.75pt, shorten >=0.75pt, C3] (bglm) -- (ggp) 
    node[pos=0.5, above] {\footnotesize\shortstack{weight/function \\space view}}%
    node[pos=0.5, below] {\footnotesize(\cref{sec:gp_predictive})};
  \node[below=0.75 of bglm] (bglm_approx) {\footnotesize $q(\vtheta) = \mathcal{N}(\vmu, \vSigma)$};
  \node[below=0.75 of ggp] (ggp_approx) {\footnotesize $q(\vf) = \mathcal{GP}(\vm_q(\vx), \vk_q(\vx, \vx'))$};
  \node[align=center] at ($(bglm.south)!0.5!(ggp_approx.north)$) {\color{C1}\footnotesize Gaussian posterior approximation (\cref{sec:glm_approximate_inference})};
  \draw[->, >={stealth'},  line width=1.25pt,shorten <=0.75pt,shorten >=0.75pt, C1] ($(bglm.south)-(0.25,0)$) -- ($(bglm_approx.north)-(0.25,0)$);
  \draw[->, >={stealth'},  line width=1.25pt,shorten <=0.75pt,shorten >=0.75pt, C1] ($(ggp.south)+(0.25,0)$) -- ($(ggp_approx.north)+(0.25,0)$);
  \end{tikzpicture}
    \vskip-0.75em
    \caption{The generalized Gauss Newton approximation (\GGN) turns a Bayesian neural network (\bnn) into a generalized linear model (\glm) with same prior and likelihood distribution, but network function $\vf(\vx_n, \vtheta)$ linearized around $\vtheta^\ast$
    (\protect\tikz[baseline=-0.5ex,inner sep=0pt]{\protect\draw[->, >={stealth'},line width=1pt,shorten <=0.75pt,shorten >=0.75pt, C4] (0,0) -- ++(0.5,0);}).
    The \glm is equivalent to a Gaussian process (\gp) (\protect\tikz[baseline=-0.5ex,inner sep=0pt]{\protect\draw[<->, dashed, >={stealth'},line width=1pt,shorten <=0.75pt,shorten >=0.75pt, C3] (0,0) -- ++(0.75,0);}).
    Inference is made tractable with a Gaussian posterior approximation ( \protect\tikz[baseline=-0.5ex,inner sep=0pt]{\protect\draw[->, >={stealth'},line width=1pt,shorten <=0.75pt,shorten >=0.75pt, C1] (0,0.2) -- ++(0,-0.4);} ), and we predict using the \glm predictive (\cref{eq:glm_predictive}).
    }
    \label{fig:bnn_bglm}
\end{figure*}

\paragraph{\ggn.}
The network Hessian $\hess$ in \cref{eq:likelihood_hessian} is infeasible to compute in practice, %
such that many approaches employ the \emph{generalized Gauss-Newton} (\ggn) approximation, which drops this term \citep{schraudolph2002fast, martens2014new} and approximates \cref{eq:likelihood_hessian} as:
\begin{align}
    \nabla^2_{\vtheta\vtheta}  \log\onelikelihood  & \approx - \jac \transpose \vLambda(\vy; \vf)\jac. \label{eq:likelihood_hessian_ggn}
\end{align}
This \ggn approximation to the Hessian is also guaranteed to be positive semi-definite, whereas the original Hessian \cref{eq:likelihood_hessian} is not.
The \ggn is often further approximated, and in
this paper, we consider the most common cases \citep{ritter2018scalable, zhang2018noisy},  diagonal and Kronecker-factored (\kfac) approximations~\citep{martens2015optimizing, botev2017practical}. %
\kfac approximations are block-diagonal to enable efficient storage and computation of inverses and decompositions while maintaining expressivity compared to a diagonal approximation.
Each block corresponding to a parameter group, e.g., a neural network layer, is Kronecker factored; the \ggn of the $l$-th parameter group is approximated as
\begin{equation}
\label{eq:kfac_ggn}
  \left[{\textstyle \sum_{n=1}^N}\jacshort_{\vtheta}(\vx_n) \transpose \vLambda(\vy_n; \vf_n)\jacshort_{\vtheta}(\vx_n)\right]_l \approx \vQ_l \kron \vW_l,
\end{equation}
where $\vQ_l$ is the uncentered covariance of the activations and $\vW_l$ is computed recursively~\citep{botev2017practical}.
Therefore, $\vQ_l$ is quadratic in the size of the input and $\vW_l$ in the output of the layer, and both are positive semidefinite.
Inversion of the Kronecker approximation is cheap because we only need to invert its factors individually.
The Kronecker approximation can be combined with the prior exactly~\citep{grosse2016kronecker} or using \emph{dampening}~\citep{ritter2018scalable}.
We use the exact version, see \cref{app:kronecker} for a discussion.

\paragraph{Posterior predictive.} Regardless of the posterior approximation, we usually obtain a predictive distribution by integrating the approximate posterior \approxposterior against the model likelihood $\likelihood$:
\begin{tcolorbox}[enhanced,colback=white,%
    colframe=C1!75!black, attach boxed title to top right={yshift=-\tcboxedtitleheight/2, xshift=-.75cm}, title=\bnn predictive, coltitle=C1!75!black, boxed title style={size=small,colback=white,opacityback=1, opacityframe=0}, size=title, enlarge top initially by=-\tcboxedtitleheight/2, left=-5pt,  after skip=1.5ex plus 0.5ex]
 \vskip-1.em
 \begin{equation}
\begin{aligned}
    &p_{\bnn}(\vy \given \vx, \D) = \Exp_{\approxposterior}\left[p(\vy \given \vf(\vx, \vtheta))\right]\\
    & \qquad\approx \tfrac{1}{S} \textstyle\sum_s p(\vy \given \vf(\vx, \vtheta_s)), \qquad \vtheta_s \sim \approxposterior
\end{aligned} \label{eq:bnn_predictive}
 \end{equation}
\end{tcolorbox}
where we have approximated the (intractable) expectation by Monte Carlo sampling. To distinguish this predictive from our proposed method, we refer to \cref{eq:bnn_predictive} as \bnn predictive. Typically, the \bnn predictive distribution is non-Gaussian, because the likelihood can be non-Gaussian and/or $\vf$ depends non-linearly on $\vtheta_s$.

\section{Methods}
\label{sec:methods}
Here, we discuss the effects of the \ggn approximation in more detail (\cref{sec:ggn_bnn_glm}) and introduce our main contributions, the \glm predictive (\cref{sec:glm_predictive}) and its \gp counterpart (\cref{sec:gp_predictive}); see \cref{fig:bnn_bglm} for an overview.

\begin{figure*}[tb]
    \centering
    \pgfdeclarelayer{background}
	\pgfdeclarelayer{foreground}
	\pgfsetlayers{background,main,foreground}   %
\begin{tikzpicture}
	\node (a){\begin{tikzpicture}
	\node (data) at (0,0) {\mbox{\adjincludegraphics[scale=0.8, Clip={0.001\width} {.75\height} {.775\width} {0.08\height}, clip]{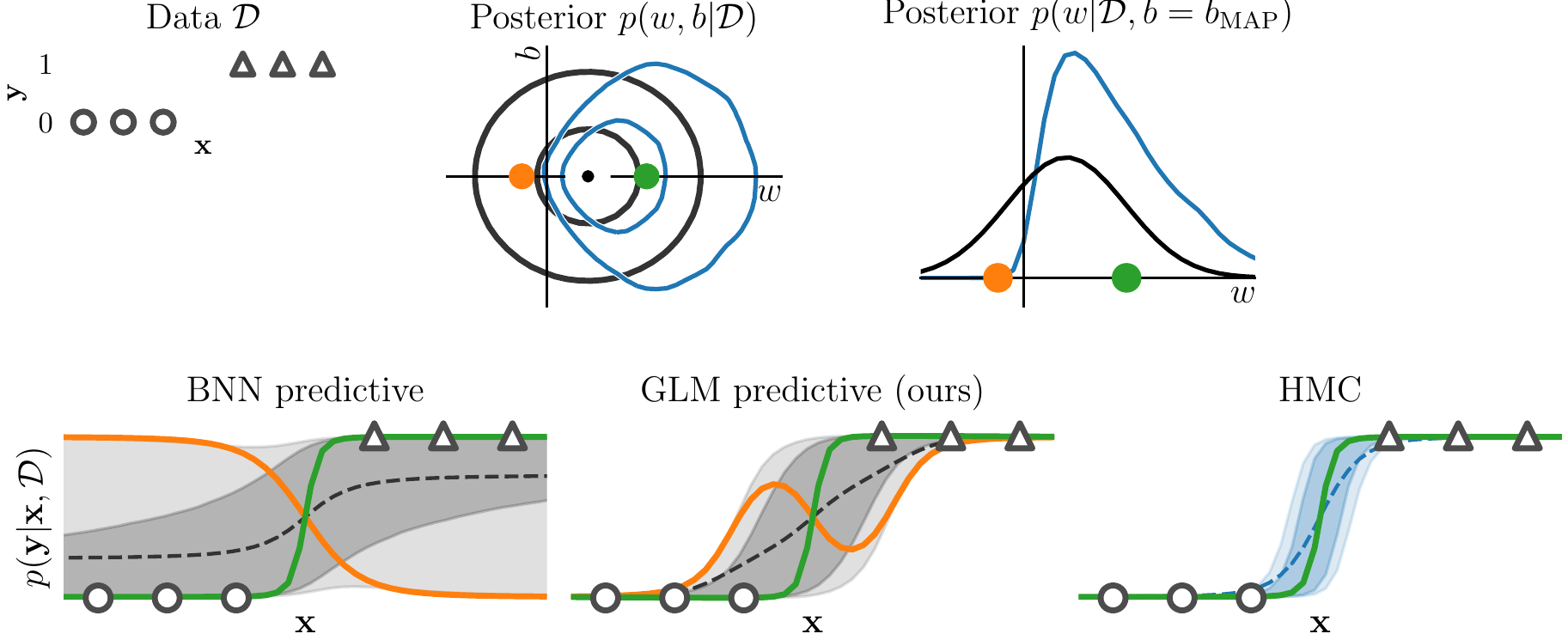}}};
	\node[below=0.15 of data.south west, align=left, anchor=north west] (model) {{Model} \\[1.5ex] 
		$\begin{aligned}
			f(\vx; w, b) &= 5\tanh(w\vx+b)\\
			p(\vy=1|\vf) & =\sigma(\vf)\\
			p(w) = p(b)&= \mathcal{N}(0,1)
		\end{aligned}$};
	\begin{pgfonlayer}{background}
		\node[rectangle, line width=1pt, draw=white, fit=(data) (model) , inner sep=3mm, rounded corners=5](1) {};
		\node[left=0.5 of 1.north west, anchor=west, fill=white] { \circled{inner sep=1pt}{\textbf{1}} \textbf{Data $\mathcal{D}$ and model}};
	\end{pgfonlayer}
	\end{tikzpicture}};

	\node[right=0.25 of a.north east, anchor=north west](b){
	\begin{tikzpicture}
		\node (posterior1){\mbox{\adjincludegraphics[scale=.8, Clip={0.27\width} {.48\height} {0.5\width} {0.06\height}, clip]{figures/toy_classification_v4.pdf}}};
		\node[right=0.1 of posterior1] (posterior2){\mbox{\adjincludegraphics[scale=.8, Clip={0.57\width} {.48\height} {0.2\width} {0.06\height}, clip]{figures/toy_classification_v4.pdf}}};
		\node[above=-0.2 of posterior1.north, anchor=south](text2){$p(w, b|\mathcal{D})$};
		\node[above=-0.2 of posterior2.north, anchor=south](text3){$p(w|\mathcal{D}, b=b_\mathrm{MAP})$};
		\node[below=0. of posterior1.south west, anchor=north west] (postlegend) {\tikz[baseline=-0.5ex,inner sep=2pt]{\draw[C1, line width=1] (0,0) -- (0.75,0);} \footnotesize true (HMC) \hspace{1em}
			\tikz[baseline=-0.5ex,inner sep=2pt]{\draw[black, line width=1] (0,0) -- (0.75,0);} \footnotesize approximate (Laplace-GGN)};
		\begin{pgfonlayer}{background}
			\node[rectangle, line width=1pt, draw=white, fit=(posterior1) (postlegend) (text3), inner sep=2mm, rounded corners=5](2) {};
			\node[left=0.5 of 2.north west, anchor=west, fill=white] { \circled{inner sep=1pt}{\textbf{2}} \textbf{Posterior inference}};
		\end{pgfonlayer}
	\end{tikzpicture}	
	};
	\node[below=-0.2 of a.south west, anchor=north west](c){
	\begin{tikzpicture}

	\node(predictives){\mbox{\adjincludegraphics[scale=.775, Clip={0.01\width} {0.01\height} {0.\width} {0.575\height}, clip]{figures/toy_classification_v4.pdf}}};
	
	\begin{pgfonlayer}{background}

	\node[rectangle, line width=0pt, draw=white, fit=(predictives), inner sep=0.1mm, inner ysep=1mm, rounded corners=5](3) {};
	\node[left=0.5 of 3.north west, anchor=west, fill=white] { \circled{inner sep=1pt}{\textbf{3}} \textbf{Posterior predictive distribution}};
	\end{pgfonlayer}
\end{tikzpicture}};
\end{tikzpicture}
    \vskip-.85em
    \caption[]{
    The \bnn predictive underfits because some samples can give extremely wrong predictions (an example shown in orange, \protect\tikz[baseline=-0.5ex,inner sep=2pt]{
        \protect\draw[line width=1.5pt, C2] (0,0) -- node[scale=1.1,pos=0.5] {\protect\pgfuseplotmark{*}} ++(0.75,0);
        }). 
    The \glm predictive corrects this.\\
    \circled{inner sep=1pt}{2} Laplace-\ggn posterior
    (\protect\tikz[baseline=-0.5ex,inner sep=0pt]{\protect\draw[line width=1pt, black] (0,0) ellipse (0.2cm and 0.1cm);\protect\draw[line width=1pt, black] (0,0) ellipse (0.35cm and 0.175cm);})
    vs. the true posterior 
    (\!\protect\tikz[baseline=-0.5ex]{\protect\draw[rounded corners=4, line width=1pt,C1] (0,0)--(0.3,-0.15)--(0.6,-0.2)--(0.85,0)--(0.6,0.2)--(0.3,0.15)--cycle; \protect\draw[rounded corners=3, line width=1pt, C1, scale=0.75] (0.1,0)--(0.5,-0.15)--(0.75,0)--(0.5,0.15)--cycle;}\!)
    through $10^5$ HMC samples: the Laplace-\ggn is symmetric and extends beyond the true, skewed posterior with same MAP. We highlight two posterior samples, one where both distributions have mass
    (\protect\tikz[baseline=-0.5ex,inner sep=2pt]{\protect\node[C3, scale=1.5] {\pgfuseplotmark{*}};})
    and another where only the Laplace-\ggn has mass
    (\protect\tikz[baseline=-0.5ex,inner sep=2pt]{\protect\node[C2, scale=1.5] {\pgfuseplotmark{*}};}).\\
    \circled{inner sep=1pt}{3} Posterior predictives $p(\vy\given\vx, \mathcal{D})$. The \bnn and \glm predictive both use the same Laplace-\ggn posterior; while the proposed \glm predictive closely resembles HMC (using the true posterior), the \bnn predictive underfits. Underfitting is due to samples from the mismatched region of the posteriors 
    (\protect\tikz[baseline=-0.5ex,inner sep=2pt]{
        \protect\draw[line width=1.5pt, C2] (0,0) -- node[scale=1.1,pos=0.5] {\protect\pgfuseplotmark{*}} ++(0.75,0);
        }); while the \glm predictive reasonably extrapolates the behaviour around the MAP, the \bnn predictive behaves qualitatively different.\\
    \protect\tikz[baseline=-0.5ex,inner sep=0pt]{\protect\draw[line width=1.pt, C1, dashed] (0,0.05) -- ++(0.5,0);\protect\draw[line width=1.pt, black, dashed] (0,-0.05) -- ++(0.5,0);}
    predictive means;
    \protect\tikz[baseline=0ex,inner sep=0pt]{\protect\draw[C1vlight, fill] (0,-0.1) rectangle ++(0.5,0.4); \protect\draw[C1lighter, fill] (0,0.025) rectangle ++(0.5,0.15);\protect\draw[lightgray, fill] (-0.5,-0.1) rectangle ++(0.5,0.4); \protect\draw[mediumgray, fill] (-0.5,0.025) rectangle ++(0.5,0.15);\useasboundingbox (0,0) rectangle ++(0.5,0.15);}
    innermost $50\%$/$66\%$ of samples.
    }
    \label{fig:toy_classification}
\end{figure*}

\subsection{Generalized Gauss-Newton turns {\bnn}s into generalized linear models}
\label{sec:ggn_bnn_glm}
In \cref{sec:background} we introduced the \ggn as a positive semi-definite approximation to the Hessian by simply dropping the term $\hess\transpose \vr(\vy; \vf)$ in \cref{eq:likelihood_hessian_ggn}; in other words, we assume that $\hess\transpose \vr(\vy; \vf) = 0$. Two independently sufficient conditions are commonly used as justification ~\citep{bottou2018optimization}: (i) The residual vanishes for all data points, $\vr(\vy;\network) = 0 \, \forall(\vx,\vy)$, which is true if the network is a perfect predictor.
However, this is neither desired, as it indicates overfitting, nor is it realistic. (ii) The Hessian vanishes, $\hess = 0\, \forall \vx$, which is true for linear networks and can be enforced by linearizing the network. Hence, an alternative definition uses this second condition as a starting point and defines the \ggn through the linearization of the network \citep{Martens2011_rnn_hessian_free}.

In this work, we follow this alternative definition and motivate the \ggn approximation as a \emph{local linearization} of the network function $\network$,
\begin{align}
    \networklin = \networkast + \jacast(\vtheta - \vtheta^\ast),
    \label{eq:local_linearization}
\end{align}
at a parameter setting $\vtheta^\ast$ (\protect\tikz[baseline=-0.5ex,inner sep=0pt]{\protect\draw[->, >={stealth'},line width=1pt,shorten <=0.75pt,shorten >=0.75pt, C4] (0,0) -- ++(0.5,0);} in \cref{fig:bnn_bglm}).
This linearization reduces the \bnn to a Bayesian generalized linear model (\glm) with log joint distribution $\ell_{\glm}(\vtheta, \D)$%
\begin{align}
    \hspace*{-0.5em}\ell_{\glm}(\vtheta, \D) =\textstyle\sum_{n=1}^N \log p(\vy_n \given \networklinn) + \log p(\vtheta),\raisetag{2.5em}
    \label{eq:log_joint_glm}
\end{align}
where $\networklin$ is linear in the parameters $\vtheta$ but not in the inputs $\vx$.
In practice, we often choose the linearization point $\thetastar$ to be the MAP estimate found by optimization of \cref{eq:map_objective}. At $\thetastar$ the \ggn approximation to the Hessian of the linearized model, \cref{eq:log_joint_glm}, is identical to that of the full model, \cref{eq:map_objective}.

\remark{Applying the \ggn approximation to the likelihood Hessian turns the underlying probabilistic model locally from a \bnn into a \glm}.

\subsection{Approximate inference in the \glm}
\label{sec:glm_approximate_inference}

Previous works, e.g. \citet{ritter2018scalable,khan2019approximate}, apply the Laplace and the \ggn approximation jointly. We refer to the resulting posterior $q(\vtheta) = \mathcal{N}(\thetamap, \vSigma_{\ggn})$ as the ``Laplace-\ggn posterior'', where $\vSigma_{\ggn}$ denotes one of the \ggn approximations to the covariance introduced in \cref{sec:background} (full, diagonal, or \kfac). The full covariance case is given~by:
\begin{align}
  \label{eq:sigma_post}
   \hspace*{-0.5em} \vSigma_{\ggn}^{-1} & = \textstyle\sum_{n=1}^N\jacastshort(\vx_n) \transpose \vLambda(\vy_n; \vf_n)\jacastshort(\vx_n) + \vS_0^{-1} %
\end{align}
with prior covariance $\vS_0$. In our \glm setting, this corresponds to linearizing the original \bnn around $\thetastar=\thetamap$ and using the same Laplace-\ggn posterior. For large-scale experiments we use this posterior as it is simpler and computationally more feasible than the refinement we describe next. Note that our main contribution is to propose a different predictive (see \cref{sec:glm_predictive}), not a different posterior.

We can use the \glm perspective to \emph{refine} the posterior, because in practise we are only ever approximately able to find $\thetamap$ of \cref{eq:map_objective}. 
We linearize the network around its state after MAP training, $\thetastar\approx\thetamap$, and perform inference in the \glm, which typically results in a posterior with mode different from $\thetastar$. The \glm objective \cref{eq:log_joint_glm} is convex and therefore easier to optimize and guarantees convergence. For general likelihoods, posterior inference is still intractable and %
we resort to Laplace and variational approximations (see \cref{sec:background}). Both lead to Gaussian posterior approximations $q(\vtheta)$ to $\posterior$ (\protect\tikz[baseline=-0.5ex,inner sep=0pt]{\protect\draw[->, >={stealth'},line width=1pt,shorten <=0.75pt,shorten >=0.75pt, C1] (0,0) -- ++(0.5,0);} in \cref{fig:bnn_bglm}) and are computed iteratively for general likelihoods, see e.g. \citet[Chapter 4]{bishop2006pattern}; for Gaussian likelihoods they can be evaluated in a single step. On small-scale experiments (\cref{sec:exp:uci}) we found that refinement can improve performance but at a higher computational cost; we discuss computational constraints in \cref{sec:computational_considerations}.
Nonetheless, the refinement view allows us to consider the \ggn approximation separately from the Laplace approximation: the \ggn approximation linearizes the network around $\thetastar$, whereas the Laplace approximation is only one of several possible posterior approximations.

\remark{The \ggn approximation should be treated as an approximation to the model. It locally linearizes the network features and is independent of posterior inference approximations such as the Laplace approximation or variational inference.}

\subsection{The \glm predictive distribution}
\label{sec:glm_predictive}
To make predictions, we combine the approximate posterior with the likelihood; the posterior is the Laplace-\ggn posterior or a refinement thereof. %
Previous works have used the full network features in the likelihood resulting in the \bnn predictive (\cref{eq:bnn_predictive}), which was shown to severely underfit \citep{ritter2018scalable}.
Because we have effectively done inference in the \ggn-linearized model, we should instead predict using these modified features:
\begin{tcolorbox}[enhanced,colback=white,%
    colframe=C1!75!black, attach boxed title to top right={yshift=-\tcboxedtitleheight/2, xshift=-.75cm}, title=\glm predictive, coltitle=C1!75!black, boxed title style={size=small,colback=white,opacityback=1, opacityframe=0}, size=title, enlarge top initially by=-\tcboxedtitleheight/2, left=-5pt, after skip=1.5ex plus 0.5ex]
 \vskip-1.em
 \begin{equation}
\begin{aligned}
    & p_{\glm}(\vy \given \vx, \D) = \Exp_{q(\vtheta)}\left[p(\vy \given \flin(\vx, \vtheta))\right]  \\
    & \quad \approx \tfrac{1}{S} \textstyle\sum_s p(\vy \given \flin(\vx, \vtheta_s)), \quad \vtheta_s \sim q(\vtheta).
\end{aligned}\label{eq:glm_predictive}
 \end{equation}
\end{tcolorbox}
We stress that the \glm predictive in \cref{eq:glm_predictive} uses the same approximate posterior as the \bnn predictive, \cref{eq:bnn_predictive}, but locally linearized features in the likelihood.

\remark{Because the Laplace-\ggn posterior corresponds to the posterior of a linearized model, we should use this linearized model to make predictions.
In this sense, the \glm predictive is consistent with Laplace-\ggn inference, while the \bnn predictive is not.}

\subsection{Illustrative example}
\label{sec:illustrative_example}

In \cref{fig:toy_classification} we illustrate the underfitting problem of the \bnn predictive on a simple 1d binary classification problem and show how the \glm predictive resolves it.

We consider data sampled from a step function ($y = 0$ for $x<0$ and $y=1$ for $x\geq 0$) and use a 2-parameter feature function $\vf(\vx; \vtheta) = 5\text{\texttt{tanh}}(w\vx+b)$, $\vtheta=(w, b)$, Bernoulli likelihood, and factorized Gaussian prior on the parameters. The data (\protect\tikz[baseline=-0.5ex,inner sep=2pt]{\protect\node[darkgray, scale=1.5] {\pgfuseplotmark{*}};} vs \protect\tikz[baseline=-0.5ex,inner sep=2pt]{\protect\node[darkgray, scale=1.5] {\pgfuseplotmark{triangle*}};} in \cref{fig:toy_classification} \textit{left}) is ambiguous as to where the step from $0$ to $1$ occurs, such that both parameters $w$ and $b$ are uncertain.

We obtain the true parameter posterior through HMC sampling \citep{NealHMC} and find that it is symmetric \wrt the shift parameter $b$ but \emph{skewed} \wrt to the slope $w$ (see \cref{fig:toy_classification} \textit{left}). The skewness makes sense as we expect only positive slopes $w$. The corresponding posterior predictive is certain where we observe data but uncertain around the step, and the predictive mean monotonically increases from $0$ to $1$ (see \cref{fig:toy_classification} \textit{right}).

The Laplace-\ggn posterior as a Gaussian approximation is \emph{symmetric} \wrt the slope parameter $w$. It also extends to regions of the parameter space with negative slopes, $w<0$, which have no mass under the true posterior (see \cref{fig:toy_classification} \textit{left}). Samples \protect\tikz[baseline=-0.5ex,inner sep=2pt]{\protect\node[C2, scale=1.5] {\pgfuseplotmark{*}};} 
from this mismatched region result in a monotonically decreasing predictive when using the non-linear features of the \bnn predictive 
(\protect\tikz[baseline=-0.5ex,inner sep=2pt]{\protect\draw[line width=1.5pt, C2] (0,0) -- ++(0.5,0);} in \cref{fig:toy_classification} \textit{right}). In contrast, the linearized features of the \glm predictive extrapolate the behaviour around the MAP and result in a more sensible predictive in this case. Samples \protect\tikz[baseline=-0.5ex,inner sep=2pt]{\protect\node[C3, scale=1.5] {\pgfuseplotmark{*}};} 
from the matched region behave sensibly for both predictives (\protect\tikz[baseline=-0.5ex,inner sep=2pt]{\protect\draw[line width=1.5pt, C3] (0,0) -- ++(0.5,0);} in \cref{fig:toy_classification} \textit{right}). See \cref{app:exp:one_d_toy} for further details and an extended discussion.

We derive the following general intuition from this example: The Laplace-\ggn approximate posterior may be overly broad compared to the true posterior. Because the feature function $\vf(\vx; \vtheta)$ in the \bnn predictive is highly non-linear in $\vtheta$, samples $\vtheta_s$ from this mismatched region of the posterior can ulimately result in underfitting. While the \glm predictive maintains non-linearity in the inputs $\vx$, its features $\flin(\vx; \vtheta)$ are \emph{linear} in the parameters, allowing it to behave more gracefully for samples $\vtheta_s$ from the mismatched region. In other words, the \glm predictive linearly extrapolates the behavior around the MAP, while the \bnn predictive with its non-linear features can behave almost arbitrarily away from the MAP. 

\remark{The underfitting of the \bnn predictive is not a failure of the Laplace-\ggn posterior per se but is due to using a mismatched predictive model.}

\subsection{Gaussian process formulation of the \glm{}}
\label{sec:gp_predictive}
A Bayesian \glm in weight space is equivalent to a Gaussian process (\gp) in function space with a particular kernel (\protect\tikz[baseline=-0.5ex,inner sep=0pt]{\protect\draw[<->, >={stealth'},line width=1pt,shorten <=0.75pt,shorten >=0.75pt, C3, dashed] (0,0) -- ++(0.75,0);} in \cref{fig:bnn_bglm}) \citep{Rasmussen2006}.
The corresponding log joint is given by $\sum_{n=1}^N \log p(\vy_n \given \vf_n) + \log p(\vf)$, where the GP prior $p(\vf)$ is specified by its mean and covariance function that can be computed based on the expectation and covariance of \cref{eq:local_linearization} under the parametric prior $\prior=\mathcal{N}(\vm_0, \vS_0)$:
\begin{equation}
\begin{aligned}
  \vm(\vx) & = \Exp_{p(\vtheta)} [\flin\!(\vx; \vtheta)] = \flin\!(\vx; \vm_0) \\
  \vk(\vx, \vxp) & = \Cov_{p(\vtheta)} [\flin\!(\vx; \vtheta), \flin\!(\vxp; \vtheta)] \\
    & = \jacast\vS_0 \jacastshort(\vxp)\transpose.
\end{aligned}
\label{eq:ggp_mean_cov}
\end{equation}
As for the \glm, we now perform approximate inference in this \gp model or solve it in closed-form for regression; we denote the \gp posterior (approximation) by $q(\vf)$. For a single output and at $\thetastar=\thetamap$ the Laplace-\ggn approximation to \gp posterior $q(\vf^\ast)$ at a new location $\vx^\ast$ is given by~\citep{Rasmussen2006}: %
\begin{align}
    \vf^\ast \given \vxstar, \D & \sim \mathcal{N} \left(\vf(\vxstar; \thetastar), \vsigma_\ast^2 \right)\label{eq:GP_posterior}\\
    \vsigma_\ast^2 & = \vK_{\ast \ast} - \vK_{\ast N} (\vK_{NN} + \vLambda_{NN}\inv)\inv \vK_{N \ast}, \nonumber
\end{align}
where $\vK_{\ast N}$ denotes the kernel $\vk(\cdot, \cdot)$ evaluated between $\vxstar$ and the $N$ training points, and $\vLambda_{NN}$ is a diagonal matrix with entries $\vLambda(\vy_n; \vf_n)$~(\cref{eq:likelihood_hessian}).
See \cref{app:sec:gp} for the derivation and an extension to multiple outputs. Further, we can perform posterior refinement in function space by optimizing w.r.t. $\vf(X) = \jacastshort(X) \vtheta$ on a set of data points $X$,  which follows from the linearized formulation in \cref{eq:local_linearization}. Analogous to the \glm predictive, we define the \gp predictive:
\begin{tcolorbox}[enhanced,colback=white,%
    colframe=C1!75!black, attach boxed title to top right={yshift=-\tcboxedtitleheight/2, xshift=-.75cm}, title=\gp predictive, coltitle=C1!75!black, boxed title style={size=small,colback=white,opacityback=1, opacityframe=0}, size=title, enlarge top initially by=-\tcboxedtitleheight/2, left=-5pt, after skip=1.5ex plus 0.5ex, before skip=1.5ex plus 0.5ex]
 \vskip-1.em
\begin{align}
    & p_{\gp}(\vy \given \vx, \D) = \Exp_{q(\vf)}\left[p(\vy \given \vf)\right] \label{eq:GP_predictive} \\
    & \quad \approx \tfrac{1}{S} \textstyle\sum_s p(\vy \given \vf_s), \quad \vf_s \sim q(\vf). \nonumber
\end{align}
\end{tcolorbox}
Functional approximations of a \gp model are  orthogonal to parametric approximations in weight space: While parametric posterior approximations sparsify the covariances of the parameters (e.g. \kfac), functional posterior approximations consider sparsity in data space (e.g. subset of data); also see~\cref{sec:computational_considerations}.

\remark{The \glm in weight space is equivalent to a \gp in function space that enables complementary approximations.}

\subsection{Computational considerations}
\label{sec:computational_considerations}
Scalability is a major concern for inference in {\bnn}s for large-scale problems. Here, we briefly discuss practical aspects of the Laplace-\ggn computations and highlight the influence of approximations as well as implementation details; see \cref{app:sec:complexities} for further details.

\paragraph{Jacobians.}
A key component of the Laplace-\ggn approximation and our \glm are the neural network Jacobians $\jac$. %
For common architectures, the complexity of computing and storing a Jacobian is \order{PC} per datapoint for a network with $C$ class outputs and $P$ parameters.
Therefore, ad-hoc computation of Jacobians is possible while storage of all Jacobians for an entire data set of size $N$ is often prohibitive (\order{NPC}).

\paragraph{Laplace-\ggn.}
Inversion of the full covariance Laplace-\ggn approximation (\cref{eq:sigma_post}) scales cubically in the number of parameters (\order{P^3}) and is prohibitive for large neural networks; we only consider it for small problems. 
The diagonal approximation is a cheap alternative for storage and inversion (\order{P}) but misses important posterior correlations and performs worse~(see \citealt{mackay1995probable}, \cref{sec:exp:uci}, and \cref{app:exp:image_classification}).
\kfac approximations trade off between feasible computation/storage and the ability to model important dependencies within blocks, e.g.,  layers~\citep{martens2015optimizing, botev2017practical}.
Storage and computation only depend on the size of the Kronecker factors and the blocks can be inverted individually.
For scalable computation of \ggn approximations we use \texttt{backpack} for \texttt{pytorch} which makes use of additional performance improvements and does not require explicit computation of Jacobians~\citep{dangel2019backpack}.

\paragraph{Parametric predictives.}
We use $S$ Monte Carlo samples to evaluate the predictives; naively, computation of the \bnn predictive (\order{SP}) is cheaper than of the \glm predictive (\order{SPC}) due to the Jacobians. However, in both cases we can use local reparameterization~\citep{kingma2015variational} to sample either the activations per layer (\bnn predictive) or the final preactivations directly (\glm predictive) instead. 

\paragraph{Functional inference.}
\gp inference replaces inversion of the Hessian in parameter space (\order{P^3}) with inversion of the kernel matrix (\order{N^3} in computation and \order{N^2} in memory).
Additionally, we need to compute the inner products of $N$ Jacobians to evaluate the kernel.
For scalability, we consider a subset of $M \ll N$ training points to construct the kernel~(\cref{app:sec:gp}) and obtain the \gp posterior in $\mathcal{O}(M^3 + M^2P)$ and predictives per new location in  $\mathcal{O}(MP + M^2)$. 
We found that $M \geq 50$ already improves performance over the MAP (see ablations in \cref{app:exp:image_classification}) even when $N$ was orders of magnitude larger; increasing $M$ strictly improved performance.
Instead of a naive subset approximation we could also use sparse approximations \citep{titsias2009_variational_gp,hensman2015scalable} to scale the kernel computations.

\paragraph{\gp and \glm refinement.}
To perform posterior refinement (cf. \cref{sec:glm_approximate_inference,sec:gp_predictive}) efficiently, we have to compute and store the Jacobians on all data, as we require them in every iterative update step. For large networks and datasets we are memory bound and, thus, only consider refinement for small problems in~\cref{sec:exp:uci}.

\begin{figure*}[ht]
    \centering
    \begin{tikzpicture}
    \node (left) {\includegraphics[width=3.7in]{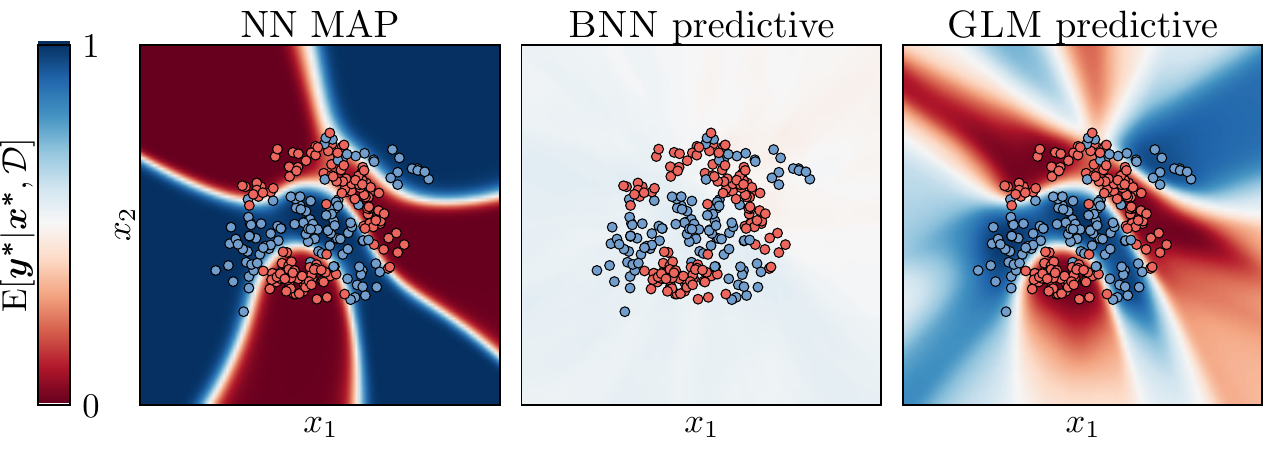}};
    \node[right=.75cm of left] (right){\includegraphics[width=2.6in]{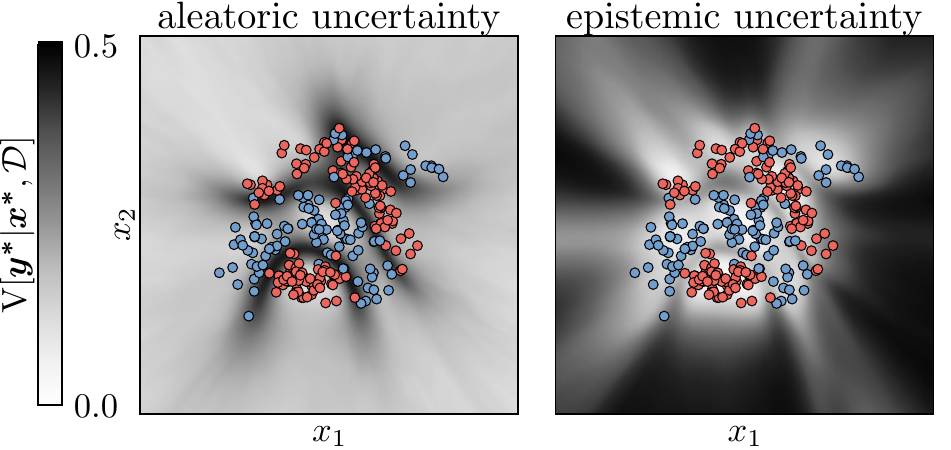}};
    \node[anchor=south] at ($(left.north)+(0.45,-0.1)$) {Predictive mean};
    \node[anchor=south] at ($(right.north)+(0.4,-0.1)$) {\glm predictive variance};
    \end{tikzpicture}
    \vskip-1.25em
    \caption{Binary classification on the banana dataset. \textit{left:} Predictive means; the \bnn predictive severely underfits; like the MAP, our proposed \glm predictive makes meaningful predictions but also becomes less certain away from the data. \textit{right:} The \glm predictive variance (see \cref{app:exp_details_toy}) decomposes into meaningful aleatoric (data-inherent) uncertainty at class boundaries and epistemic (model-specific) uncertainty away from data.}
    \label{fig:toy_new}
\end{figure*}

\begin{table*}[ht]
    \centering
    \small
    \begin{tabular}{l C C C C C C C C}
    \toprule
    \textbf{Dataset} & \textbf{NN MAP} & \textbf{MFVI} & \textbf{\bnn}& \textbf{\glm} & \textbf{\glm diag} & \textbf{\glm refine} & \textbf{\glm refine d} \\\midrule
    \textbf{australian} & \mathbf{0.31 \pms{0.01}} & 0.34 \pms{0.01} & 0.42 \pms{0.00} & \mathbf{0.32 \pms{0.02}} & 0.33 \pms{0.01} & \mathbf{0.32 \pms{0.02}} & \mathbf{0.31 \pms{0.01}}  \\
    \textbf{cancer} & \mathbf{0.11 \pms{0.02}} & \mathbf{0.11 \pms{0.01}} & 0.19 \pms{0.00} & \mathbf{0.10 \pms{0.01}} & \mathbf{0.11 \pms{0.01}} & \mathbf{0.11 \pms{0.01}} & 0.12 \pms{0.02}  \\
    \textbf{ionosphere} & 0.35 \pms{0.02} & 0.41 \pms{0.01} & 0.50 \pms{0.00} & \mathbf{0.29 \pms{0.01}} & 0.35 \pms{0.01} & 0.35 \pms{0.05} & 0.32 \pms{0.03}  \\
    \textbf{glass} & 0.95 \pms{0.03} & 1.06 \pms{0.01} & 1.41 \pms{0.00} & 0.86 \pms{0.01} & 0.99 \pms{0.01} & 0.98 \pms{0.07} & \mathbf{0.83 \pms{0.02}}  \\
    \textbf{vehicle} & 0.420 \pms{0.007} & 0.504 \pms{0.006} & 0.885 \pms{0.002} & 0.428 \pms{0.005} & 0.618 \pms{0.003} & \mathbf{0.402 \pms{0.007}} & 0.432 \pms{0.005}  \\
    \textbf{waveform} & \mathbf{0.335 \pms{0.004}} & 0.393 \pms{0.003} & 0.516 \pms{0.002} & \mathbf{0.339 \pms{0.004}} & 0.388 \pms{0.003} & \mathbf{0.335 \pms{0.004}} & 0.364 \pms{0.008}  \\
    \textbf{digits} & \mathbf{0.094 \pms{0.003}} & 0.219 \pms{0.004} & 0.875 \pms{0.002} & 0.250 \pms{0.002} & 0.409 \pms{0.002} & 0.150 \pms{0.002} & 0.149 \pms{0.008}  \\
    \textbf{satellite} & 0.230 \pms{0.002} & 0.307 \pms{0.002} & 0.482 \pms{0.001} & 0.241 \pms{0.001} & 0.327 \pms{0.002} & \mathbf{0.227 \pms{0.002}} & 0.248 \pms{0.002}  \\
    \bottomrule
    \end{tabular}
    \vskip-0.5em
    \caption{Negative test log likelihood (lower is better) on UCI %
    classification tasks ($2$ hidden layers, $50$ \texttt{tanh}).
    The \glm predictive clearly outperforms the \bnn predictive; %
    the \glm posterior refined with variational inference is overall the best method. This also holds for accuracy and calibration and on other architectures, see \cref{app:exp_details_uci}.}
    \label{tab:uci}
\end{table*}

\section{Experiments}
We empirically evaluate the proposed \glm predictive for the Laplace-\ggn approximated posterior in weight space (\cref{eq:glm_predictive}) and the corresponding \gp predictive in function space (\cref{eq:GP_predictive}).
We compare them to the \bnn predictive (\cref{eq:bnn_predictive}) with same posterior for several sparsity structures of the Laplace and variational approximation as well as mean-field VI (BBB, \citet{blundell2015weight}) and a dampened \kfac Laplace-\ggn approximation with \bnn predictive~\citep{ritter2018scalable}.

We consider a second example on $2d$ binary classification (\cref{sec:exp:banana}), several small-scale classification problems (\cref{sec:exp:uci}), for which posterior refinement is possible, as well as larger image classification tasks (\cref{sec:exp:image}). We close with an application of the \glm predictive to out-of-distribution (OOD) detection (\cref{sec:exp:ood}).
Because the \glm predictive for $\thetastar=\thetamap$ is identical to \citet{foong2019between} and \citet{khan2019approximate}, we focus on classification and refer to their works for regression.

In all experiments, we use a diagonal prior, $\prior=\mathcal{N}(0, \delta^{-1} \vI_P)$, and choose its precision $\delta$ based on the negative log likelihood on a validation set for each dataset, architecture, and method. 
The prior precision $\delta$ corresponds to \emph{weight-decay} with factor $\tfrac{\delta}{N}$.
For each task, we first train the network to find a MAP estimate using the objective \cref{eq:map_objective} and the Adam optimizer~\citep{kingma2014adam}.
We then compute the different posteriors and predictives using the values of the parameters after training, $\thetastar$ (details in \cref{app:experiments}).

The proposed \glm and \gp predictives consistently resolve underfitting problems of the \bnn predictive, and are on par or better than other methods considered.%

\begin{table*}[ht]
    \vskip-0.5em
    \centering
    \small
    \begin{tabular}{>{\bfseries}l l C C C C}
    \toprule
    \textbf{Dataset} & \textbf{Method} &  \textbf{Accuracy} \uparrow    &   \textbf{NLL} \downarrow   &   \textbf{ECE} \downarrow    &    \textbf{OOD-AUC} \uparrow   \\
    \midrule
    \multirow{5}{*}{FMNIST}& MAP & 91.39\pms{0.11} & 0.258\pms{0.004} & 0.017\pms{0.001} & 0.864\pms{0.014} \\
    &             \bnn predictive & 84.42\pms{0.12} & 0.942\pms{0.016} & 0.411\pms{0.008} & 0.945\pms{0.002} \\
    &             \bnn predictive (\citeauthor{ritter2018scalable}) & 91.20\pms{0.07} & 0.265\pms{0.004} & 0.024\pms{0.002} & 0.947\pms{0.006} \\
    &             \glm predictive (\emph{ours}) & \mathbf{92.25\pms{0.10}} & \mathbf{0.244\pms{0.003}} & 0.012\pms{0.003} & \mathbf{0.955\pms{0.006}} \\
    &             \gp predictive (\emph{ours}) & 91.36\pms{0.11} & 0.250\pms{0.004} & \mathbf{0.007\pms{0.001}} & 0.918\pms{0.010} \\
    \midrule
    \multirow{5}{*}{CIFAR10} & MAP & 80.92\pms{0.32} & 0.605\pms{0.007} & 0.066\pms{0.004} & 0.792\pms{0.008} \\
    &             \bnn predictive & 21.74\pms{0.80} & 2.114\pms{0.021} & 0.095\pms{0.012} & 0.689\pms{0.020} \\
    &             \bnn predictive (\citeauthor{ritter2018scalable}) & 80.78\pms{0.36} & 0.588\pms{0.005} & 0.052\pms{0.005} & 0.783\pms{0.007} \\
    &             \glm predictive (\emph{ours}) & \mathbf{81.37\pms{0.15}} & 0.601\pms{0.008} & 0.084\pms{0.010} & \mathbf{0.843\pms{0.016}} \\
    &             \gp predictive (\emph{ours}) & 81.01\pms{0.32} & \mathbf{0.555\pms{0.008}} & \mathbf{0.017\pms{0.003}} & 0.820\pms{0.013} \\
    \bottomrule
    \end{tabular}
    \vskip-0.5em
    \caption{Accuracy, negative test log likelihood (NLL), expected calibration error (ECE) on the test set, and area under the curve for out-of-distribution detection (OOD-AUC).
    The proposed methods (\glm and \gp predictive) outperform the \bnn predictive with same posterior and with dampened (concentrated) posterior~\protect\citep{ritter2018scalable} as well as the MAP (point-)estimate posterior on most tasks and metrics. See~\protect\cref{app:exp:image_classification} for further results.
    }
    \label{tab:image_datasets}
\end{table*}

\subsection{Second illustrative example}
\label{sec:exp:banana}
First, we consider $2d$ binary classification on the banana dataset in \cref{fig:toy_new}.
We use a neural network with $2$ hidden layers of $50$ \texttt{tanh} units each and compare the \bnn and the \glm predictive for the same full Laplace-\ggn posterior (experimental details and additional results for MFVI and diagonal posteriors in \cref{app:exp_details_toy}).

Like in the $1d$ example (\cref{fig:toy_classification}), the \bnn predictive severely underfits compared to the MAP; its predictive mean is completely washed out and its variance is very large everywhere (see \cref{app:exp_details_toy}).
Using the same posterior but the proposed \glm predictive instead resolves this problem. %
In contrast to the MAP point-estimate, our \glm predictive with Laplace-\ggn posterior leads to growing predictive variances away from the data in line with previous observations for regression~\citep{foong2019between, khan2019approximate}.
Moreover, the \glm predictive variance decomposes into meaningful aleatoric (data-inherent) uncertainty at the boundaries between classes and epistemic (model-specific) uncertainty away from the data \citep{kwon2020uncertainty} (\cref{fig:toy_new} \textit{(right)}).
In \cref{app:exp_details_toy} we show that the \glm predictive easily adapts to deeper and shallower architectures and yields qualitatively similar results in all cases, whereas the \bnn predictive performs even worse for deeper (more non-linear) architectures. MFVI requires extensive tuning and yields lower quality results.

\subsection{UCI classification}
\label{sec:exp:uci}
We now compare the different methods on a set of UCI classification tasks on a network with $2$ hidden layers of $50$ \texttt{tanh} units.
On this scale, posterior refinement in the \glm using variational inference is feasible as discussed in \cref{sec:glm_approximate_inference}.
In \cref{tab:uci}, we report the test log predictive probabilities over 10 splits ($70\%$ train/$15\%$ valid/$15\%$ test).
See \cref{app:exp_details_uci} for details and results for accuracy and calibration as well as on other architectures.

\begin{figure}[htb]
    \vskip-0.5em
     \centering
     \includegraphics[width=3.25in]{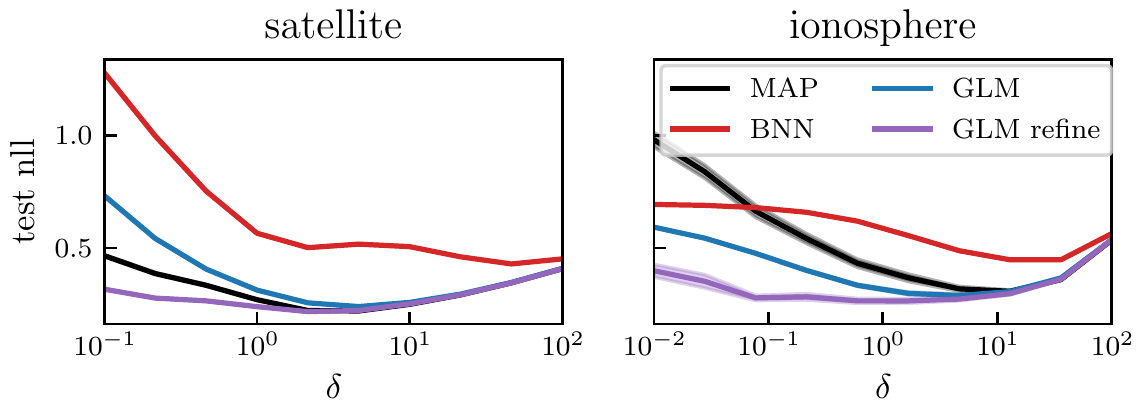}
     \vskip-0.75em
     \caption{The \glm predictive
     (\protect\tikz[baseline=-0.5ex,inner sep=0pt]{\protect\draw[line width=1.5pt, C1] (0,0) -- ++(0.4,0);}/\protect\tikz[baseline=-0.5ex,inner sep=0pt]{\protect\draw[line width=1.5pt, C5] (0,0) -- ++(0.4,0);})
     outperforms the \bnn predictive
   (\protect\tikz[baseline=-0.5ex,inner sep=0pt]{\protect\draw[line width=1.5pt, C4] (0,0) -- ++(0.4,0);}) \emph{for all} settings of the prior precision hyperparameter $\delta$.}
     \label{fig:uci_sweep}
     \vspace{-1em}
 \end{figure}

Using the same Laplace-\ggn posterior, the \glm predictive (``\glm'' in \cref{tab:uci}) clearly outperforms the \bnn predictive (``\bnn'') on almost all datasets and metrics considered.
Moreover, the proposed posterior refinement using variational inference in the \glm (``\glm refine'') can further boost performance.
The proposed methods also perform consistently better than MFVI on most datasets, even when considering only a diagonal posterior approximation (``... d(iag)''); and they easily adapt to deeper architectures, unlike MFVI, which is often hard to tune (see \cref{app:exp_details_uci}).
In \cref{fig:uci_sweep} we highlight that the \glm predictive consistently outperforms the \bnn predictive for \emph{any} setting of the prior precision hyperparameter $\delta$ and that posterior refinement consistently improves over the MAP estimate.
 \begin{figure*}[ht]
    \centering
    \includegraphics[width=6.7in]{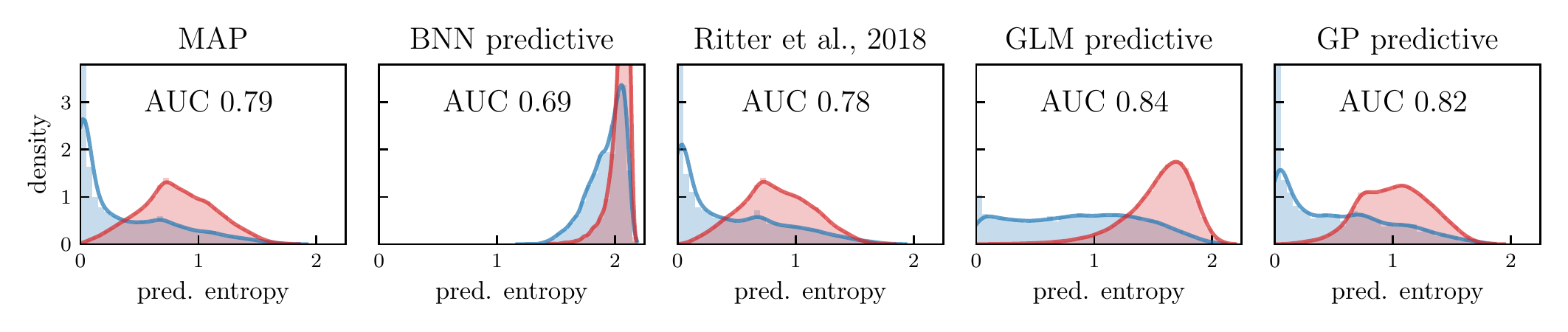}
    \vspace{-1em}
    \caption{In-distribution
    (\protect\tikz[baseline=.5ex,inner sep=0pt]{\protect\draw[line width=1.5pt, C1, name path=A] (0,0.3) -- ++(0.75,0); \protect\draw[name path=B, draw=none] (0,0) -- ++(0.75,0);\protect\tikzfillbetween[of=A and B]{C1, opacity=0.3};}, CIFAR10)
    vs out-of-distribution (\protect\tikz[baseline=.5ex,inner sep=0pt]{\protect\draw[line width=1.5pt, C4, name path=A] (0,0.3) -- ++(0.75,0); \protect\draw[name path=B, draw=none] (0,0) -- ++(0.75,0);\protect\tikzfillbetween[of=A and B]{C4, opacity=0.3};}, SVHN) detection using a fully convolutional architecture.
    The MAP is overconfident while the \bnn predictive is underconfident.
    \gp and \glm predictives show best out-of-distribution detection (area under the curve, AUC), also see \cref{tab:image_datasets}.}
    \label{fig:cifar10_ood}
\end{figure*}
\subsection{Image classification}%
\label{sec:exp:image}
As larger scale problems, we consider image classification on MNIST \citep{lecun2010mnist}, FashionMNIST \citep{FashionMNIST}, and CIFAR10 \citep{Cifar10}. We use a \kfac Laplace-\ggn approximation for parametric models and a subset posterior approximation with $M=3200$ data points for the \gp. We compare to the MAP estimate and to the \bnn predictive with same posterior as well as with dampened posterior \citep{ritter2018scalable} and present results for several performance metrics on CNNs in \cref{tab:image_datasets}; see \cref{app:exp:image_classification} for details, additional results on MNIST, other network architectures, and diagonal approximation.%

As for the other problems, the \glm predictive consistently outperforms the \bnn predictive by a wide margin using the same posterior. It also typically outperforms the \bnn predictive with dampened (concentrated) posterior \citep{ritter2018scalable}, in particular for fully connected networks, see \cref{app:exp:image_classification}. While the \glm predictive performs best on most tasks in terms of accuracy and negative test log likelihood, the \gp predictive interestingly achieves better expected calibration error (ECE) \citep{naeini2015obtaining}. We attribute the improved calibration to the \gp implicitly using a full-covariance Laplace-\ggn, while the parametric approaches are limited to a \kfac approximation of the posterior covariance. However, the \gp is limited to a subset of the training data to make predictions; we hypothesize that better sparse approximations could further improve its performance on accuracy and negative log likelihood.

\subsection{Out-of-distribution detection}
\label{sec:exp:ood}
We further evaluate the predictives on out-of-distribution (OOD) detection on the following in-distribution (ID)/OOD pairs: MNIST/FMNIST, FMNIST/MNIST, and CIFAR10/SVHN. Following~\citet{osawa2019practical, ritter2018scalable}, we compare the entropies of the predictive distributions on ID vs OOD data and the associated OOD detection performance measured in terms of the area under the curve (OOD-AUC). We use the same \kfac posterior approximations as in~\cref{sec:exp:image}; see \cref{app:exp:image_classification,app:exp:ood} for details and additional results on other ID/OOD pairs.

We provide OOD detection performance (OOD-AUC) in~\cref{tab:image_datasets} for FMNIST/MNIST and CIFAR10/SVHN and compare the predictive entropy histograms for CIFAR10/SVHN in~\cref{fig:cifar10_ood}.
Across all tasks considered, we find that the \glm predictive achieves the best OOD detection performance, while the \bnn predictive consistently performs worst. The \bnn predictive with concentrated (dampened) Laplace-\ggn posterior \citep{ritter2018scalable} improves over the undampened posterior, but performs worse than the \glm predictive. %
\section{Related Work}
The Laplace approximation for {\bnn}s was first introduced by \citet{mackay1992bayesian} who applied it to small networks using the full Hessian but also suggested an approximation similar to the generalized Gauss Newton~\citep{mackay1992evidence}.
\citet{foresee1997gauss} later used the Gauss-Newton for Bayesian regression neural networks with Gaussian likelihoods.
The generalized Gauss-Newton~\citep{martens2014new} in conjunction with scalable factorizations or diagonal Hessian approximations \citep{martens2015optimizing, botev2017practical} enabled a revival of the Laplace approximation for modern neural networks~\citep{ritter2018scalable, khan2019approximate}. \citet{bottou2018optimization} discuss the linearizing effect of the \ggn approximation for MAP or maximum likelihood optimization; here we use this interpretation to obtain a consistent Bayesian predictive.

To address underfitting problems of the Laplace \citep{lawrence2001phd} that are particularly egregious when combined with the \ggn, \citet{ritter2018scalable} introduced a Kronecker factored Laplace-\ggn approximation, which does not seem to suffer in the same way despite using the same \bnn predictive.
Our analysis and experiments suggest that this is because of an additional ad-hoc approximation they introduce, dampening, which can reduce the posterior covariance (see \cref{app:kronecker}). 
Dampening is typically used in optimization procedures using Kronecker-factored Hessian approximations~\citep{martens2015optimizing} but can lead to significant distortions when applied to a posterior approximation. In contrast, we use an undampened Laplace-\ggn posterior in combination with the \glm predictive to resolve underfitting.

For Gaussian likelihoods our \glm predictive recovers the analytically tractable ``linearized Laplace'' model \citep{foong2019between} as well as \dnntogp \citep{khan2019approximate}. Both apply the Laplace and \ggn approximations jointly at the posterior mode $\thetastar=\thetamap$ and are limited to regression. %
We separate the \ggn from approximate inference %
to derive an explicit \glm model for general likelihoods and to justify the \glm predictive. Our experiments generalize their observations to general likelihoods.
\citet{khan2019approximate} introduce \dnntogp to relate inference in (linearized) {\bnn}s to {\gp}s but are limited to Gaussian likelihoods. Our approach builds on their work but considers general likelihoods; therefore, we obtain a similar \gp covariance function that is related to the neural tangent kernel (NTK) \citep{jacot2018neural}.
Our proposed \emph{refinement} is related to training an empirical NTK~\citep{lee2019wide}.
In contrast to the empirical NTK, the \ggn corresponds to a local linearization at the MAP and not at a random initialization.
Therefore, we expect that these learned feature maps represent the data better.

Out-of-distribution detection has become a benchmark for predictive uncertainties~\citep{nalisnick2018deep}, on which many recent \bnn approaches are evaluated, e.g.,~\citet{ritter2018scalable, osawa2019practical, wenzel2020good}.
Our simple change in the predictive also leads to improved OOD detection.

\section{Conclusion}
In this paper we argued that in Bayesian deep learning, the frequently utilized generalized Gauss-Newton (\ggn) approximation should be understood as a modification of the underlying probabilistic model and should be considered separately from approximate posterior inference. 
Applying the \ggn approximation turns a Bayesian neural network (\bnn) locally into a  generalized linear model or, equivalently, a Gaussian process. Because we then use this linearized model for inference, we should also predict using these modified features in the likelihood rather than the original \bnn features. 
The proposed \glm predictive extends previous results by \citet{khan2019approximate} and \citet{foong2019between} to general likelihoods and resolves underfitting problems observed e.g. by \citet{ritter2018scalable}. We conclude that underfitting is not due to the Laplace-\ggn posterior but is caused by using a mismatched model in the predictive distribution.
We illustrated our approach on several simple examples, demonstrated its effectiveness on UCI and image classification tasks, and showed that it can be used for out-of-distribution detection. 
In future work, we aim to scale our approach further.

\acknowledgments{We thank Emtiyaz Khan for the many fruitful discussions that lead to this work as well as Michalis Titsias and Andrew Foong for feedback on the manuscript.
We are also thankful for the RAIDEN computing system and its support team at the RIKEN AIP.}

\subsubsection*{References}
\printbibliography[heading=none]

\clearpage
\appendix
\onecolumn
\renewcommand\thefigure{\thesection\arabic{figure}}
\setcounter{figure}{0}
\renewcommand\theequation{\thesection\arabic{equation}}
\setcounter{equation}{0}
\renewcommand\thetable{\thesection\arabic{table}}
\setcounter{table}{0}
\linewidth\hsize \toptitlebar {\centering
{\Large\bfseries \ourtitle\\ (Appendix) \par}}
 \bottomtitlebar

\makeatletter
\renewcommand*\l@section{\@dottedtocline{1}{1.5em}{2.3em}}
\makeatother

\setcounter{page}{1}
\counterwithin{figure}{section}
\counterwithin{equation}{section}

\renewcommand{\contentsname}{Contents of the Appendix}
\addtocontents{toc}{\protect\setcounter{tocdepth}{2}}
\tableofcontents

\section{Derivations and additional details}

In this section we provide additional derivations and details on the background as well as proposed methods, the \glm predictive and its \gp equivalent.
As discussed in the main paper, the Laplace-GGN approximation to the posterior at the MAP, $\thetastar = \thetamap$, takes the form:
\begin{equation}
  \label{app:eq:lap_ggn}
  q_{\ggn}(\vtheta) = \mathcal{N}(\thetamap, \vS) = \mathcal{N} \bigg( \thetamap, \bigg(\sum_{(\vx, \vy) \in \D} \jac \transpose \vLambda(\vy; \vf)\jac + \vS_0\inv \bigg)\inv \bigg),\tag{\ref{eq:sigma_post}}
\end{equation}
where $\vS_0$ denotes the prior covariance of the parameters, $\prior=\mathcal{N}(\vmu_0, \vS_0)$.
Because using the full posterior covariance is infeasible in practise, we have to apply further approximations.
In \cref{app:kronecker} we discuss the \kfac approximation that is both expressive and scalable and therefore our choice for large-scale experiments.
In \cref{app:sec:gp}, we discuss \gp inference with special attention to the subset-of-data approach.
At the end of this section, we discuss the computational complexities of the proposed predictives and compare them to the \bnn predictive.

\subsection{\kfac approximation of the Laplace-\ggn}%
\label{app:kronecker}

\citet{ritter2018scalable} first proposed a Laplace-\ggn approximation that utilizes a \kfac approximation to the posterior enabled by scalable \kfac approximations to the \ggn and Fisher information matrix~\citep{botev2017practical, martens2015optimizing}.
Here, we discuss the \kfac posterior approximation that we use and compare it to theirs.
In particular, we do not use nor require dampening and post-hoc adjustment of the prior precision which would render the MAP estimate invalid.
We use the \kfac posterior approximation for large-scale image classification tasks because it can model parameter covariances per layer while maintaining scalability;
a diagonal posterior approximation fails to model these relationships while the full \ggn is intractable for larger networks.

\citet{botev2017practical} and \citet{martens2015optimizing} propose a \kfac approximation to the expected \ggn Hessian approximation under the data distribution and (for the likelihoods considered) Fisher information matrix, respectively.
The \kfac approximation typically is block-diagonal and factorizes across neural network layers.
Critically, the \kfac approximation is based on the observation that the \ggn approximation to the Hessian of the likelihood for \emph{a single data point} can be written as a Kronecker product exactly.
For a single data point $(\vx_n, \vy_n)$ and parameter group $l$, we can write
\begin{equation}
    \left[ \jacshort_{\vtheta}(\vx_n) \transpose \vLambda(\vy_n; \vf_n)\jacshort_{\vtheta}(\vx_n) \right]_l = \vQ_l^{(n)} \kron \vW_l^{(n)},
    \label{app:eq:kron_individual}
\end{equation}
where $\vQ_l^{(n)}$ depends on the pre-activations of layer $l$ and $\vW_l^{(n)}$ can be computed recursively; for details, see \citet{botev2017practical}.
In the case of a fully connected layer parameterized by $\vW \in \Reals^{D_\textrm{out}^{(l)} \times D_\textrm{in}^{(l)}}$, we map from an internal representation of dimensionality $D_\textrm{in}^{(l)}$ to dimensionality $D_\textrm{out}^{(l)}$.
Then, the Kronecker factors in \cref{app:eq:kron_individual} are square matrices: $\vQ_l^{(n)} \in \Reals^{D_\textrm{out}^{(l)} \times D_\textrm{out}^{(l)}}$ and $\vW_l^{(n)} \in \Reals^{D_\textrm{in}^{(l)} \times D_\textrm{in}^{(l)}}$.
Critically, both matrices are positive semi-definite.

However, the full \ggn requires a sum of \cref{app:eq:kron_individual} over all data points.
While each individual summand allows a Kronecker factorization, this is not necessarily the case for the sum.
For tractability, \citet{martens2015optimizing} and \citet{botev2017practical} propose to approximate the sum of Kronecker products by a Kronecker product of sums:
\begin{equation}
\label{app:eq:kfac_ggn}
  \left[{\textstyle \sum_{n=1}^N}\jacshort_{\vtheta}(\vx_n) \transpose \vLambda(\vy_n; \vf_n)\jacshort_{\vtheta}(\vx_n)\right]_l
  = {\textstyle \sum_{n=1}^N}\vQ_l^{(n)} \kron \vW_l^{(n)}
  \approx \left( {\textstyle \sum_{n=1}^N}\vQ_l^{(n)} \right) \kron \left( {\textstyle \sum_{n=1}^N} \vW_l^{(n)} \right)
  =: \vQ_l \kron \vW_l,
\end{equation}
which defines $\vQ_l$ and $\vW_l$ as the sum of per-data-point Kronecker factors.
Approximating the sum of products with a product of sums is quite crude since we add additional cross-terms due to the distributivity of the Kronecker product.

The \kfac Laplace-\ggn posterior approximation is given by using the above approximation to the \ggn in the definition of the posterior covariance $\vS$ in \cref{app:eq:lap_ggn}.
As considered in the experiments and typical for Bayesian deep learning~\citep{ritter2018scalable, khan2018fast, khan2019approximate, zhang2018noisy}, we use an isotropic Gaussian prior $p(\vtheta) = \mathcal{N}(0, \delta\inv \vI)$.
We denote by $\vS_\textrm{\kfac} \approx \vS$ the \kfac Laplace-\ggn approximation which is then given by
\begin{equation}
  \label{app:eq:laplace_kfac}
    \vS\inv_\textrm{\kfac} = \vQ_l \kron \vW_l + \delta \vI_l,
\end{equation}
where $\vI_l$ is the identity matrix with the number of parameters of the $l$-th layer.
The entire term does not necessarily allow a Kronecker-factored representation.

To that end, \citet{ritter2018scalable} propose to approximate the posterior precision further, motivated by maintaining a Kronecker factored form, as:
\begin{equation}
  \vS\inv_\textrm{\kfac} \approx ( \vQ_l + \sqrt{\delta} \vI) \kron ( \vW_l + \sqrt{\delta} \vI ) = \vQ_l \kron \vW_l + \vQ_l \kron \sqrt{\delta} \vI + \sqrt{\delta} \vI \kron \vW_l + \delta \vI_l, \label{app:eq:ritter_approx}
\end{equation}
which artificially increases the posterior concentration by the cross-products.
That is, because the cross-products are positive semi-definite matrices, they increase the eigenvalues of the posterior precision and thereby reduce the posterior covariance, see \cref{app:fig:1d_toy} for a visual example.
This further approximation is commonly referred to as \emph{dampening} in the context of second-order optimization methods~\citep{botev2017practical, zhang2018noisy}.
While \citet{ritter2018scalable} do not motivate this approximation in particular, we find that it is in fact necessary to mediate underfitting issues of \bnn predictive when using a Laplace-\ggn posterior, see \cref{sec:exp:image} and \cref{app:exp:image_classification}.
Our proposed \glm predictive does not require a dampened/concentrated posterior, but also works in this case. %
We now show that the dampening approximation in \cref{app:eq:ritter_approx} is, in fact, also not necessary from a computational perspective; that is, we show how to compute $\vS_\textrm{KFAC}$ (\cref{app:eq:laplace_kfac}) without this approximation while keeping the complexity unchanged.

To avoid the additional approximation in form of \emph{dampening} that artificially concentrates the posterior, we work with an eigendecomposition of the Kronecker factors.
Let $\vQ_l = \vM_{Q_l} \vD_{Q_l} \vM_{Q_l}\transpose$ be the eigendecomposition of $\vQ_l$ and likewise for $\vW_l$, where $\vD$ is a diagonal matrix of eigenvalues that are non-negative as both Kronecker factors are positive semi-definite.
We have
\begin{align*}
  \vS\inv_\textrm{\kfac} &= (\vM_{Q_l} \vD_{Q_l} \vM_{Q_l}\transpose) \kron (\vM_{W_l} \vD_{W_l} \vM_{W_l}\transpose) + \delta \vI_l \\
          &= (\vM_{Q_l} \kron \vM_{W_l}) (\vD_{Q_l} \kron \vD_{W_l}) (\vM_{Q_l}\transpose \kron \vM_{W_l}\transpose) + \delta (\vM_{Q_l} \kron \vM_{W_l})  (\vM_{Q_l}\transpose \kron \vM_{W_l}\transpose)\\
          &= (\vM_{Q_l} \kron \vM_{W_l}) (\vD_{Q_l} \kron \vD_{W_l} + \delta \vI_l ) (\vM_{Q_l}\transpose \kron \vM_{W_l}\transpose),
\end{align*}
which follows form the fact that $\vM_{Q_l} \kron \vM_{W_l}$ forms the eigenvectors of $\vQ \kron \vW$ and is therefore unitary and can be used to write the identity $\vI_l$.

Based on the eigendecomposition of the individual Kronecker factors, the required quantities for sampling from the posterior (square root) or inversion can be computed efficiently.
In comparison to the approximation of \citet{ritter2018scalable}, we only incur additional storage costs due to the eigenvalues $\vD$ which are negligible in size since they are diagonal and the major cost drivers are the dense matrices $\vM$.

Working with the eigendecomposition further enables to understand the posterior concentration induced by dampening better.
Recall that $\vS\inv_\textrm{\kfac}$ is the posterior parameter precision and therefore higher concentration (i.e. eigenvalues) lead to less posterior uncertainty.
instead of the diagonal $\vD_{Q_l} \kron \vD_{W_l} + \delta \vI_l$, the approximation of \citet{ritter2018scalable} \emph{adds} the additional diagonal terms $\vD_{Q_l} \kron \sqrt{\delta} \vI + \sqrt{\delta} \vI \kron \vD_{W_l}$.
As all eigenvalues are positive and $\delta > 0$, this corresponds to an artificial concentration of the posterior by increasing the eigenvalues.
The concentration is also uncontrolled as we scale the eigenvalues of the Kronecker factors by the prior precision -- normally, these terms should only be combined additively.

\subsection{Gaussian process approximate inference}
\label{app:sec:gp}

In \cref{sec:gp_predictive} we presented an equivalent Gaussian process (\gp) formulation of the \glm, which can give rise to orthogonal posterior approximations.

The starting point is the Bayesian \glm with linearized features
\begin{align}
    \networklin = \networkast + \jacast(\vtheta - \vtheta^\ast), \tag{\ref{eq:local_linearization}}
\end{align}
and prior on the parameters $\prior=\mathcal{N}(\vm_0, \vS_0)$. As also discussed in the main paper, we obtain the corresponding Gaussian process prior model with mean function $\vm(\vx)$ and covariance (kernel) function $\vk(\vx, \vxp)$ by computing the mean and covariance of the features \citep{Rasmussen2006}:
\begin{equation}
\begin{aligned}
  \vm(\vx) & = \Exp_{p(\vtheta)} [\flin\!(\vx; \vtheta)] = \flin\!(\vx; \vm_0) \\
  \vk(\vx, \vxp) & = \Cov_{p(\vtheta)} [\flin\!(\vx; \vtheta), \flin\!(\vxp; \vtheta)]
    = \jacast\vS_0 \jacastshort(\vxp)\transpose.
\end{aligned}
\tag{\ref{eq:ggp_mean_cov}}
\end{equation}
This specifies the functional \gp prior $p(\vf)$ over the $C$-dimensional function outputs, which is then used together with the same likelihood as for the \glm.
In line with the \bnn and \glm model, we choose an isotropic prior on the parameters with covariance $\vS_0 = \delta\inv \vI$ and zero mean $\vm_0= 0$. The \gp mean function is therefore given by $\vm(\vx) = \networkast - \jacast\vtheta^\ast)$, such that the latent function $\vf(X)$ is modelled by this mean function and a zero centred \gp that models the fluctuations around the mean.
For multiple outputs (e.g. multi-class classification), we assume independent {\gp}s per output dimension.

The \gp log joint distribution can be written as~(see \cref{sec:gp_predictive})
\begin{align}
    \log p(\D, \vf(X)) = \sum_{n=1}^N \log p(\vy_n \given \vf_n) + \log p(\vf).
    \label{app:eq:gp_log_joint}
\end{align}
We now need to compute the functional posterior distribution over the training data set $\vf(X) \in \reals^{NC}$, which does not have a closed form for general likelihoods.

\subsubsection{Laplace approximation}
In our work, we consider the Laplace approximation to the \gp posterior~\citep{Rasmussen2006} under the assumption that $\thetastar = \thetamap$.
Here, we present details on the Laplace approximation and give formulas for the multi-output case with $C$ outputs and $N$ data points.
We follow \citet{pan2020continual} and assume that the mode of the \glm and \gp coincide, which allows us to directly construct a Laplace approximation at $\flin\!(\vx;\thetamap)$.

The posterior predictive covariance is given by the Hessian of the log joint distribution \cref{app:eq:gp_log_joint} \wrt the latent function \vf, $\nabla^2_{\vf \vf} \log p (\D, \vf(X))$, similarly to in the parametric Laplace approximation.
For multi-output predictions, we obtain a block diagonal covariance matrix, because we assume independent {\gp}s for each output.
To define the multi-output predictive covariance, we therefore only need to compute the additional stacked second derivative of the log likelihood, the $\vLambda(\vy;\vf)$ terms, in the right format.
We define the block-diagonal matrix $\vL_{NN} \in \Reals^{NC \times NC}$ where the $n$-th block is given by the $C\times C$ matrix $\vLambda(\vy_n, \vf(\vx_n,\thetastar))$.
We then obtain $\nabla^2_{\vf \vf} \log p (\D, \vf(X)) = -\vL_{NN} - \delta (\jacastshort (X) \jacastshort (X)\transpose)\inv$ similar to~\citet[chap. 3]{Rasmussen2006}.
Using the matrix inversion Lemma, we can then write the posterior as
\begin{equation}
  q(\vf^\ast \given \vxstar, \D) =   \mathcal{N}\left(\vf(\vxstar; \thetastar), \vSigma_\ast \right)
  \quad \textrm{with} \quad
    \vSigma_\ast  = \vK_{\ast \ast} - \vK_{\ast N} (\vK_{NN} + \vL_{NN}\inv)\inv \vK_{N \ast},
    \label{app:eq:gp_posterior}
\end{equation}
where $\vK_{\ast N}$ is the kernel between the input location and the $N$ training samples, i.e., $\vK_{\ast N} = \delta\inv \jacastshort(\vxstar) \jacastshort(X)\transpose$.
In comparison to \cref{eq:GP_posterior} for a single output, the kernel evaluated on the training data now has size $NC\times NC$ instead of $N \times N$.
Note that the predictive mean is simply defined by the MAP of the neural network, the \gp simply augments it with predictive uncertainty.
Assuming independent prior {\gp}s for each output, i.e., $p(\vf) = \prod_{c=1}^C p(\vf_c)$, scalability can be maintained due to parallelization as we can deal with $C$ $N \times N$ kernels independently in parallel~\citep{Rasmussen2006}.

\subsubsection{Subset of data}
However, even $N \times N$ kernel matrices are intractable for larger datasets (\order{N^2} memory and \order{N^3} inversion cost); we therefore have to resort to further approximations. In this work we arguably use the simplest such approach, subset-of-data (SoD), in which we only retain $M$ of the original $N$ training points for inference. Depending on the dataset, already a relatively small number of data points can yield reasonable results. In practise, we choose $M$ as large as computationally feasible, and obtain the following posterior covariance approximation:
\begin{equation}
  \label{app:eq:gp_pred}
  \vSigma_{\ast, \text{SoD}} = \vK_{\ast \ast} - \vK_{\ast M} (\vK_{MM} + \vL_{MM}\inv)\inv \vK_{M \ast}.
\end{equation}
We further scale the prior precision $\delta$, which is part of the \gp kernel, by a factor proportional to $\frac{N}{M}$ to alleviate problems due to using a small subset of the data.
If we only use a subset of the data, the predictive variance is not sufficiently reduced due to the subtracted gain in \cref{app:eq:gp_pred}.
Reducing the \gp prior covariance by a factor allows to alleviate this effect.
The factor is selected on the held-out validation set in our experiments but $\frac{N}{M}$ is a good rule of thumb that always gives good performance.

Instead of a subset-of-data approach, we could use more elaborate sparse Gaussian process approximations \citep{titsias2009_variational_gp,hensman2015scalable};
however, we found that a subset-of-data worked relatively well, such that we leave other sparse approximations for future work.

\subsection{Computational considerations and complexities}
\label{app:sec:complexities}

We discuss the computational complexities of the proposed and related methods in detail in this section.
The central quantity of interest for the \ggn as well as the \gp and \glm predictives are the Jacobians.
As discussed in \cref{sec:computational_considerations}, computation and storage of Jacobians for typical network architectures is \order{PC} for a single data point.
We first discuss computation, storage, and inversion of various Laplace-\ggn posterior approximations and after discuss the \glm and \gp inference (refinement) and predictives.

\subsubsection{Laplace-\ggn}
\label{app:sec:comp_lap_ggn}
Apart from computation of the Laplace-\ggn posterior precision in \cref{eq:sigma_post}, we also need to invert it and decompose it (for sampling) and store it.
The full covariance Laplace-\ggn posterior approximation is very costly as it models the covariance between all parameters in the neural network:
we need to compute the sum and matrix multiplications in \cref{eq:sigma_post} resulting in \order{NC^2P^2} complexity.
Storing this matrix and decomposing it is in \order{P^2} and \order{P^3}, respectively.
Therefore, it can only be applied to problems where the neural network has
\order{10^4} parameters.

A cheap alternative is to use the diagonal \ggn approximation which can be computed exactly in \order{NC^2P} or approximately (using sampling) in \order{NCP}~\citep{dangel2019backpack}. It is also considered in iterative algorithms for variational inference, e.g. by~\citet{khan2018fast, osawa2019practical}
Storage, decomposition, and inversion of the diagonal is as cheap as storing one set of parameter with \order{P}.

For the \kfac posterior approximation, the network architecture decides the complexity of inference:
for simplicity, we assume a fully connected network but the analysis can be readily extended to convolutional layers or architectures.
For layer $l$ that maps from a $D_\textrm{in}^{(l)}$ to a $D_\textrm{out}^{(l)}$-dimensional representation, the corresponding block in the \ggn approximation is Kronecker factored, i.e., with $\vQ_l \kron \vW_l$ (see \cref{eq:kfac_ggn}) where $\vQ_l \in \Reals^{D_\textrm{in}^{(l)} \times D_\textrm{in}^{(l)}}$ and $\vW_l \in \Reals^{D_\textrm{out}^{(l)} \times D_\textrm{out}^{(l)}}$.
As shown in \cref{app:kronecker}, all relevant quantities can be computed by decomposing both Kronecker factors individually due to the properties of the Kronecker product.
Therefore, storage complexity is simply $\mathcal{O}\big(\sum_{l=1}^L (D_\textrm{in}^{(l)})^2 + (D_\textrm{out}^{(l)})^2\big)$ and decomposition $\mathcal{O}\big(\sum_{l=1}^L (D_\textrm{in}^{(l)})^3 + (D_\textrm{out}^{(l)})^3\big)$.
Suppose $D_l := D_\textrm{in}^{(l)}=D_\textrm{in}^{(l)}$, i.e., same in- as output dimensionality of the layer, then the \kfac approximation avoids having to handle a matrix with $D_l^4$ entries by instead handling two $D_l^2$ matrices~\citep{botev2017practical}.
Therefore, \kfac approximations are our default choice for larger scale inference on the parametric side.

Above, we discussed Laplace-\ggn approximation using the full training data set.
We can also obtain a stochastic unbiased estimate of the \ggn approximation to the Hessian by subsampling $M$ data points and scaling the \ggn by factor $\frac{N}{M}$ which is used in stochastic variational inference~\citep{khan2018fast, osawa2019practical, zhang2018noisy} and leads to a reduction in complexity where we have factor $M$ instead of $N$ and $M \ll N$.

The cost of the functional Laplace-\ggn approximation in the equivalent \gp model is dominated by the number of data points $N$ instead of the number of parameters $P$.
To obtain the full functional posterior approximation is in \order{CN^3 + CN^2P} due to $C$ independent {\gp}s~\citep{Rasmussen2006} and the computation of the $N^2$ inner products of Jacobians.
The proposed subset-of-data approximation of the kernel matrix simply leads to replacement of $N$ with $M \ll N$ in the above complexities and makes inference tractable for large data sets.

\subsubsection{\glm and \gp refinement}
To refine the \glm, we need to update the parameter vector $\vtheta \in \Reals^P$.
For example, we can use gradient-based optimization to obtain the MAP of the \glm model or alternatively construct a variational or sampling-based posterior approximation.
In any case, we need to evaluate the density or gradients which require access to a mini-batch of $N_\textrm{bs} \leq N$ Jacobians; $N_\textrm{bs} \ll N$ if we can use stochastic estimators.
Hence, one step of refinement is at least in \order{N_\textrm{bs}PC}.
In contrast, one optimization step of the corresponding neural network would be \order{N_\textrm{bs}P} instead.
While the complexity is only higher than for basic neural network training by factor $C$, ad-hoc computation of Jacobians is practically much slower in current deep learning frameworks.
An alternative is to pre-compute and cache Jacobians which is feasible on small data sets (\cref{sec:exp:uci} and \ref{app:exp_details_uci}) but leads to memory problems on larger architectures.
Therefore, we only consider it on the smaller examples.

To refine the \gp, we need to update the function vector $\vf(X) \in \Reals^{MC}$ (see \cref{app:sec:gp}) where $M \leq N$ in case of a subset approximation.
For \gp refinement, we need to compute the local linearization (\cref{eq:local_linearization}) up front which is \order{MPC}.
Following steps, e.g., gradient computations, are then typically \order{MC}.
Depending on $P, M, N_\textrm{bs}$, either parametric or functional refinement may be preferred.

\subsubsection{Predictives}

We discuss the cost of obtaining samples from the posterior predictive using the proposed \glm and \gp predictives and compare them to the \bnn predictive.
First, we describe the complexity and procedure of naive parameter-level sampling and then describe how more samples can be obtained cheaply using local reparameterization \citep{kingma2015variational}.
Local reparameterization allows us to sample the outputs of a linear model directly instead of sampling the (typically much higher dimensional) weights.%
The alternative \gp predictive immediately produces samples from the output as well but has a different complexity.

The basic form of \glm and \bnn predictives relies on parameters sampled from the Laplace-\ggn posterior and sampling depends directly on the sparsity of posterior approximation~(\dig, \kfac, full).
Here, we assume we have already inverted and decomposed the posterior precision (see \cref{app:sec:comp_lap_ggn}) which enables sampling.
Evaluating the predictive for a single sample is \order{P} for the \bnn and \order{PC} for the \glm.
Typically, however, the major cost is due to sampling parameters:
sampling once from a full covariance is \order{P^2} for, from a diagonal \order{P}, and from the \kfac posterior $\mathcal{O}\big(\sum_{l=1}^L D_\textrm{out}^{(l)}(D_\textrm{in}^{(l)})^2 + D_\textrm{in}^{(l)}(D_\textrm{out}^{(l)})^2\big)$.
The complexity of sampling from the \kfac posterior is a consequence of sampling from the matrix normal distribution~\citep{gupta2000matrix}.

For both the \glm predictive and the \bnn predictive, we can make use of local reparameterization \citep{kingma2015variational} to speed up the sampling. Local reparameterization uses the fact sums of Gaussian random variables are themselves Gaussian random variables -- instead of sampling Gaussian weights individually, we can sample their sum (the outputs) directly. For the linear \glm, we can do this for all covariances of the posterior, whereas for {\bnn}s this is only effective for diagonal covariance posteriors, as we describe in the following.

For the \bnn, we can use local reparameterization to sample vectors of smaller size from the hidden representations instead of the parameter matrices~\citep{kingma2015variational}.
However, this is only possible when we have a diagonal posterior approximation where sampling a parameter is of same complexity as a forward pass \order{P}.
Here, we consider mostly \kfac and full posterior approximations where we need to fall back to the naive procedure of sampling parameters.
For the \glm, we can directly reparameterize into the $C$-dimensional output distribution by mapping from the distribution on $\vtheta$ to a distribution on $\jacastshort(\vxstar) \vtheta$.
The complexity of this transformation again depends on the sparsity of the Laplace-\ggn approximation.
We have for the full covariance \order{C^2P+CP^2}, for diagonal \order{C^2P}, and for \kfac $\mathcal{O}\big(C^2 P + \sum_{l=1}^L D_\textrm{out}^{(l)}(D_\textrm{in}^{(l)})^2 + D_\textrm{in}^{(l)}(D_\textrm{out}^{(l)})^2\big)$.
The transformation results in a multivariate Normal distribution on the $C$ outputs.
The additional $C^2$ terms arise because we obtain a $C\times C$ functional output covariance and can be reduced to \order{C} by using only a diagonal approximation to it.
Using a full multivariate normal additionally requires decomposition of the $C \times C$ covariance in \order{C^3} while the diagonal approximation incurs no additional cost to sample outputs.
Critically, having a distribution on the $C$ outputs allows much cheaper sampling in many cases.
For example, consider a full Laplace-\ggn posterior approximation:
naive sampling in the \glm with $S$ predictive samples is \order{SP^2+SPC}.
In contrast, transforming and sampling $S$ samples directly using a diagonal output distribution is \order{CP^2+SC}.
Therefore, it scales better in the number of samples and is always more efficient once we sample more outputs than we have classes which is realistic.

To predict with the \gp posterior approximation using a subset of $M \leq N$ training points, we need to compute the \gp predictive in \cref{eq:GP_predictive}.
The output distribution (before the inverse link $g(\cdot)$), in particular its covariance, can therefore be obtained in \order{CMP + CM^2} for $C$ independent outputs;
the first term arises from computing the kernel between the test point and the $M$ training points and the second term from computing the gain term in the predictive.
Sampling from the \gp predictive is \order{SC} since we sample outputs from a diagonal Gaussian; in the regression case with a Gaussian likelihood, this additional step is not necessary.

\FloatBarrier
\clearpage
\section{Experimental details and additional results}
\label{app:experiments}

\subsection{Illustrative $1d$ example}
\label{app:exp:one_d_toy}

In \cref{sec:illustrative_example} we considered a very simple $1d$ binary classification problem with inputs $\vx_i$ given by $\{-6,-4,-2,2,4,6\}$ and corresponding labels $\vy_i$ given by $0$ for $\vx_i < 0$ and $1$ for $\vx_i > 0$ (see \cref{app:fig:toy_classification} \textit{left}). This dataset has ambiguity about the exact location of the decision boundary. We use a Bernoulli likelihood, sigmoid inverse link function and features $\vf(\vx; \vtheta) = 5 \tanh(\vw\vx + \vb)$. The model parameters $\vtheta$ are given by $\vtheta=\{\vw, \vb\}$, $\vw, \vb, \vx \in\mathbb{R}$, and we use a factorized Gaussian prior $p(\vw, \vb)=\mathcal{N}(\vw; 0, 1)\mathcal{N}(\vb; 0, 1)$. Because there is ambiguity in the data, we cannot determine the parameter values with certainty, such that a Bayesian treatment is adequate.
\begin{figure}[b]
    \centering
    \includegraphics[width=\textwidth]{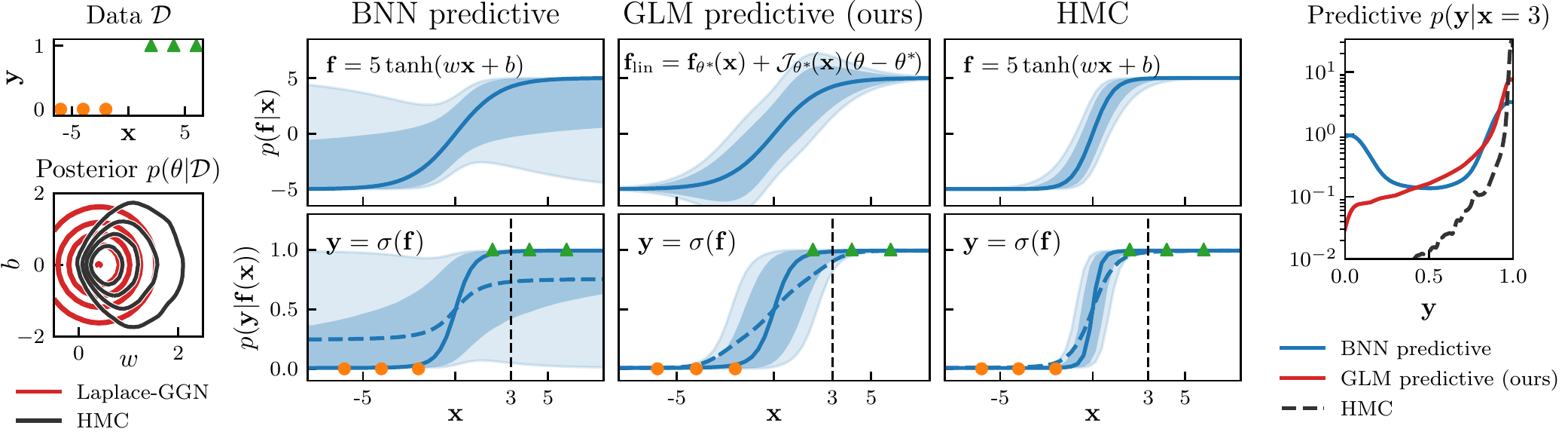}
    \vskip-0.85em
    \caption[]{Illustration how the \glm predictive overcomes underfitting of the \bnn predictive and closely resembles the true predictive (HMC). $1d$ binary classification (\protect\tikz[baseline=-0.5ex,inner sep=2pt]{\protect\node[C2] {\pgfuseplotmark{*}};} vs \protect\tikz[baseline=-0.5ex,inner sep=2pt]{\protect\node[C3] {\pgfuseplotmark{triangle*}};})
    with a $2$-parameter feature function $\vf(\vx; \vtheta) = 5\text{\texttt{tanh}}(w\vx+b)$, $\vtheta=(w, b)$, classification likelihood $p(\vy=1|\vf)=\sigma(\vf)$, and prior $p(\vtheta_i)=\mathcal{N}(0,1)$. \\
    \textit{left:} Laplace-\ggn posterior around $\thetamap$ vs. the true posterior ($10^5$ HMC samples): the Laplace-\ggn is symmetric and extends beyond the true, skewed posterior with same MAP.\\
    \textit{middle:}. The \glm predictive with \ggn posterior yields reasonable linearized preactivations $\flin$ and its predictive $p(\vy|\vf(\vx))$ closely resembles the true predictive (HMC); the \bnn predictive underfits. \protect\tikz[baseline=0ex,inner sep=0pt]{\protect\draw[C1vlight, fill] (0,-0.1) rectangle ++(0.5,0.4); \protect\draw[C1lighter, fill] (0,0.025) rectangle ++(0.5,0.15);\useasboundingbox (0,0) rectangle ++(0.5,0.15);}
    innermost $50\%$/$66\%$ of samples;
    \protect\tikz[baseline=-0.5ex,inner sep=0pt]{\protect\draw[line width=1.5pt, C1] (0,0) -- ++(0.5,0);}
    median,
    \protect\tikz[baseline=-0.5ex,inner sep=0pt]{\protect\draw[line width=1.5pt, C1, dashed] (0,0) -- ++(0.5,0);}
    mean.\\
    \textit{right:} Predictive density $p(\vy|\vx)$ at $\vx=3$: the \bnn predictive is bimodal due to the behaviour of the $\tanh$ for posterior samples with positive $w>0$ or negative $w<0$ slopes.}
    \label{app:fig:toy_classification}
\end{figure}
\begin{figure}[tb]
    \centering
    \includegraphics[width=\textwidth]{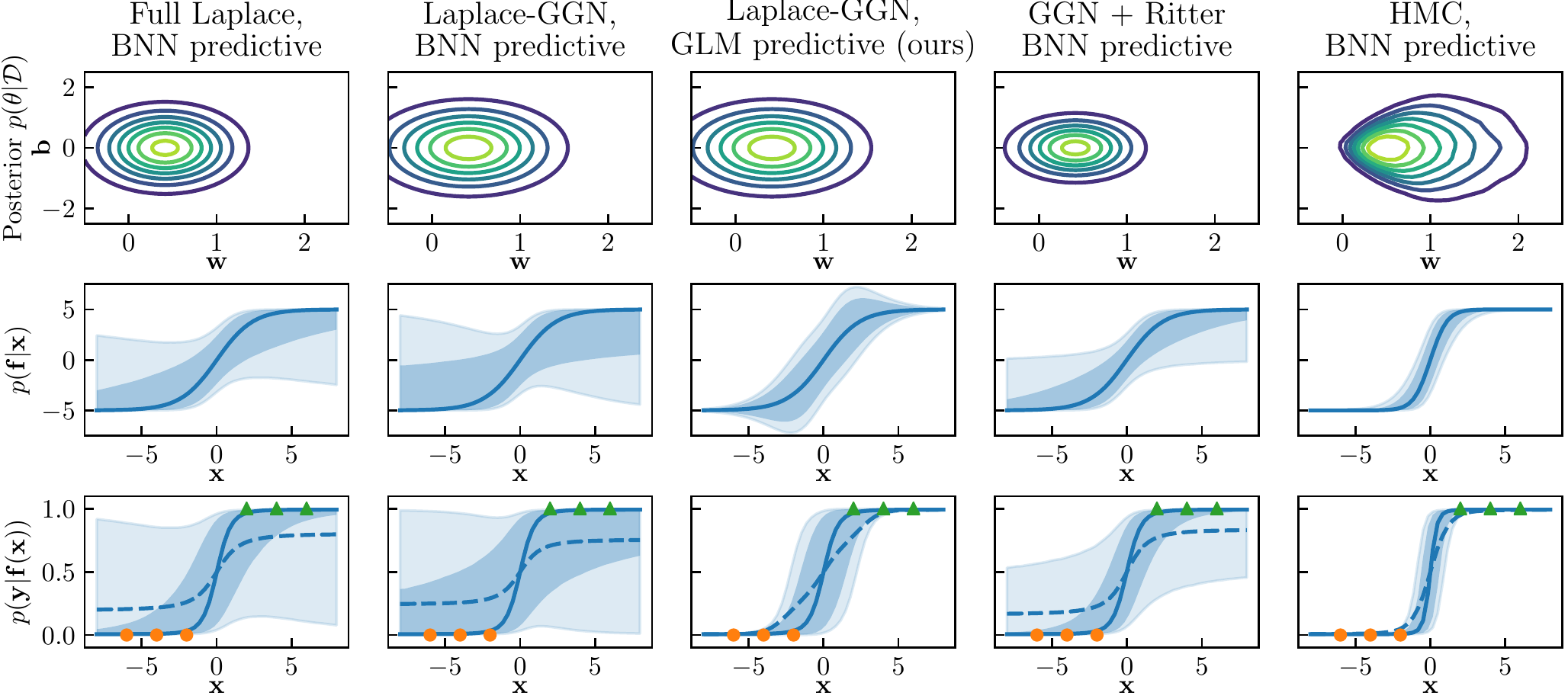}
    \caption[]{
    Continuation of \protect\cref{app:fig:toy_classification}.\\
    \textit{Left to right:} The full Laplace posterior with \bnn predictive, the Laplace-\ggn posterior with \bnn as well as \glm predictive, the dampened Laplace-\ggn posterior with \bnn predictive by \protect\citet{ritter2018scalable}, and the true (HMC) posterior ($10^5$ samples) with \bnn predictive.\\
    \textit{top:} Parameter posteriors or posterior approximations. The Laplace posteriors are symmetric and extends beyond the skewed true (HMC) posterior with same MAP, $\thetamap$. \\
    \textit{middle:} Distribution of the marginal preactivations, $p(\vf\given\vx)$. The \glm predictive with \ggn posterior yields reasonable linearized preactivations $\flin(\vx, \vtheta)$, while the other Laplace distributions already give rise to broader distributions because they use the original $\vf(\vx, \vtheta)$ combined with posteriors that extend to $\vw<0$.
    \protect\tikz[baseline=0ex,inner sep=0pt]{\protect\draw[C1vlight, fill] (0,-0.1) rectangle ++(0.5,0.4); \protect\draw[C1lighter, fill] (0,0.025) rectangle ++(0.5,0.15);\useasboundingbox (0,0) rectangle ++(0.5,0.15);}
    innermost $50\%$/$66\%$ of samples;
    \protect\tikz[baseline=-0.5ex,inner sep=0pt]{\protect\draw[line width=1.5pt, C1] (0,0) -- ++(0.5,0);}
    median. \\
    \textit{bottom:} Posterior predictives $p(\vy\given\vf(\vx))$. The \glm predictive with Laplace-\ggn posterior closely resembles the true (HMC) predictive; the \bnn predictive with any of the Laplace posteriors underfit.
    \protect\tikz[baseline=0ex,inner sep=0pt]{\protect\draw[C1vlight, fill] (0,-0.1) rectangle ++(0.5,0.4); \protect\draw[C1lighter, fill] (0,0.025) rectangle ++(0.5,0.15);\useasboundingbox (0,0) rectangle ++(0.5,0.15);}
    innermost $50\%$/$66\%$ of samples;
    \protect\tikz[baseline=-0.5ex,inner sep=0pt]{\protect\draw[line width=1.5pt, C1] (0,0) -- ++(0.5,0);}
    median,
    \protect\tikz[baseline=-0.5ex,inner sep=0pt]{\protect\draw[line width=1.5pt, C1, dashed] (0,0) -- ++(0.5,0);}
    mean.}
    \label{app:fig:1d_toy}
\end{figure}

For this low dimensional problem, we can find the ``true'' posterior $\posterior$ through HMC sampling \citep{NealHMC}, see \cref{app:fig:toy_classification} \textit{(left)}, which is symmetric \wrt the line $\vb=0$ but skewed along the $\vw$ axis and concentrated on values $\vw>0$.

Using GD we find the MAP estimate $\thetamap = (\wmap, \bmap)$ and employ the full Laplace approximation to find an approximate posterior, $q_\text{Lap}(\vtheta)$. We also compute the \ggn approximation at $\thetastar=\thetamap$ and compute the corresponding Laplace-\ggn posterior, $q_\text{\ggn}(\vtheta)$, as well as the dampened Laplace-\ggn posterior as proposed by \citet{ritter2018scalable}, $q_\text{Ritter}(\vtheta)$. As they are Gaussian approximations, they are unskewed along the $\vw$ axis but symmetric \wrt $\vb=\bmap=0$ and $\vw=\wmap$. Notably, they all put probability mass into the region $\vw < 0$. In stark contrast, the true posterior (HMC) does not put any probability mass in this region. We find that the Laplace-\ggn posterior is less concentrated than the full Laplace posterior; the dampened posterior by \citet{ritter2018scalable} is artificially concentrated and even narrower than the full Laplace posterior due to dampening. See \cref{app:fig:1d_toy} for a contour plot of all posteriors considered.

We can compute the posterior distribution over the preactivations $\vf$, $p(\vf\given\vx)$, as well as the posterior predictive for the class label $\vy$, $p(\vy\given\vf(\vx))$. We use the \bnn predictive on all posteriors and evaluate the proposed \glm predictive on the Laplace-\ggn posterior. The \glm predictive uses linearized features $\flin$ in the predictive and we show these linearized features for $p(\vf\given\vx)$ in this case, see \cref{app:fig:1d_toy} \textit{(middle row)} for the preactivations and \cref{app:fig:1d_toy} \textit{(bottom row)} for the predictives. We also show the preactivations and predictives for HMC and the Laplace-\ggn posterior (together with their feature functions) in \cref{app:fig:toy_classification}.

While the true (HMC) posterior with the corresponding \bnn predictive yields reasonable uncertainties, all Laplace posteriors using the \bnn predictive underfit or severely underfit; this is due to the posterior probability density having mass in $\vw<0$, which means that we obtain a bimodal predictive distribution. The dampened posterior \citep{ritter2018scalable} has the most concentrated posterior, such that it suffers the least. In stark contrast, the proposed \glm predictive despite using the widest posterior produces reasonable preactivations and a posterior predictive that does not underfit and is much closer to the true (HMC) predictive. The median predictions for all methods (\protect\tikz[baseline=-0.5ex,inner sep=0pt]{\protect\draw[line width=1.5pt, C1] (0,0) -- ++(0.5,0);} in \cref{app:fig:1d_toy}) saturate at $\vy\approx 0$ or $\vy\approx 1$, whereas this is only true for the mean (\protect\tikz[baseline=-0.5ex,inner sep=0pt]{\protect\draw[line width=1.5pt, C1, dashed] (0,0) -- ++(0.5,0);} in \cref{app:fig:1d_toy}) when using HMC or the proposed \glm predictive.

In \cref{app:fig:toy_classification} \textit{(right)} we show the posterior predictive densities at a specific input location $\vx=3$. While the HMC and \glm predictive have a single mode at $\vy=1$, the \bnn predictive underfits and is, in fact, bimodal with a mode at $\vy=1$ and a second mode $\vy=0$. This second underfitting mode is due to posterior samples from the mismatched region with $w<0$ as also discussed in the main text.

\FloatBarrier

\subsection{Illustrative $2d$ example: the banana dataset}
\label{app:exp_details_toy}

In \cref{sec:exp:banana}, we considered binary classification on the synthetic banana dataset. Here we provide further experimental details and results.

\subsubsection{Experimental details}
We use a synthetic dataset known as 'banana' and separate 5\% ($N=265$) of it as training data and 5\% as validation set. For NN MAP, we tune the prior precision $\delta$ using the validation dataset on a uniformly-spaced grid of 10 values in range $[0.1, 2.0]$ for all architectures with at least two layers, otherwise we use a smaller range $[0.02, 0.4]$; for BBB \citep{blundell2015weight} we use 10 log-spaced values between $10^{-3}$ and $1$. We optimize the models using full-batch gradient descent with the Adam optimizer with initial learning rate $10^{-2}$ (NN MAP) and $10^{-1}$ (BBB) for 3000 epochs, decaying the learning rate by a factor of $10$ after $2400$ and $2800$ epochs. %
We show the predictive distribution for a $100 \times 100$ grid with input features $x_1$ and $x_2$ in $[-4,4]$ range using $1000$ Monte Carlo-samples to estimate the posterior predictive distribution for all methods.

\subsubsection{Aleatoric and epistemic uncertainty}
We can decompose the overall uncertainty into aleatoric and epistemic uncertainty. The aleatoric uncertainty is due to inherent noise in the data (e.g. two or more different classes overlapping in the input domain) and will always be there regardless of how many data points we sample. Therefore, we might be able to quantify this uncertainty better by sampling more data, but we will not be able to reduce it. On the other hand, epistemic uncertainty is caused by lack of knowledge (e.g. missing data) and can be minimized by sampling more data.
Therefore, the decomposition into aleatoric and epistemic uncertainty allows us to establish to what extent a model is uncertain about its predictions is due to inherent noise in the data and to what extend the uncertainty is due to the lack of data. This distinction can be helpful in certain areas of machine learning such as active learning.

To decompose the uncertainty of a Bernoulli variable, we follow \citet{kwon2020uncertainty}, who derive the following decomposition of the total variance into aleatoric and epistemic uncertainty:

\begin{align}
    \mathrm{Var}_{p_{\glm}(\vy^\ast|\vx^\ast,\mathcal{D})}(\vy^\ast) =  \underbrace{\int\mathrm{Var}_{p_{\glm}(\vy^\ast|\vx^\ast,\vtheta)}(\vy^\ast)q(\vtheta)\calcd\vtheta}_\text{aleatoric} + \underbrace{\int[\mathbb{E}_{p_{\glm}(\vy^\ast|\vx^\ast,\vtheta)}(\vy^\ast) - \mathbb{E}_{p_{\glm}(\vy^\ast|\vx^\ast,\mathcal{D})}(\vy^\ast)]^2 q(\vtheta)\calcd\vtheta}_\text{epistemic}
\end{align}

\subsubsection{Additional experimental results}
Here, we present additional results using different posterior approximations (diagonal and full covariance Laplace-\ggn approximation and MFVI as well as MAP) and predictives (\glm predictive, \bnn predictive) as well as on different architectures.

\begin{figure}[htb]
    \centering
    \begin{tikzpicture}
    \node (a) {\includegraphics[trim=0cm 0cm 10.25cm 0cm, clip=true]{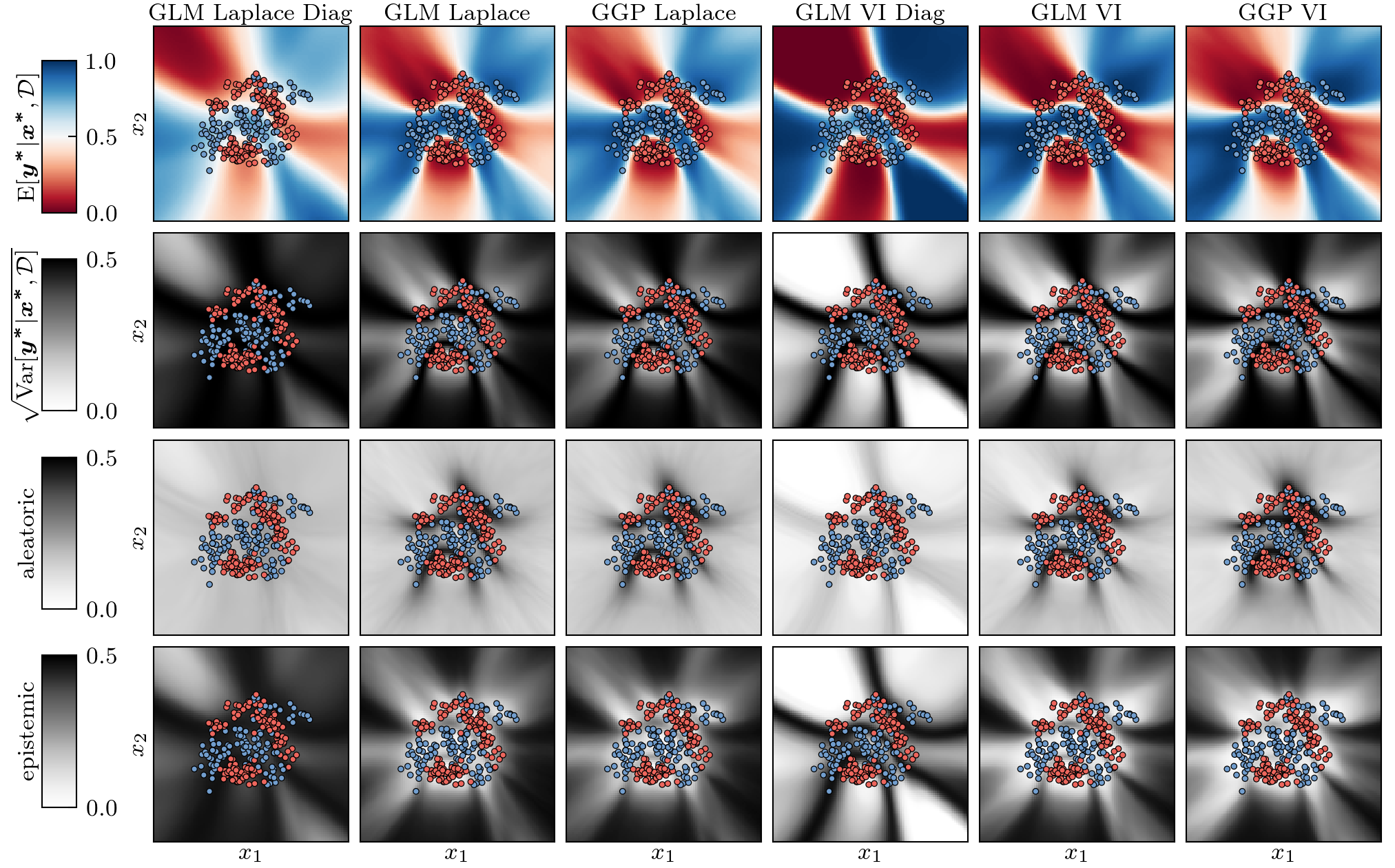}};
    \node[right=0.0of a] (b) {\includegraphics[trim=9.45cm 0cm 2.65cm 0cm, clip=true]{figures/toy_example_glm_ggp_2h50_tanh.png}};
    \end{tikzpicture}
    \vskip-1em
    \caption{\textit{From top to bottom:} Mean, standard deviation, aleatoric uncertainty, and epistemic uncertainty on the banana dataset for the \glm predictive. We compare full and diagonal covariances for the Laplace-\ggn posterior and a posterior with additional VI refinement for a network with \textbf{2 hidden layers with 50 units and \texttt{tanh} activation function}. %
    }
    \label{fig:toy_glm_ggp}
\end{figure}

\begin{figure}[h]
    \centering
    \includegraphics[scale=0.9]{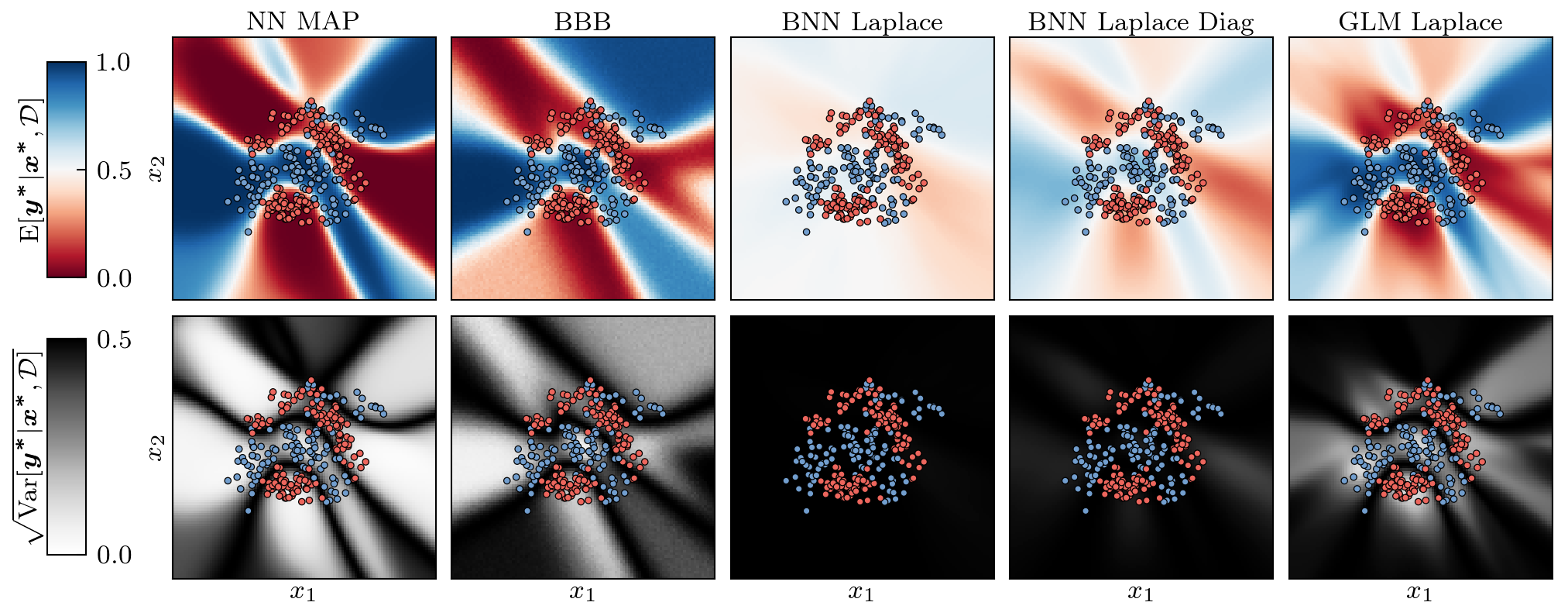}
    \vskip-0.5em
    \caption{Mean and standard deviation of the predictives on the banana dataset for a network with \textbf{1 hidden layer with 50 units  and \texttt{tanh} activation} function. A significant decrease of parameters compared to a network with 2 layers results in a decrease in confidence of NN MAP as well as the corresponding \glm predictions; the \bnn Laplace predictive improves slightly, but the variance is still greatly overestimated.}
    \label{fig:toy_1layer}
\end{figure}
\begin{figure}[h]
    \centering
    \includegraphics[scale=0.9]{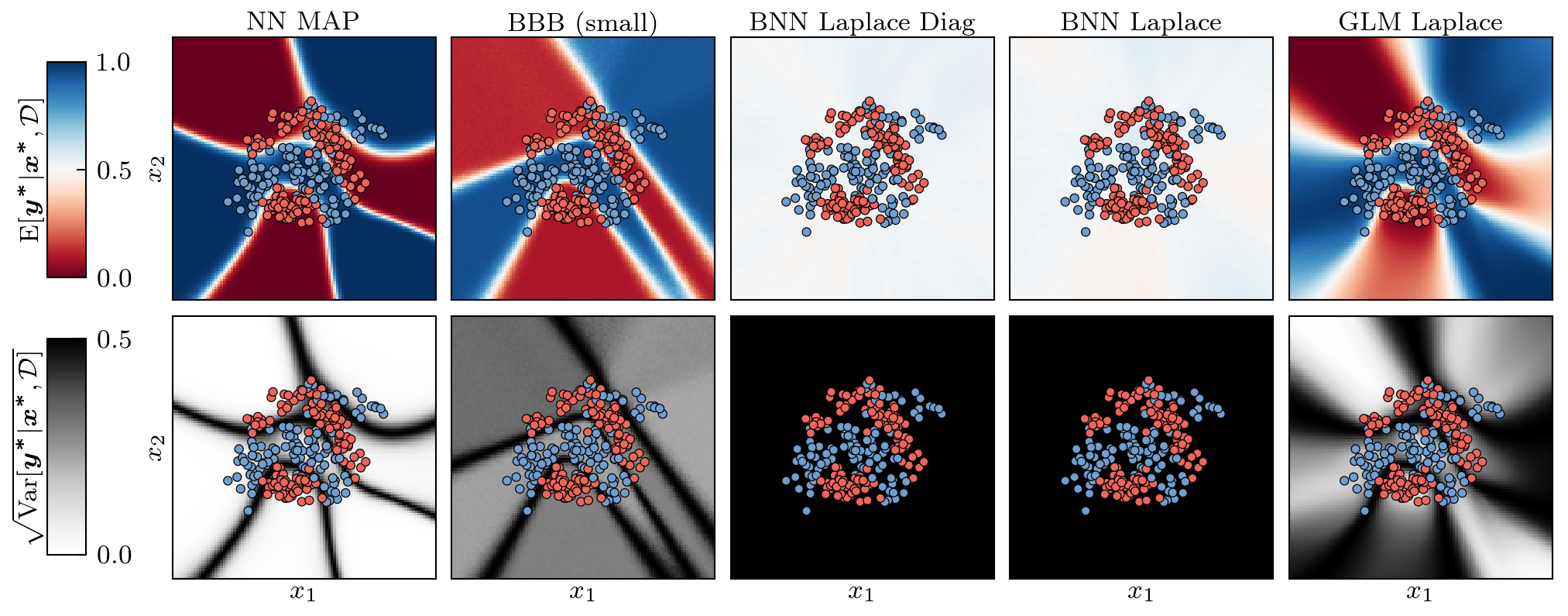}
    \vskip-0.5em
    \caption{Mean and standard deviation of the predictives on the banana dataset for a network with \textbf{3 hidden layers with 50 units each and \texttt{tanh} activation function} (with the exception of BBB where number of hidden units per layer was limited to 5 due to failing to fit the data with more units per layer). Increasing the model capacity leads to more confident predictions for the NN MAP and the corresponding \glm predictive (narrower decision boundary), yet the \glm predictive is still able to produce reasonable uncertainties away from data.}
    \label{fig:toy_3layers}
\end{figure}

In \cref{fig:toy_glm_ggp} we show the \glm with Laplace-\ggn posterior or a refined \ggn posterior with variational inference. In addition to the full covariance Laplace-\ggn, we also consider diagonal posterior approximations.
In general, the Laplace approximation results in less confident predictions than the refined posterior, which is trained using variational inference. We attribute this to the mode seeking behaviour of variational inference that likely results in a narrower posterior.

We also find that the diagonal posterior approximations generally performs worse compared to the full covariance posteriors; for the Laplace the diagonal posterior is underconfident, whereas for the refined posterior it is overconfident. In both cases the epistemic and aleatoric uncertainties are not very meaningful for the diagonal posterior. We attribute this to the diagonal approximation neglecting important correlations between some of the weights, which are captured by the full covariance posteriors. This highlights that diagonal approximations are often too crude to be useful in practise.
While the two diagonal posteriors behave very differently, we find that the full covariance versions behave relatively similar.

In \cref{fig:toy_1layer,fig:toy_3layers}, we compare the proposed \glm predictive with Laplace-\ggn posterior to the \bnn predictive with same posterior as well as to the NN MAP and mean field variational (BBB) for networks with one and three layers, respectively.

A lower number of parameters ($1$ layer, \cref{fig:toy_1layer}) slightly improves the performance of \bnn predictive, but still the variance is severely overestimated. The \glm predictive has higher variance compared to the model with two layers (\cref{fig:toy_new}) similarly to NN MAP on which it was based. Mean field VI behaves reasonably, though it tends to be overly confident away from the data.

When the model is greatly overparameterized (3 layers, \cref{fig:toy_3layers}), the \bnn predictive completely fails even for the diagonal approximation. In contrast, the \glm predictive becomes somewhat more confident (similarly to the NN MAP), though it still maintains reasonable uncertainties away from the data. Mean field VI was very hard to train in this overparameterized setting; we had to reduce the layer width dramatically in order for it not to underfit, and show results for a much narrower network, which is nonetheless overly confident. We attribute this behaviour of MFVI to the large KL penalty in the ELBO objective, which becomes more egregious the larger the number of parameters becomes relative to the number of datapoints.

\FloatBarrier
\subsection{Classification on UCI}
\label{app:exp_details_uci}

In \cref{sec:exp:uci} we considered a suite of binary and multi-class classification tasks; here we provide details about the datasets, further experimental details, as well as additional experimental results and different evaluation metrics.

We use the following 8 UCI classification data sets available from UCI Machine Learning Repository: \url{https://archive.ics.uci.edu/ml/datasets.php} for our comparison, see \cref{tab:app:uci_datasets}
\begin{table}[htb]
    \centering
    \begin{tabular}{lLLL}
    \toprule
    \textbf{Dataset} & \textbf{number of datapoints } N & \textbf{input dimension } D & \textbf{number of classes } C\\\midrule
    australian credit approval & 690 & 14 & 2\\
    breast cancer Wisconsin & 569 & 32 & 2 \\
    ionosphere & 351 & 34 & 2 \\\midrule
    glass identification & 214 & 6 & 10\\
    vehicles silhouettes & 846 & 14 & 4 \\
    waveform & 1000 & 31 & 3 \\
    digits & 1797 & 64 & 10 \\
    satellite & 6435 & 36 & 6 \\
    \bottomrule
    \end{tabular}
    \caption{Overview of UCI classification datasets used}
    \label{tab:app:uci_datasets}
\end{table}

\subsubsection{Experimental details}
We compare the neural network MAP, a BNN with mean-field variational inference trained with \textit{Bayes-by-Backprop}~\citep{blundell2015weight} (BBB) using the local reparameterization trick~\citep{kingma2015variational}, and a \bnn with Laplace approximation using the \ggn~\citep{ritter2018scalable, foresee1997gauss} (Eq.~\ref{eq:bnn_predictive}) with our \glm and \gp predictive compared to the \bnn predictive.
We compare both a refined Laplace and Gaussian variational approximation of the \glm formulation (\cref{sec:methods}) with diagonal and full posterior covariance approximations.
For all methods, we train until convergence (for BBB 5000 steps, for MAP 10000 steps) with the Adam optimizer~\citep{kingma2014adam} using a learning rate of \num{e-3} for MAP and \num{e-3} for BBB.

For refinement in the \glm, we use 1000 iterations for the Laplace approximation and 250 for the variational approximations.
For the Laplace approximation, we optimize the GLM using Adam and a learning rate of \num{e-3}.
For the variational approximation the GLM, we use a natural gradient VI algorithm to update the mean and covariance of the posterior approximation~\citep{amari1998natural, khan2018fast, zhang2018noisy} with learning rates \num{e-3} for a full and \num{e-2} for a diagonal covariance.

We split each dataset 10 times randomly into \emph{train/validation/test} sets with ratios 70\%/15\%/15\% and stratify by the labels to obtain proportional number of samples with particular classes.
For each split, we train all above-mentioned methods with 10 different prior precisions $\delta$ on a log-spaced grid from \num{e-2} to \num{e2}, except for the larger datasets \emph{satellite} and \emph{digits} where the grid is from \num{e-1} to \num{e2}.
In the resulting tables, we report the performance with the standard error on the test set after selecting the best hyperparameter $\delta$ on the validation set.
We run above experiment for two network architectures:
one hidden layer with \texttt{tanh} activation and 50 hidden units and a two-layer network with \texttt{tanh} activation and 50 units on each hidden layer.

\subsubsection{Additional results}
\cref{app:tab:2hidden} complements \cref{tab:uci} with results on accuracy and expected calibration error (ECE) which measures how well predicted uncertainty corresponds to empirical accuracy~\citep{naeini2015obtaining}.
We use $10$ bins to estimate the ECE.
The other metrics reveal that the MAP provides good performance on accuracy despite being worse than the \glm predictive on NLL; the \glm predictive is best calibrated and overall outperforms the \bnn predictive as expected.

In \cref{app:tab:1hidden}, we additionally show results on a shallower, single hidden layer model which is expected to be beneficial for MFVI since it is brittle and hard to tune.
With a single hidden layer, the performance of the all methods generally goes down slightly while the performance of MFVI goes up.
Nonetheless, the \glm predictive and MAP still perform best and are preferred to MFVI due to lower runtime but overall better performance.

\begin{table*}[ht]
  \begin{subtable}[h]{0.99\textwidth}
    \centering
    \small
    \begin{tabular}{l C C C C C C C C}
    \toprule
    \textbf{Dataset} & \textbf{NN MAP} & \textbf{MFVI} & \textbf{\bnn}& \textbf{\glm} & \textbf{\glm diag} & \textbf{\glm refine} & \textbf{\glm refine d} \\\midrule
    \textbf{australian} & \mathbf{0.31 \pms{0.01}} & 0.34 \pms{0.01} & 0.42 \pms{0.00} & \mathbf{0.32 \pms{0.02}} & 0.33 \pms{0.01} & \mathbf{0.32 \pms{0.02}} & \mathbf{0.31 \pms{0.01}}  \\
    \textbf{cancer} & \mathbf{0.11 \pms{0.02}} & \mathbf{0.11 \pms{0.01}} & 0.19 \pms{0.00} & \mathbf{0.10 \pms{0.01}} & \mathbf{0.11 \pms{0.01}} & \mathbf{0.11 \pms{0.01}} & 0.12 \pms{0.02}  \\
    \textbf{ionosphere} & 0.35 \pms{0.02} & 0.41 \pms{0.01} & 0.50 \pms{0.00} & \mathbf{0.29 \pms{0.01}} & 0.35 \pms{0.01} & 0.35 \pms{0.05} & 0.32 \pms{0.03}  \\
    \textbf{glass} & 0.95 \pms{0.03} & 1.06 \pms{0.01} & 1.41 \pms{0.00} & 0.86 \pms{0.01} & 0.99 \pms{0.01} & 0.98 \pms{0.07} & \mathbf{0.83 \pms{0.02}}  \\
    \textbf{vehicle} & 0.420 \pms{0.007} & 0.504 \pms{0.006} & 0.885 \pms{0.002} & 0.428 \pms{0.005} & 0.618 \pms{0.003} & \mathbf{0.402 \pms{0.007}} & 0.432 \pms{0.005}  \\
    \textbf{waveform} & \mathbf{0.335 \pms{0.004}} & 0.393 \pms{0.003} & 0.516 \pms{0.002} & \mathbf{0.339 \pms{0.004}} & 0.388 \pms{0.003} & \mathbf{0.335 \pms{0.004}} & 0.364 \pms{0.008}  \\
    \textbf{digits} & \mathbf{0.094 \pms{0.003}} & 0.219 \pms{0.004} & 0.875 \pms{0.002} & 0.250 \pms{0.002} & 0.409 \pms{0.002} & 0.150 \pms{0.002} & 0.149 \pms{0.008}  \\
    \textbf{satellite} & 0.230 \pms{0.002} & 0.307 \pms{0.002} & 0.482 \pms{0.001} & 0.241 \pms{0.001} & 0.327 \pms{0.002} & \mathbf{0.227 \pms{0.002}} & 0.248 \pms{0.002}  \\
    \bottomrule
    \end{tabular}
    \caption{Negative test log likelihood}
    \label{app:tab:uci_nll_2h}
\end{subtable}
\vspace{0.75em}

\begin{subtable}[h]{0.99\textwidth}
    \centering
    \small
    \begin{tabular}{l C C C C C C C C}
    \toprule
    \textbf{Dataset} & \textbf{NN MAP} & \textbf{MFVI} & \textbf{\bnn}& \textbf{\glm} & \textbf{\glm diag} & \textbf{\glm refine} & \textbf{\glm ref. d} \\\midrule
     \textbf{australian} & \mathbf{0.884 \pms{0.010}} & \mathbf{0.885 \pms{0.008}} & \mathbf{0.887 \pms{0.009}} & \mathbf{0.883 \pms{0.009}} & \mathbf{0.885 \pms{0.009}} & \mathbf{0.888 \pms{0.008}} & \mathbf{0.886 \pms{0.008}}  \\
     \textbf{cancer} & \mathbf{0.971 \pms{0.004}} & \mathbf{0.969 \pms{0.003}} & \mathbf{0.972 \pms{0.003}} & \mathbf{0.969 \pms{0.003}} & \mathbf{0.969 \pms{0.003}} & \mathbf{0.971 \pms{0.003}} & \mathbf{0.971 \pms{0.003}}  \\
     \textbf{ionosphere} & \mathbf{0.887 \pms{0.006}} & 0.879 \pms{0.007} & 0.866 \pms{0.009} & \mathbf{0.887 \pms{0.009}} & 0.883 \pms{0.008} & \mathbf{0.891 \pms{0.007}} & \mathbf{0.892 \pms{0.005}}  \\
     \textbf{glass} & \mathbf{0.684 \pms{0.010}} & 0.534 \pms{0.013} & 0.459 \pms{0.009} & \mathbf{0.678 \pms{0.013}} & 0.666 \pms{0.013} & \mathbf{0.675 \pms{0.015}} & 0.669 \pms{0.012}  \\
     \textbf{vehicle} & \mathbf{0.827 \pms{0.005}} & 0.717 \pms{0.003} & 0.712 \pms{0.003} & \mathbf{0.828 \pms{0.005}} & 0.814 \pms{0.006} & \mathbf{0.824 \pms{0.005}} & 0.820 \pms{0.005}  \\
     \textbf{waveform} & 0.855 \pms{0.003} & \mathbf{0.861 \pms{0.003}} & \mathbf{0.858 \pms{0.003}} & 0.854 \pms{0.003} & 0.852 \pms{0.004} & 0.853 \pms{0.003} & 0.849 \pms{0.003}  \\
     \textbf{digits} & \mathbf{0.974 \pms{0.001}} & 0.951 \pms{0.002} & 0.924 \pms{0.001} & 0.964 \pms{0.001} & 0.960 \pms{0.001} & 0.966 \pms{0.001} & 0.970 \pms{0.001}  \\
     \textbf{satellite} & \mathbf{0.916 \pms{0.001}} & 0.891 \pms{0.001} & 0.842 \pms{0.001} & 0.915 \pms{0.001} & 0.910 \pms{0.001} & \mathbf{0.917 \pms{0.001}} & 0.911 \pms{0.001}  \\
    \bottomrule
    \end{tabular}
    \caption{Test accuracy}
    \label{app:tab:uci_acc_2h}
\end{subtable}
\vspace{0.75em}

\begin{subtable}[h]{0.99\textwidth}
    \centering
    \small
    \begin{tabular}{l C C C C C C C C}
    \toprule
    \textbf{Dataset} & \textbf{NN MAP} & \textbf{MFVI} & \textbf{\bnn}& \textbf{\glm} & \textbf{\glm diag} & \textbf{\glm refine} & \textbf{\glm ref. d} \\\midrule
     \textbf{australian} & 0.056 \pms{0.004} & 0.099 \pms{0.005} & 0.185 \pms{0.008} & 0.055 \pms{0.005} & 0.094 \pms{0.007} & 0.060 \pms{0.005} & \mathbf{0.050 \pms{0.003}}  \\
     \textbf{cancer} & 0.033 \pms{0.002} & 0.033 \pms{0.002} & 0.127 \pms{0.003} & 0.033 \pms{0.002} & 0.053 \pms{0.003} & \mathbf{0.027 \pms{0.003}} & \mathbf{0.027 \pms{0.003}}  \\
     \textbf{ionosphere} & 0.092 \pms{0.004} & 0.155 \pms{0.009} & 0.238 \pms{0.008} & 0.089 \pms{0.003} & 0.131 \pms{0.006} & \mathbf{0.081 \pms{0.004}} & \mathbf{0.081 \pms{0.004}}  \\
     \textbf{glass} & 0.192 \pms{0.008} & \mathbf{0.140 \pms{0.008}} & 0.175 \pms{0.009} & 0.159 \pms{0.007} & 0.222 \pms{0.012} & 0.170 \pms{0.007} & 0.160 \pms{0.005}  \\
     \textbf{vehicle} & 0.064 \pms{0.003} & 0.058 \pms{0.002} & 0.266 \pms{0.003} & 0.078 \pms{0.003} & 0.210 \pms{0.005} & \mathbf{0.052 \pms{0.002}} & 0.071 \pms{0.002}  \\
     \textbf{waveform} & 0.041 \pms{0.002} & 0.134 \pms{0.002} & 0.233 \pms{0.003} & 0.049 \pms{0.002} & 0.109 \pms{0.004} & \mathbf{0.038 \pms{0.001}} & 0.049 \pms{0.002}  \\
     \textbf{digits} & \mathbf{0.016 \pms{0.001}} & 0.091 \pms{0.002} & 0.478 \pms{0.001} & 0.141 \pms{0.000} & 0.256 \pms{0.001} & 0.054 \pms{0.001} & 0.027 \pms{0.000}  \\
     \textbf{satellite} & \mathbf{0.018 \pms{0.001}} & 0.033 \pms{0.001} & 0.125 \pms{0.001} & 0.037 \pms{0.001} & 0.118 \pms{0.001} & \mathbf{0.019 \pms{0.001}} & 0.022 \pms{0.001}  \\
    \bottomrule
    \end{tabular}
    \caption{Test expected calibration error}
    \label{tab:ece2t}
\end{subtable}
\caption{Additional results on the UCI classification experiment presented in \cref{sec:exp:uci}:
We have a \textbf{2 hidden layer MLP with \texttt{tanh} activation function and 50 neurons each}.
In Table~(b), we additionally show results on accuracy and in (c) on expected calibration error (ECE).
The MAP estimate works best on most datasets in terms of accuracy but the \glm predictives are best calibrated and attain a better log likelihood.}
\label{app:tab:2hidden}
\end{table*}

\begin{table*}[ht]
  \begin{subtable}[ht]{0.99\textwidth}
    \centering
    \small
    \begin{tabular}{l C C C C C C C C}
    \toprule
    \textbf{Dataset} & \textbf{NN MAP} & \textbf{MFVI} & \textbf{\bnn}& \textbf{\glm} & \textbf{\glm diag} & \textbf{\glm refine} & \textbf{\glm ref. d} \\\midrule
 \textbf{australian} & \mathbf{0.32 \pms{0.01}} & \mathbf{0.32 \pms{0.01}} & 0.38 \pms{0.00} & \mathbf{0.32 \pms{0.01}} & 0.33 \pms{0.01} & \mathbf{0.32 \pms{0.01}} & \mathbf{0.31 \pms{0.01}}  \\
 \textbf{cancer} & 0.15 \pms{0.03} & \mathbf{0.11 \pms{0.01}} & 0.15 \pms{0.00} & \mathbf{0.10 \pms{0.01}} & \mathbf{0.11 \pms{0.01}} & 0.17 \pms{0.05} & 0.16 \pms{0.05}  \\
 \textbf{ionosphere} & 0.32 \pms{0.02} & \mathbf{0.27 \pms{0.02}} & 0.45 \pms{0.00} & 0.29 \pms{0.01} & 0.35 \pms{0.01} & \mathbf{0.27 \pms{0.02}} & \mathbf{0.26 \pms{0.02}}  \\
 \textbf{glass} & 0.91 \pms{0.03} & 0.90 \pms{0.06} & 1.25 \pms{0.00} & 0.87 \pms{0.01} & 0.96 \pms{0.01} & \mathbf{0.75 \pms{0.02}} & 0.79 \pms{0.03}  \\
 \textbf{vehicle} & 0.433 \pms{0.007} & \mathbf{0.394 \pms{0.005}} & 0.795 \pms{0.003} & 0.451 \pms{0.004} & 0.587 \pms{0.004} & \mathbf{0.396 \pms{0.007}} & 0.421 \pms{0.006}  \\
 \textbf{waveform} & \mathbf{0.338 \pms{0.004}} & 0.360 \pms{0.007} & 0.434 \pms{0.002} & \mathbf{0.342 \pms{0.004}} & 0.368 \pms{0.003} & \mathbf{0.339 \pms{0.004}} & 0.343 \pms{0.005}  \\
 \textbf{digits} & \mathbf{0.086 \pms{0.003}} & 0.137 \pms{0.004} & 0.671 \pms{0.002} & 0.256 \pms{0.002} & 0.401 \pms{0.003} & 0.170 \pms{0.011} & 0.143 \pms{0.004}  \\
 \textbf{satellite} & 0.222 \pms{0.002} & 0.274 \pms{0.002} & 0.429 \pms{0.001} & 0.240 \pms{0.001} & 0.281 \pms{0.001} & \mathbf{0.219 \pms{0.001}} & 0.238 \pms{0.002}  \\
    \bottomrule
    \end{tabular}
    \caption{Negative test log likelihood}
    \label{tab:nll1t}
\end{subtable}
\vspace{0.75em}

\begin{subtable}[ht]{0.99\textwidth}
    \centering
    \small
    \begin{tabular}{l C C C C C C C C}
    \toprule
    \textbf{Dataset} & \textbf{NN MAP} & \textbf{MFVI} & \textbf{\bnn}& \textbf{\glm} & \textbf{\glm diag} & \textbf{\glm refine} & \textbf{\glm ref. d} \\\midrule
 \textbf{australian} & \mathbf{0.881 \pms{0.010}} & \mathbf{0.883 \pms{0.008}} & \mathbf{0.888 \pms{0.008}} & \mathbf{0.882 \pms{0.010}} & 0.879 \pms{0.010} & \mathbf{0.880 \pms{0.010}} & \mathbf{0.880 \pms{0.009}}  \\
 \textbf{cancer} & 0.967 \pms{0.004} & \mathbf{0.970 \pms{0.003}} & \mathbf{0.973 \pms{0.003}} & 0.968 \pms{0.003} & \mathbf{0.972 \pms{0.003}} & 0.963 \pms{0.003} & \mathbf{0.971 \pms{0.003}}  \\
 \textbf{ionosphere} & 0.875 \pms{0.010} & \mathbf{0.909 \pms{0.007}} & 0.881 \pms{0.008} & 0.872 \pms{0.010} & 0.883 \pms{0.009} & 0.898 \pms{0.007} & 0.896 \pms{0.007}  \\
 \textbf{glass} & \mathbf{0.706 \pms{0.011}} & 0.672 \pms{0.015} & 0.550 \pms{0.014} & 0.691 \pms{0.010} & 0.691 \pms{0.013} & \mathbf{0.700 \pms{0.013}} & 0.669 \pms{0.013}  \\
 \textbf{vehicle} & 0.809 \pms{0.005} & 0.802 \pms{0.003} & 0.750 \pms{0.003} & 0.815 \pms{0.003} & 0.817 \pms{0.004} & \mathbf{0.827 \pms{0.004}} & 0.808 \pms{0.004}  \\
 \textbf{waveform} & \mathbf{0.862 \pms{0.003}} & 0.852 \pms{0.003} & 0.858 \pms{0.002} & \mathbf{0.860 \pms{0.003}} & 0.845 \pms{0.003} & 0.857 \pms{0.003} & \mathbf{0.859 \pms{0.003}}  \\
 \textbf{digits} & \mathbf{0.977 \pms{0.001}} & 0.969 \pms{0.001} & 0.954 \pms{0.001} & 0.971 \pms{0.001} & 0.963 \pms{0.001} & 0.966 \pms{0.001} & 0.967 \pms{0.001}  \\
 \textbf{satellite} & \mathbf{0.919 \pms{0.001}} & 0.900 \pms{0.001} & 0.863 \pms{0.001} & \mathbf{0.918 \pms{0.001}} & 0.917 \pms{0.001} & \mathbf{0.918 \pms{0.001}} & 0.912 \pms{0.001}  \\
    \bottomrule
    \end{tabular}
    \caption{Test accuracy}
    \label{tab:acc1t}
\end{subtable}
\vspace{0.75em}

\begin{subtable}[ht]{0.99\textwidth}
    \centering
    \small
    \begin{tabular}{l C C C C C C C C}
    \toprule
    \textbf{Dataset} & \textbf{NN MAP} & \textbf{MFVI} & \textbf{\bnn}& \textbf{\glm} & \textbf{\glm diag} & \textbf{\glm refine} & \textbf{\glm ref. d} \\\midrule
 \textbf{australian} & 0.063 \pms{0.005} & 0.061 \pms{0.003} & 0.144 \pms{0.007} & 0.061 \pms{0.006} & 0.082 \pms{0.006} & \mathbf{0.055 \pms{0.005}} & \mathbf{0.058 \pms{0.003}}  \\
 \textbf{cancer} & 0.035 \pms{0.003} & 0.032 \pms{0.003} & 0.086 \pms{0.004} & 0.034 \pms{0.003} & 0.048 \pms{0.003} & \mathbf{0.026 \pms{0.002}} & \mathbf{0.027 \pms{0.002}}  \\
 \textbf{ionosphere} & 0.080 \pms{0.006} & 0.083 \pms{0.002} & 0.214 \pms{0.006} & 0.081 \pms{0.004} & 0.132 \pms{0.007} & 0.077 \pms{0.004} & \mathbf{0.067 \pms{0.005}}  \\
 \textbf{glass} & \mathbf{0.155 \pms{0.010}} & 0.180 \pms{0.008} & 0.215 \pms{0.010} & 0.173 \pms{0.008} & 0.232 \pms{0.010} & \mathbf{0.157 \pms{0.005}} & 0.189 \pms{0.008}  \\
 \textbf{vehicle} & 0.078 \pms{0.005} & \mathbf{0.060 \pms{0.002}} & 0.255 \pms{0.003} & 0.088 \pms{0.003} & 0.181 \pms{0.005} & 0.065 \pms{0.003} & 0.070 \pms{0.003}  \\
 \textbf{waveform} & \mathbf{0.044 \pms{0.002}} & 0.064 \pms{0.002} & 0.160 \pms{0.002} & 0.055 \pms{0.001} & 0.078 \pms{0.003} & 0.052 \pms{0.002} & 0.049 \pms{0.002}  \\
 \textbf{digits} & \mathbf{0.014 \pms{0.000}} & 0.041 \pms{0.001} & 0.410 \pms{0.001} & 0.156 \pms{0.001} & 0.257 \pms{0.001} & 0.049 \pms{0.002} & 0.033 \pms{0.002}  \\
 \textbf{satellite} & 0.020 \pms{0.000} & 0.021 \pms{0.001} & 0.119 \pms{0.001} & 0.044 \pms{0.001} & 0.090 \pms{0.001} & \mathbf{0.018 \pms{0.001}} & 0.021 \pms{0.001}  \\
    \bottomrule
    \end{tabular}
    \caption{Test expected calibration error}
    \label{tab:ece1t}
\end{subtable}
\caption{Performance on UCI classification task with \textbf{a smaller single hidden layer and \texttt{tanh} activation}.
  The \glm predictives remain consistently better than the \bnn predictives.
  The \glm predictives are also preferred to MFVI, even when only considering a diagonal refined posterior which is also cheaper than MFVI.}
\label{app:tab:1hidden}
\end{table*}

\FloatBarrier

\subsection{Image classification}
\label{app:exp:image_classification}

For the large scale image benchmarks, we use several architectures, hyperparameters, and methods.
Here, we show and discuss additional results to the ones presented in \cref{sec:exp:image} and \cref{tab:image_datasets}.
After, we describe the used architectures that are similar to the ones in the DeepOBS benchmark suite~\citep{schneider2018deepobs}.
We evaluate and compare our methods on the three common image benchmarks MNIST~\citep{lecun2010mnist}, FashionMNIST~\citep{FashionMNIST}, and CIFAR-10~\citep{Cifar10}.

\subsubsection{Additional results}
In \cref{app:tab:image_datasets}, additional performance results are listed that are in line with the results presented in \cref{sec:exp:image}:
the \glm predictive is preferred across tasks, datasets, architectures, and performance metrics overall.
It provides significantly better results than the \bnn predictive as expected:
across all tasks, we find that the \bnn predictive without posterior concentration underfits extremely due to the mismatched predictive model.
Both, the proposed \glm and \gp predictive fix this problem.
The \glm predictive typically performs best in terms of accuracy and OOD detection while the \gp predictive gives strong results on NLL and is very well calibrated.
Dampening the posterior as done by \citet{ritter2018scalable} improves the performance of the \bnn predictive in some cases but not consistently;
for example, using an MLP yields bad performance despite dampening.
We additionally present results on a simple diagonal posterior approximation.
Here, the \glm predictive also works consistently better than the \bnn predictive.

\begin{table*}[ht]
    \centering
    \small
    \begin{tabular}{>{\bfseries}l >{\bfseries}l l C C C C}
    \toprule
    \textbf{Dataset} & \textbf{Model} & \textbf{Method} &  \textbf{Accuracy} \uparrow    &   \textbf{NLL} \downarrow   &   \textbf{ECE} \downarrow    &    \textbf{OD-AUC} \uparrow   \\
    \midrule
    \multirow{15}{*}{MNIST} & \multirow{7}{*}{MLP} & MAP & \mathbf{98.22 \pms{0.13}} & 0.061 \pms{0.004} & 0.006 \pms{0.001} & 0.806 \pms{0.015} \\
        &             & \bnn predictive \dig & 93.03 \pms{0.13}  & 0.369 \pms{0.003}  & 0.168 \pms{0.001}  & 0.902 \pms{0.002}  \\
        &             & \bnn predictive \kfac & 92.89\pms{0.12} & 0.576\pms{0.008} & 0.315\pms{0.006} & 0.885\pms{0.004} \\
        &             & \bnn predictive (\citeauthor{ritter2018scalable}) & 93.14\pms{0.05} & 0.304\pms{0.002} & 0.111\pms{0.003} & 0.927\pms{0.001} \\
        &             & \glm predictive \dig (\emph{ours}) & 96.58 \pms{0.20}  & 0.189 \pms{0.002}  & 0.094 \pms{0.005}  & 0.913 \pms{0.003}  \\
        &             & \glm predictive \kfac (\emph{ours}) & \mathbf{98.40\pms{0.05}} & \mathbf{0.054\pms{0.002}} & 0.007\pms{0.001} & \mathbf{0.970\pms{0.002}} \\
        &             & \gp predictive (\emph{ours}) & \mathbf{98.22\pms{0.13}} & 0.058\pms{0.001} & \mathbf{0.003\pms{0.000}} & 0.809\pms{0.014} \\
    \cmidrule{2-7}
        & \multirow{7}{*}{CNN} & MAP & \mathbf{99.40\pms{0.03}} & \mathbf{0.017\pms{0.001}} & \mathbf{0.002\pms{0.000}} & 0.989\pms{0.00} \\
        &             & \bnn predictive \dig & 97.78 \pms{0.03}  & 0.947 \pms{0.005}  & 0.576 \pms{0.002}  & 0.939 \pms{0.004}  \\
        &             & \bnn predictive \kfac & 97.75\pms{0.08} & 1.176\pms{0.026} & 0.660\pms{0.007} & 0.938\pms{0.003} \\
        &             & \bnn predictive (\citeauthor{ritter2018scalable}) & 99.06\pms{0.04} & 0.055\pms{0.001} & 0.031\pms{0.001} & \mathbf{0.993\pms{0.001}} \\
        &             & \glm predictive \dig (\emph{ours}) & 98.12 \pms{0.05}  & 0.160 \pms{0.002}  & 0.103 \pms{0.002}  & 0.981 \pms{0.001}  \\
        &             & \glm predictive \kfac (\emph{ours}) & \mathbf{99.40\pms{0.02}} & \mathbf{0.017\pms{0.001}} & \mathbf{0.003\pms{0.000}} & 0.990\pms{0.001} \\
        &             & \gp predictive (\emph{ours}) & \mathbf{99.41\pms{0.03}} & \mathbf{0.016\pms{0.001}} & \mathbf{0.002\pms{0.000}} & 0.989\pms{0.001} \\
    \midrule
    \multirow{15}{*}{FMNIST} & \multirow{7}{*}{MLP} & MAP & 88.09\pms{0.10} & 0.347\pms{0.005} & 0.026\pms{0.004} & 0.869\pms{0.006} \\
        &             & \bnn predictive \dig & 83.08 \pms{0.14}  & 0.602 \pms{0.002}  & 0.170 \pms{0.002}  & 0.906 \pms{0.009}  \\
        &             & \bnn predictive \kfac & 81.98\pms{0.28} & 0.842\pms{0.009} & 0.319\pms{0.004} & 0.919\pms{0.009} \\
        &             & \bnn predictive (\citeauthor{ritter2018scalable}) & 83.34\pms{0.13} & 0.518\pms{0.003} & 0.094\pms{0.002} & \mathbf{0.949\pms{0.006}} \\
        &             & \glm predictive \dig (\emph{ours}) & 87.50 \pms{0.22}  & 0.418 \pms{0.004}  & 0.110 \pms{0.002}  & 0.910 \pms{0.003}  \\
        &             & \glm predictive \kfac (\emph{ours}) & \mathbf{88.64\pms{0.27}} & \mathbf{0.340\pms{0.006}} & 0.021\pms{0.004} & 0.907\pms{0.020} \\
        &             & \gp predictive (\emph{ours}) & 88.04\pms{0.09} & \mathbf{0.338\pms{0.003}} & \mathbf{0.010\pms{0.001}} & 0.889\pms{0.006} \\
    \cmidrule{2-7}
        & \multirow{7}{*}{CNN} & MAP & 91.39\pms{0.11} & 0.258\pms{0.004} & 0.017\pms{0.001} & 0.864\pms{0.014} \\
        &             & \bnn predictive \dig & 84.37 \pms{0.11}  & 0.809 \pms{0.002}  & 0.340 \pms{0.002}  & 0.924 \pms{0.007}  \\
        &             & \bnn predictive \kfac & 84.42\pms{0.12} & 0.942\pms{0.016} & 0.411\pms{0.008} & 0.945\pms{0.002} \\
        &             & \bnn predictive (\citeauthor{ritter2018scalable}) & 91.20\pms{0.07} & 0.265\pms{0.004} & 0.024\pms{0.002} & 0.947\pms{0.006} \\
        &             & \glm predictive \dig (\emph{ours}) & 89.45 \pms{0.14}  & 0.397 \pms{0.002}  & 0.136 \pms{0.002}  & 0.944 \pms{0.007}  \\
        &             & \glm predictive \kfac (\emph{ours}) & \mathbf{92.25\pms{0.10}} & \mathbf{0.244\pms{0.003}} & 0.012\pms{0.003} & \mathbf{0.955\pms{0.006}} \\
        &             & \gp predictive (\emph{ours}) & 91.36\pms{0.11} & 0.250\pms{0.004} & \mathbf{0.007\pms{0.001}} & 0.918\pms{0.010} \\
    \midrule
    \multirow{15}{*}{CIFAR10} & \multirow{7}{*}{CNN} & MAP & \mathbf{77.41\pms{0.06}} & 0.680\pms{0.004} & 0.045\pms{0.004} & \mathbf{0.809\pms{0.006}} \\
        &             & \bnn predictive \dig & 69.38 \pms{0.25}  & 1.333 \pms{0.003}  & 0.373 \pms{0.004}  & 0.514 \pms{0.008}  \\
        &             & \bnn predictive \kfac & 72.49\pms{0.20} & 1.274\pms{0.010} & 0.390\pms{0.003} & 0.548\pms{0.005} \\
        &             & \bnn predictive (\citeauthor{ritter2018scalable}) & 77.38\pms{0.06} & \mathbf{0.661\pms{0.003}} & \mathbf{0.012\pms{0.003}} & 0.796\pms{0.005} \\
        &             & \glm predictive \dig (\emph{ours}) & \mathbf{77.47 \pms{0.11}} & 0.783 \pms{0.001}  & 0.184 \pms{0.003}  & 0.699 \pms{0.009}  \\
        &             & \glm predictive \kfac (\emph{ours}) & \mathbf{77.44\pms{0.05}} & 0.679\pms{0.004} & 0.043\pms{0.003} & \mathbf{0.809\pms{0.005}} \\
        &             & \gp predictive (\emph{ours}) & \mathbf{77.42\pms{0.05}} & \mathbf{0.660\pms{0.003}} & \mathbf{0.013\pms{0.003}} & 0.798\pms{0.005} \\
    \cmidrule{2-7}
        & \multirow{7}{*}{AllCNN} & MAP & 80.92\pms{0.32} & 0.605\pms{0.007} & 0.066\pms{0.004} & 0.792\pms{0.008} \\
        &             & \bnn predictive \dig & 21.71 \pms{0.79}  & 2.110 \pms{0.021}  & 0.097 \pms{0.011}  & 0.700 \pms{0.019}  \\
        &             & \bnn predictive \kfac & 21.74\pms{0.80} & 2.114\pms{0.021} & 0.095\pms{0.012} & 0.689\pms{0.020} \\
        &             & \bnn predictive (\citeauthor{ritter2018scalable}) & 80.78\pms{0.36} & 0.588\pms{0.005} & 0.052\pms{0.005} & 0.783\pms{0.007} \\
        &             & \glm predictive \dig (\emph{ours}) & 80.26 \pms{0.30}  & 0.968 \pms{0.006}  & 0.348 \pms{0.005} & 0.743 \pms{0.022}  \\
        &             & \glm predictive \kfac (\emph{ours}) & \mathbf{81.37\pms{0.15}} & 0.601\pms{0.008} & 0.084\pms{0.010} & \mathbf{0.843\pms{0.016}} \\
        &             & \gp predictive (\emph{ours}) & 81.01\pms{0.32} & \mathbf{0.555\pms{0.008}} & \mathbf{0.017\pms{0.003}} & 0.820\pms{0.013} \\
    \bottomrule
    \end{tabular}

    \caption{Accuracy, negative test log likelihood (NLL), and expected calibration error (ECE) on the test set as well as area under the curve for out-of-distribution detection (OD-AUC).
    Bold numbers indicate best performance on particular dataset-model combination.
    The proposed methods (\glm and \gp) typically take the first and second best spot in terms of accuracy and in most cases perform best in the other measured performances as well.
    In terms of out-of-distribution detection, the \glm provides overall the best performance.}
    \label{app:tab:image_datasets}
\end{table*}

\subsubsection{Architectures}
We use different architectures depending on the data sets:
For MNIST and FMNIST, we use one fully connected and one mixed architecture (convolutional and fully connected layers).
On CIFAR-10, we use the same mixed architecture and a fully convolutional architecture.

For the MNIST and FMNIST data sets, we train a multilayer perceptron (MLP) with $4$ hidden layers of sizes $[1024, 512, 256, 128]$.
The MLP is an interesting benchmark due to the number of parameters per layer and the potential problem of distorted predictives.

On both data sets, we also use a standard mixed architecture with three convolutional and three fully connected layers as implemented in the DeepOBS benchmark suite~\citep{schneider2018deepobs}.

On CIFAR-10, we do not use an MLP as it is hard to fit and performs significantly worse than convolutional architectures.
Instead, we use the same mixed architecture (CNN) and another fully convolutional architecture (AllCNN), which is again as standardized by \citet{schneider2018deepobs}.
The AllCNN architecture uses 9 convolutional blocks followed by average pooling.
In contrast to \citet{schneider2018deepobs}, we do not use additional dropout.
All convolutional layers are followed by \texttt{ReLU} activation and fully connected layers by \texttt{tanh} activation, except for the final layers.

\subsubsection{MAP estimation}
We train all our models in \texttt{pytorch}~\citep{paszke2017automatic} using the Adam optimizer~\citep{kingma2014adam} on the log joint objective in \cref{eq:map_objective}.
We use the default Adam learning rate of $10^{-3}$ and run the training procedure for $500$ epochs to ensure convergence and use a batch size of $512$.
In line with standard Bayesian deep learning methods, we assume an isotropic Gaussian prior $p(\vtheta) = \mathcal{N} (0, \delta\inv \vI)$ on all parameters. %
To estimate standard errors presented in the results, we run every experiment with $5$ different random seeds.
Except in the CIFAR-10 experiment with the fully convolutional architecture which is expensive to train, we train each for $16$ values of $\delta$ log-spaced between $10^{-2}$ and $10^3$.
On the fully convolutional network, we train $6$ networks.

\subsubsection{Inference}
We construct the considered Laplace-\ggn posterior approximations on all the trained models and evaluate their performance on a validation set to select the best prior precision $\delta$ per model-dataset-method combination.
The \kfac and \dig approximations to the \ggn are computed using \texttt{backpack} for \texttt{pytorch}~\citep{dangel2019backpack}.
The \gp subset of data Laplace approximation is constructed only on the best MAP network per seed and we compare the impact of different subset sizes $M$ below.

\subsubsection{Ablation study: Subset-of-data \gp approximation}
\begin{figure*}[ht]
    \centering
    \begin{tikzpicture}
      \node (left) {\includegraphics[width=2.2in]{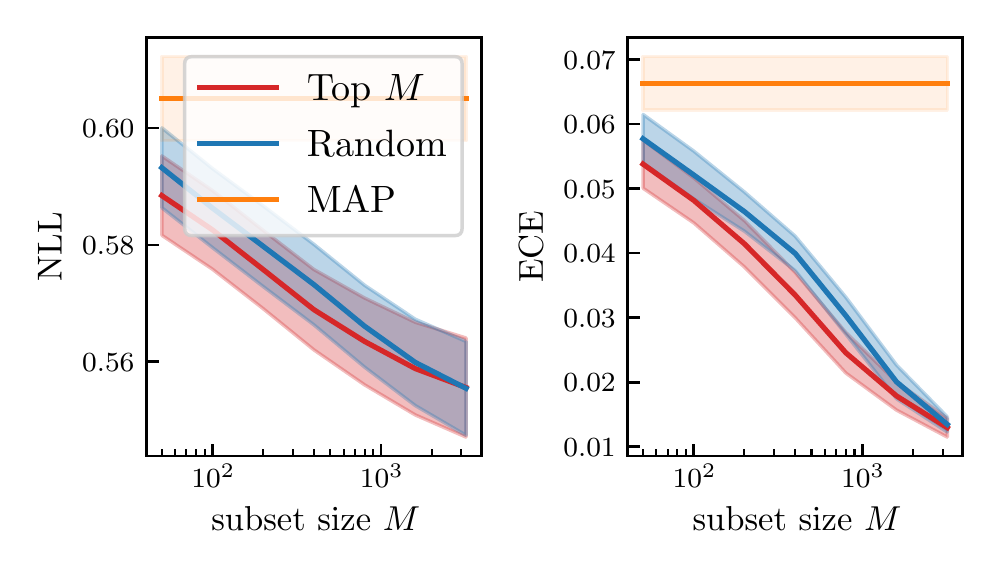}};
      \node[right=-.3cm of left] (mid){\includegraphics[width=2.2in]{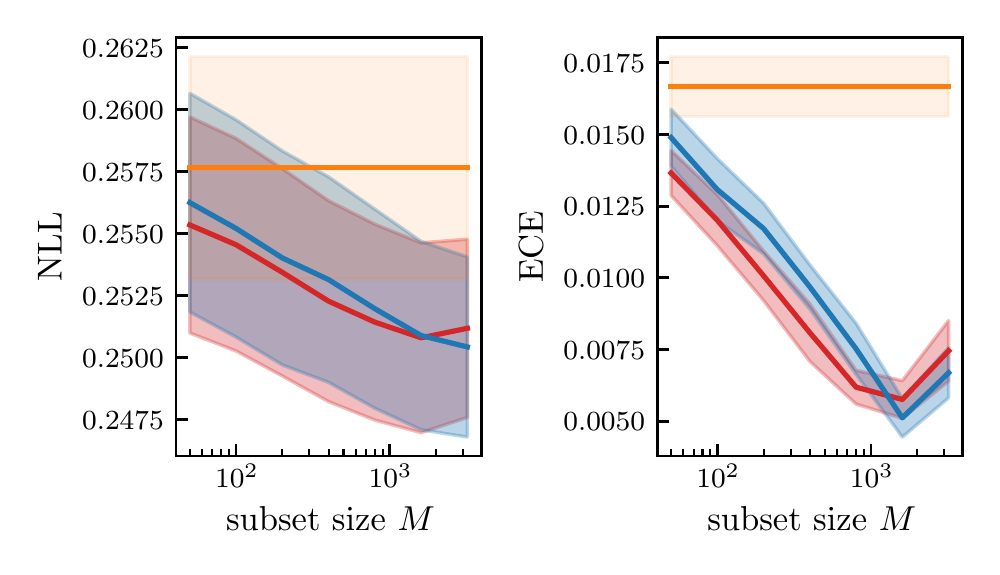}};
      \node[right=-.3cm of mid] (right){\includegraphics[width=2.2in]{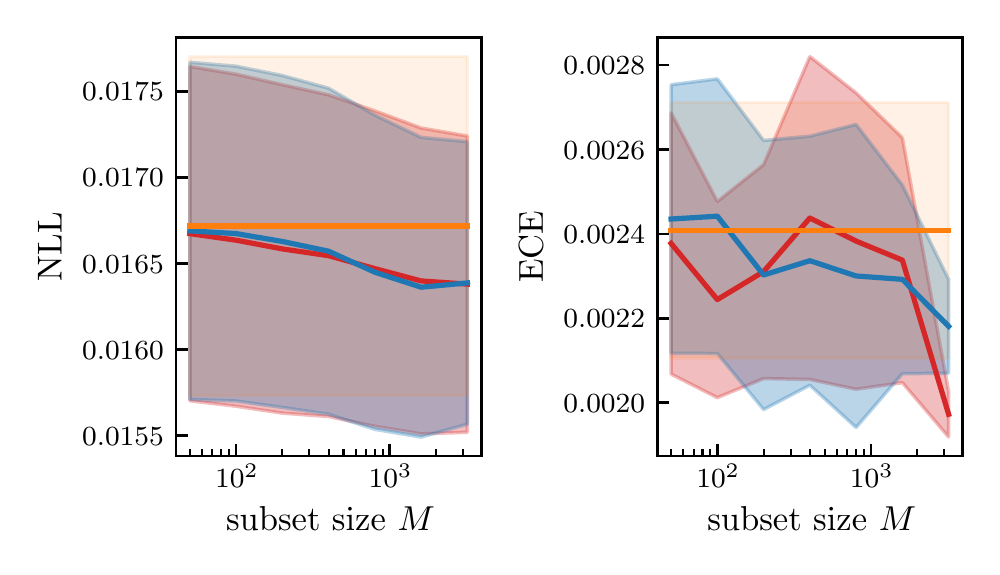}};
    \node[anchor=south] at ($(left.north)+(0.35,-0.1)$) {CIFAR10};
    \node[anchor=south] at ($(mid.north)+(0.35,-0.1)$) {FMNIST};
    \node[anchor=south] at ($(right.north)+(0.35,-0.1)$) {MNIST};
    \end{tikzpicture}
    \caption{Performance vs. subset size $M$ for the \gp predictive.
      We compare the \gp predictive to the respective MAP baseline in terms of negative log likelihood (NLL) and expected calibration error (ECE) and compare a \emph{random}
        (\protect\tikz[baseline=.5ex,inner sep=0pt]{\protect\draw[line width=1.5pt, C1, name path=A] (0,0.3) -- ++(0.75,0); \protect\draw[name path=B, draw=none] (0,0) -- ++(0.75,0);\protect\tikzfillbetween[of=A and B]{C1, opacity=0.3};})
      to the \emph{top-$M$}
      (\protect\tikz[baseline=.5ex,inner sep=0pt]{\protect\draw[line width=1.5pt, C4, name path=A] (0,0.3) -- ++(0.75,0); \protect\draw[name path=B, draw=none] (0,0) -- ++(0.75,0);\protect\tikzfillbetween[of=A and B]{C4, opacity=0.3};}, \citep{pan2020continual}) subset selection approach.
      Both selection methods lead to performance improvements where possible, especially with larger subset sizes $M$.
      On CIFAR10 and FMNIST we can obtain considerable improvements over the MAP, especially in terms of calibration where the improvement is up to $6$-fold.
      On MNIST, the MAP is already sufficient and there is no room for improvement.
      We show results on CIFAR10 (\textit{left}), FMNIST (\textit{middle}), and MNIST (\textit{right}) using the best performing architectures on the respective tasks (see \cref{app:tab:image_datasets}).}
    \label{app:fig:gp_sod}
\end{figure*}

As described in \cref{app:sec:gp}, we use a subset of data \gp predictive with $M$ training samples to construct the posterior and the posterior predictive. Here, we investigate the influence of this subset size on performance, see \cref{app:fig:gp_sod} for a comparison against the neural network MAP.
As already observed in the experimental results, the \gp predictive greatly improves the calibration in terms of ECE but also improves the NLL over the MAP baseline network even with as few as $M=50$ data points.
Further, the improvement typically consistently increases with improved subset size $M$.
We consider two strategies for selecting the subset of data.
We either sample $M$ training data points at random or use the top-$M$ selection criterion as proposed by \citet{pan2020continual} for a \gp arising from \dnntogp.
In line with \citet{pan2020continual}, we find that selecting the top-$M$ relevant points is mostly beneficial for small $M$.
However, we also find that selecting $M$ points uniformly at random is sufficient for good performance and is slightly cheaper as we do not need to pass through the entire training data.

\subsubsection[]{Ablation study: Predictive distribution for dampened posteriors \protect\citep{ritter2018scalable}}
\label{app:sec:ablation_kfac}
Finally, we investigate how well the dampened \kfac Laplace-\ggn posterior by \citet{ritter2018scalable} performs when used in conjunction with our proposed \glm predictive instead of the \bnn predictive originally used by the authors, see \cref{app:tab:kfac_damp} for results. The performance is not consistent over all metrics; on accuracy, the \glm predictive always improves over the \bnn predictive, especially on fully connected networks.

Note that dampening corresponds to an ad-hoc approximation that is not necessary in this setting; using the un-dampened posterior with \glm predictive generally performs similar or better on most datasets.

\begin{table*}[ht]
    \centering
    \small
    \begin{tabular}{>{\bfseries}l >{\bfseries}l l C C C C}
    \toprule
    \textbf{Dataset} & \textbf{Model} & \textbf{Method} &  \textbf{Accuracy} \uparrow    &   \textbf{NLL} \downarrow   &   \textbf{ECE} \downarrow    &    \textbf{OD-AUC} \uparrow   \\
    \midrule
    \multirow{4}{*}{MNIST} & \multirow{2}{*}{MLP} & \bnn predictive  & 93.14\pms{0.05} & 0.304\pms{0.002} & 0.111\pms{0.003} & \mathbf{0.927\pms{0.001}} \\
                           &             & \glm predictive & \mathbf{98.39 \pms{0.05}}  & \mathbf{0.052 \pms{0.002}}  & \mathbf{0.004 \pms{0.000}}  & 0.921 \pms{0.010}  \\
    \cmidrule{2-7}
        & \multirow{2}{*}{CNN} & \bnn predictive & 99.06\pms{0.04} & 0.055\pms{0.001} & 0.031\pms{0.001} & \mathbf{0.993\pms{0.001}} \\
        &             & \glm predictive  & \mathbf{99.41 \pms{0.03}} & \mathbf{0.016 \pms{0.001}}  & \mathbf{0.002 \pms{0.000}}  & 0.990 \pms{0.001} \\
    \midrule
    \multirow{4}{*}{FMNIST} & \multirow{2}{*}{MLP} & \bnn predictive & 83.34\pms{0.13} & 0.518\pms{0.003} & 0.094\pms{0.002} & \mathbf{0.949\pms{0.006}} \\
                            &             & \glm predictive  & \mathbf{88.77 \pms{0.26}} & \mathbf{0.337 \pms{0.003}} & \mathbf{0.022 \pms{0.004}}  & 0.922 \pms{0.020}  \\
    \cmidrule{2-7}
                            & \multirow{2}{*}{CNN} & \bnn predictive  & 91.20\pms{0.07} & 0.265\pms{0.004} & 0.024\pms{0.002} & \mathbf{0.947\pms{0.006}} \\
        &             & \glm predictive  & \mathbf{92.24 \pms{0.10}} & \mathbf{0.242 \pms{0.004}} & \mathbf{0.016 \pms{0.001}} & 0.905 \pms{0.010}  \\
    \midrule
    \multirow{4}{*}{CIFAR10} & \multirow{2}{*}{CNN} & \bnn predictive & 77.38\pms{0.06} & \mathbf{0.661\pms{0.003}} & \mathbf{0.012\pms{0.003}} & 0.796\pms{0.005} \\
        &             & \glm predictive  & \mathbf{77.44 \pms{0.05}}  & 0.679 \pms{0.004}  & 0.044 \pms{0.003}  & \mathbf{0.810 \pms{0.005}} \\
    \cmidrule{2-7}
        & \multirow{2}{*}{AllCNN} & \bnn predictive  & 80.78\pms{0.36} & \mathbf{0.588\pms{0.005}} & \mathbf{0.052\pms{0.005}} & 0.783\pms{0.007} \\
        &             & \glm predictive & \mathbf{81.23 \pms{0.14}}  & 0.599 \pms{0.008}  & 0.077 \pms{0.008}  & \mathbf{0.843 \pms{0.016}}  \\
    \bottomrule
    \end{tabular}
    \caption{Performance comparison of \bnn and \glm predictive using the dampened posterior proposed by \citep{ritter2018scalable}.
      On accuracy, the \glm performs strictly better while on other metrics there is a trade-off: either one of the methods is better on OOD-detection or on NLL and ECE but not on both.
    We find that the \glm predictive also works with the dampenend posterior and in that case is still most of the times the better alternative.}
    \label{app:tab:kfac_damp}
\end{table*}

\FloatBarrier

\subsection{Out-of-distribution detection}
\label{app:exp:ood}

We evaluate the predictives on out-of-distribution (OOD) detection on the following in-distribution (ID)/OOD pairs: MNIST/FMNIST, FMNIST/MNIST, and CIFAR10/SVHN.
Following \citet{osawa2019practical}, we compare the predictive entropies of the distribution on in-distribution (ID) data vs OOD data.
To that end, we compute the predictive entropy of the output probability vector $\vp$ that is either estimated by sampling;  a point estimate for the MAP case can be computed as $-\sum_{c=1}^C p_c \log (p_c)$.
The predictive entropy is at its maximum in case of a uniform output distribution.
Higher predictive entropies are desired for OOD data and lower entropies for ID data.
We then use the entropy as a score to distinguish ID and OOD data~\citep{osawa2019practical}.
Varying the score threshold, we can estimate the area under the receiver-operator characteristic as a measure of performance (OOD-AUC).

\begin{figure*}[ht]
    \centering
    \includegraphics[width=6.7in]{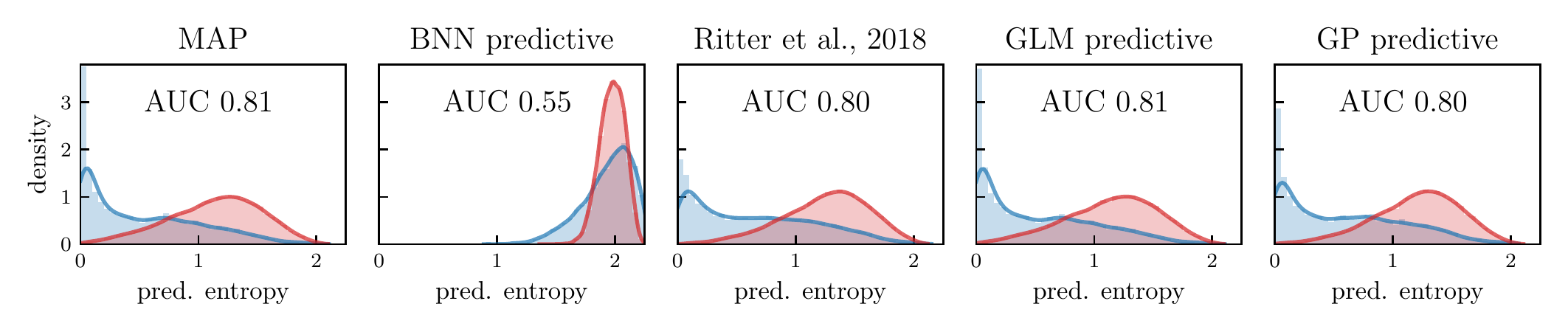}
    \vspace{-1em}
    \caption{In-distribution
    (CIFAR10~\protect\tikz[baseline=.5ex,inner sep=0pt]{\protect\draw[line width=1.5pt, C1, name path=A] (0,0.3) -- ++(0.75,0); \protect\draw[name path=B, draw=none] (0,0) -- ++(0.75,0);\protect\tikzfillbetween[of=A and B]{C1, opacity=0.3};})
    vs out-of-distribution (SVHN~\protect\tikz[baseline=.5ex,inner sep=0pt]{\protect\draw[line width=1.5pt, C4, name path=A] (0,0.3) -- ++(0.75,0); \protect\draw[name path=B, draw=none] (0,0) -- ++(0.75,0);\protect\tikzfillbetween[of=A and B]{C4, opacity=0.3};}) detection and calibration on CIFAR10 using a convolutional architecture with 2 fully connected classification layers.
  All predictive distributions are well-calibrated and detect OOD data except for the underconfident vanilla \bnn predictive.}
    \label{app:fig:app:cifar10_ood}
    \vspace{-1em}
\end{figure*}

\begin{figure*}[ht]
    \centering
    \includegraphics[width=6.7in]{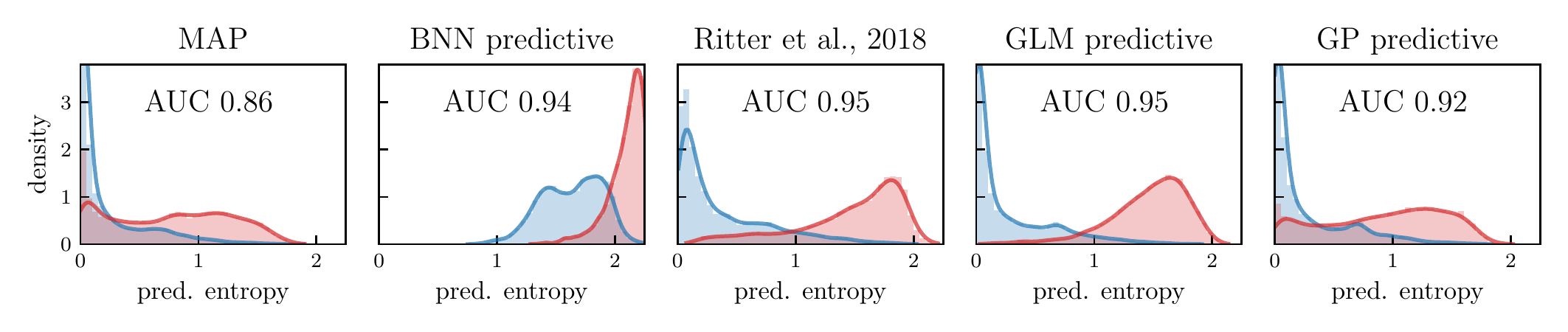}
    \vspace{-1em}
    \caption{In-distribution
    (FMNIST~\protect\tikz[baseline=.5ex,inner sep=0pt]{\protect\draw[line width=1.5pt, C1, name path=A] (0,0.3) -- ++(0.75,0); \protect\draw[name path=B, draw=none] (0,0) -- ++(0.75,0);\protect\tikzfillbetween[of=A and B]{C1, opacity=0.3};})
    vs out-of-distribution (MNIST~\protect\tikz[baseline=.5ex,inner sep=0pt]{\protect\draw[line width=1.5pt, C4, name path=A] (0,0.3) -- ++(0.75,0); \protect\draw[name path=B, draw=none] (0,0) -- ++(0.75,0);\protect\tikzfillbetween[of=A and B]{C4, opacity=0.3};}) detection and calibration on FMNIST using convolutional architecture with 2 fully connected classification layers on top.
    MAP is overconfident while BNN predictive is underconfident but still detects out-of-distribution data well.
  The \glm predictive performs best.}
  \label{fig:app:fmnist_ood_cnn}
    \vspace{-1em}
\end{figure*}

\begin{figure*}[ht]
    \centering
    \includegraphics[width=6.7in]{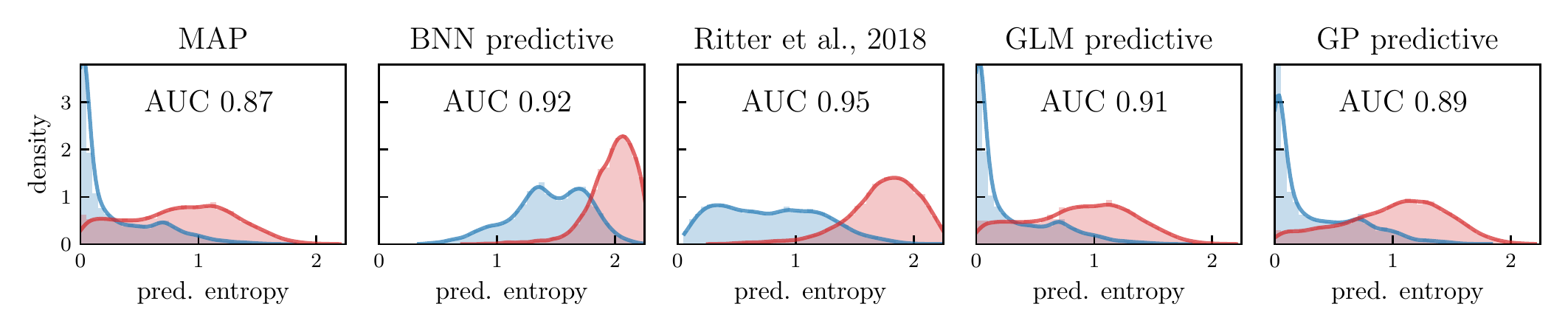}
    \vspace{-1em}
    \caption{In-distribution
    (FMNIST~\protect\tikz[baseline=.5ex,inner sep=0pt]{\protect\draw[line width=1.5pt, C1, name path=A] (0,0.3) -- ++(0.75,0); \protect\draw[name path=B, draw=none] (0,0) -- ++(0.75,0);\protect\tikzfillbetween[of=A and B]{C1, opacity=0.3};})
    vs out-of-distribution (MNIST~\protect\tikz[baseline=.5ex,inner sep=0pt]{\protect\draw[line width=1.5pt, C4, name path=A] (0,0.3) -- ++(0.75,0); \protect\draw[name path=B, draw=none] (0,0) -- ++(0.75,0);\protect\tikzfillbetween[of=A and B]{C4, opacity=0.3};}) detection and calibration on FMNIST using fully connected architecture.
    \bnn predictives are underconfident but detect OOD data well.
  The \gp predictive is both well-calibrated and good on OOD detection.}
    \label{fig:app:fmnist_ood_mlp}
    \vspace{-1em}
\end{figure*}

\begin{figure*}[ht]
    \centering
    \includegraphics[width=6.7in]{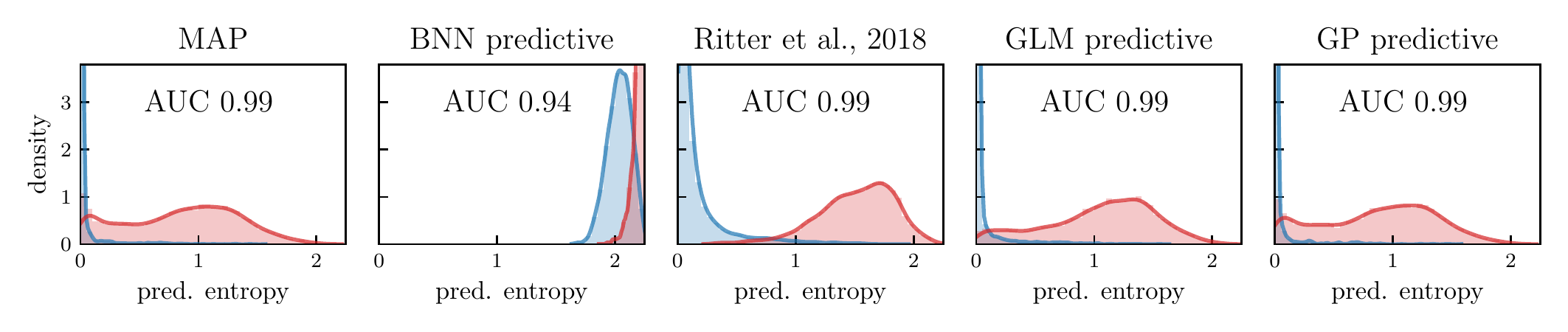}
    \vspace{-1em}
    \caption{In-distribution
    (MNIST~\protect\tikz[baseline=.5ex,inner sep=0pt]{\protect\draw[line width=1.5pt, C1, name path=A] (0,0.3) -- ++(0.75,0); \protect\draw[name path=B, draw=none] (0,0) -- ++(0.75,0);\protect\tikzfillbetween[of=A and B]{C1, opacity=0.3};})
    vs out-of-distribution (FMNIST~\protect\tikz[baseline=.5ex,inner sep=0pt]{\protect\draw[line width=1.5pt, C4, name path=A] (0,0.3) -- ++(0.75,0); \protect\draw[name path=B, draw=none] (0,0) -- ++(0.75,0);\protect\tikzfillbetween[of=A and B]{C4, opacity=0.3};}) detection and calibration on MNIST using convolutional architecture with 2 fully connected classification layers.
  MAP, \glm, and \gp predictives are perfecty calibrated and detect OOD data while \bnn predictives are slighty suboptimal.}
  \label{fig:app:mnist_ood_cnn}
    \vspace{-1em}
\end{figure*}

\begin{figure*}[ht]
    \centering
    \includegraphics[width=6.7in]{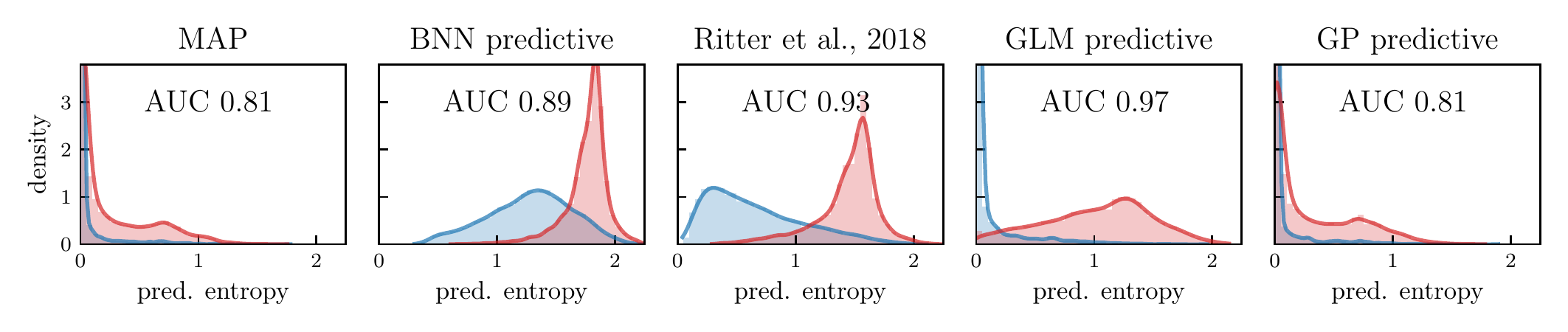}
    \vspace{-1em}
    \caption{In-distribution
    (MNIST~\protect\tikz[baseline=.5ex,inner sep=0pt]{\protect\draw[line width=1.5pt, C1, name path=A] (0,0.3) -- ++(0.75,0); \protect\draw[name path=B, draw=none] (0,0) -- ++(0.75,0);\protect\tikzfillbetween[of=A and B]{C1, opacity=0.3};})
    vs out-of-distribution (FMNIST~\protect\tikz[baseline=.5ex,inner sep=0pt]{\protect\draw[line width=1.5pt, C4, name path=A] (0,0.3) -- ++(0.75,0); \protect\draw[name path=B, draw=none] (0,0) -- ++(0.75,0);\protect\tikzfillbetween[of=A and B]{C4, opacity=0.3};}) detection and calibration on MNIST using fully connected architecture.
    \glm predictive is best on OOD detection and \gp is best calibrated.
  \bnn predictives are underconfident while the MAP is overconfident.}
  \label{fig:app:mnist_ood_mlp}
    \vspace{-1em}
\end{figure*}

Figs.~\ref{app:fig:app:cifar10_ood} to~\ref{fig:app:mnist_ood_mlp} show histograms of ID and OOD predictive entropies as well as the OOD-AUC on the different ID/OOD dataset pairs and different architectures.
For CIFAR10, we additionally provide results on a mixed architecture (\cref{app:fig:app:cifar10_ood}).
On MNIST and FMNIST, we show results on the MLP and mixed architectures:
MLP is used in \cref{fig:app:fmnist_ood_mlp} and \cref{fig:app:mnist_ood_mlp};
mixed architecture in \cref{fig:app:fmnist_ood_cnn} and \cref{fig:app:mnist_ood_cnn}.

Overall, the \glm predictive provides the best OOD detection performance followed by the dampened \bnn predictive by~\citep{ritter2018scalable}.
However as the histograms suggest, their predictive is underconfident which is reflected in the results on calibration (ECE) in \cref{app:tab:image_datasets}.
The naive \bnn predictive performs worst overall.
As for other metrics, the \gp predictive consistently improves over the MAP estimate.

\end{document}